\documentclass[preprint,12pt]{elsarticle}

\usepackage{amssymb}

\usepackage{comment}
\usepackage{xurl}
\usepackage{amsmath}
\usepackage{booktabs}
\usepackage{bm}
\usepackage{multirow}
\usepackage[export]{adjustbox}
\usepackage{subfigure}
\usepackage{hyperref}
\usepackage{footmisc}
\usepackage{cleveref}
\usepackage[section]{placeins} 
\usepackage{tablefootnote}
\usepackage[inline]{enumitem}
\usepackage{makecell}
\usepackage{longtable}
\usepackage[dvipsnames]{xcolor}

\usepackage{setspace}

\usepackage{filecontents}
\usepackage[font=small,labelfont=bf]{caption} 

\newcommand\minput[1]{%
  \input{#1}%
  \ifhmode\ifnum\lastnodetype=11 \unskip\fi\fi}

\journal{Neurocomputing}

\begin{document}

\begin{frontmatter}



\title{Tabular Data Generation Models: An In-Depth Survey and Performance Benchmarks with Extensive Tuning}

\author[rennes,orange]{G. Charbel N. Kindji\corref{cor1}}
\author[orange]{Lina M. Rojas-Barahona}
\author[rennes]{Elisa Fromont}
\author[orange]{Tanguy Urvoy}
\affiliation[rennes]{organization={Univ Rennes, IUF, Inria, CNRS, IRISA},
            city={Rennes},
            postcode={35000},
            country={France}}

\affiliation[orange]{organization={Orange Labs},
            city={Lannion},
            postcode={22300},
            country={France}}

\author{}
\cortext[cor1]{Corresponding author. Email: charbel.kindji.orange@gmail.com}


\begin{abstract}

Generating realistic, safe, and useful tabular data is important for downstream tasks such as privacy preserving, imputation, oversampling, explainability, and simulation. However, the structure of tabular data, marked by heterogeneous types, non-smooth distributions, complex feature dependencies, and categorical imbalance, poses significant challenges. Although many generative approaches have been proposed, a fair and unified evaluation across datasets remains missing. This work benchmarks five recent model families on 16 diverse datasets (average 80K rows), with careful optimization of hyperparameters, feature encodings, and architectures. We show that dataset-specific tuning leads to substantial performance gains, particularly for diffusion-based models. We further introduce constrained hyperparameter spaces that retain competitive performance while significantly reducing tuning cost, enabling efficient model selection under fixed GPU budgets. As future perspectives, we can study cross-domain and cross-table generation.

\end{abstract}


\end{frontmatter}



\section{Introduction}
\label{s:intro}

The demand for generative models is rapidly increasing. Particularly, generating realistic, safe, and useful tabular data is crucial for industries, wherein this type of data is most common. Among the direct applications of tabular data generation we can cite data privacy, imputation, 
oversampling, explainability or simulation~\cite{libes2017issues,surveytabularDNN22}. 
Because of their ability to learn valuable data representations, another significant application of these models is  pre-training and fine-tuning for various downstream tasks \cite{zhang2019table2vec, Chitlangia2022, zhang2023generative}.
However, generating high-quality tabular data presents several technical challenges that are not encountered with text or images~\cite{grinsztajn2022why}.
First, tabular columns are encoded through heterogeneous data types
with distributions that are often non-smooth with mixed (continuous/discrete) behaviours and various modalities.
There can also be complex dependencies between columns, and the categorical features are often highly imbalanced.
Finally, the wide variety of problems represented in tabular form makes it challenging to establish a universal data encoding and architecture suitable for pre-training across all scenarios.

To handle these challenges, many models have been proposed in the literature, which are often evaluated on different datasets with inconsistent metrics, tuning and training budgets. Indeed, the performance of the allegedly best tabular generation models seem very unstable from one dataset to another. Moreover, these models seem quite sensitive to the feature-encodings and hyperparameter-choices made by the authors.

In this work we propose a unified evaluation benchmark for tabular data generation. Our main contributions are :
\begin{itemize}
    \item We present the state-of-the-art and propose the following typology of the existing models : \textit{non-iterative} and iterative neural, which groups \textit{auto-regressive} and \textit{diffusion} models; as well as \textit{non-neural} models.
    \item We study the impact of dataset-specific feature  encoding, hyperparameter and  architecture  tuning on tabular data generation models. For each model we answer the following questions: 
        \begin{enumerate*}[label=(\roman*)]
            \item is it worth optimizing the hyperparameters/preprocessing specifically for each dataset?
            \item can we propose a reduced search space that fits well for all datasets?
            \item is there a clear trade-off between training/sampling costs, and synthetic data quality?
        \end{enumerate*}
    \item  We benchmark $5$ models that are representative of the recent literature on $16$ datasets with a strict $3$-fold cross-validation procedure. The datasets were chosen based on their size, purpose and diversity.
    \item We consider in our benchmark both extensive and limited-budget optimizations for each fold and dataset through hundreds of trials.

\end{itemize}

It is worth noting that this paper is not intended to be a survey. Rather, it benchmarks selected models for tabular data generation through \textit{meticulous fine-grained tuning}.
Previous surveys on tabular data generation are either focused on privacy~\cite{mckay2019comparative, arnold2020really, tao2022benchmarking, hu2023sok, du2024towards, zabergja2024tabular} or do not cover recent diffusion models~\cite{surveytabularDNN22, fonseca2023tabular}. 
Most survey only report the author's evaluations with default hyperparameters. A few benchmarks like \cite{kotelnikov2023tabddpm, du2024towards} perform hyperparameter search, with simple train/validation/test split, reduced search space, and small number of trials (usually from 20 to 50).
Instead, we provide large-scale extensive benchmarks. We select representative models of the literature and we evaluate them on distinct datasets in cross validation, in which for each fold and dataset, we optimized the model's hyperparameters, feature encoding, and architecture through hundreds of trials. 
Unlike previous work, our benchmarks compare not only the \textit{realism}, the \textit{utility} and the \textit{anonymity}, but also the \textit{costs} and the \textit{carbon footprint}.  

Finally, we analyze the results of these benchmarks and derive some interesting insights about the models. While diffusion-based models like TabSyn and TabDDPM~\cite{zhang2023mixed} generally outperform other models when left unconstrained, they do not significantly surpass their simpler counterparts when tuning and training budgets are limited. This is because models that do not rely on Transformers have a smaller memory footprint, allowing for more thorough optimization within the same GPU budget.

The paper is structured as follows. First we present the models' typology in Section~\ref{sec:sota}. The benchmark's challengers, metrics, datasets and other experimental settings are presented in section~\ref{sec:settings}. The extensive and limited-budget benchmarks are detailed in Sections~\ref{sec:results} and ~\ref{sec:hpsens} respectively. Finally the conclusions is presented in Section~\ref{sec:concl} and the future work in Section~\ref{sec:future}.

\section{Typology of Models for Tabular Data Generation}
\label{sec:sota}

Tabular data generation is a booming research field which gives birth every month to a host of new data synthesis algorithms.
In this section we survey the existing families of tabular data generation methods with a specific focus on the models that we selected for our study. This includes models already covered in \cite{surveytabularDNN22} and \cite{fonseca2023tabular} as well as more recent diffusion-based models and LLM-based ones \cite{fang2024large}. 
We chose models known for their strong performance, widespread usage, and availability of code that can be easily adapted for both architecture and hyperparameter tuning.

The following neural and non-neural approaches have been proposed for tabular data generation. Among the neural approaches, we make a distinction between the ``\textit{push-forward}'' models which directly map noise into data, and the iterative models which require a decoding phase.

\subsection{Non-iterative Neural Models}
\label{s:nn}

The most popular non-iterative or ``\textit{push-forward}'' neural networks for tabular data generation are Variational Auto-Encoders (VAE)~\cite{kingma2014auto} and Generative Adversarial Networks (GAN)~\cite{goodfellow2020generative}. 
A few papers also consider self-normalizing flows \cite{tabak2013family}. One of the simplest VAE architecture for tabular data is TVAE~\cite{CTGAN}. In its original implementation~\cite{SDV}, it consists of a one-hot encoding for categorical variables coupled with a Gaussian Mixture Model normalization scheme (GMM) for continuous features. The encoder/decoder architecture is a simple stack of linear layers. Several variants have been proposed to improve from this baseline. In \cite{ma2020vaem} the GMM normalization is replaced by a two-step training that first fits the marginals then fits the inter-dependencies. In \cite{zhao2021ctab, c2st22}, several normalization schemes were tested as a replacement for the GMM normalization.
Other variants of VAE use differentiable oblivious trees to ensure privacy \cite{vardhan2020generating}.
In \cite{liu2023goggle} the VAE is coupled with a Graph Neural Network (GNN).

The most popular method to generate tabular data is certainly 
adversarial training \cite{Jordon2018PATEGANGS, park2018data,camino2018generating,baowaly2019synthesizing,chen2019faketables, CTGAN,koivu2020synthetic,zhao2021ctab,sauber2022use, fonseca2023tabular, watson2023adversarial}. 
It would not be exaggerated to affirm that every exotic variant of GANs has been tested on tabular data generation, but the most successful architectures seem to be the ones based on Wasserstein GANs~\cite{arjovsky2017wasserstein} such as CTGAN~\cite{CTGAN}. CTGAN is the base architecture that we selected for our benchmark.
As mentioned in \cite{zhao2021ctab, c2st22}, the feature encoding scheme is critical, especially for numerical features. For this reason we customized the CTGAN code as we did for TVAE, to allow for a choice of architecture and feature encoders.

\subsection{Iterative Neural Models}
\label{sec:iterative}

It has been shown that iterative generative models (either auto-regressive or by diffusion) almost systematically outperform push forward models when it comes to raw text, sound, or image generation~\cite{genAIsurvey22}. We confirm here that tabular data do not escape this rule. However, this performance comes with a cost: the decoding phase is often slow and highly energy consuming.

\subsubsection{Auto-regressive Language Models}

After the recent breakthrough of large language models (LLM)~\cite{ouyang2022training}, the usage of token-based language models to generate tabular data seems inevitable.
Their main advantage against prior \textit{ad-hoc} models is that they come without specific feature encodings for numerical or categorical columns: these values are directly fed to the model as raw sequences of tokens.
Even if there is no clear agreement yet on how tabular instances should be serialized for LLMs, this general encoding ability opens the possibility for a pretraining/finetuning paradigm on heterogeneous tabular datasets \cite{zhang2019table2vec, zhang2023generative}.

In a recent preprint survey \cite{fang2024large}, the authors  tried to map the exuberant 
flow of preprint papers on tabular data and LLMs.
Several of these preprints present only prompt engineering tricks that generate small tables, nonetheless some of the proposed models seem promising at a larger scale.
To name a few of them, GReaT~\cite{great23} (for "Generation of Realistic Tabular data") proposes to fine-tunes GPT-2 \cite{radford4language} for tabular data generation.
REaLTabFormer~\cite{solatorio2023realtabformer} 
 extends GReaT to multiple tables with shared indexes, and TAPTAP~\cite{zhang2023generative} experiments a pretraining of GReaT on 450 open tabular datasets.
 
We made a few experiments with GReaT and its variants and confirmed the remarks of 
\cite{zhang2023mixed, du2024towards}, which state that they struggle to capture the joint probability distribution on datasets where the categorical values names do not carry semantic information. 
Given the prohibitive computational cost of LLMs, the large number of hyperparameters and serialization schemes to consider, as well as the problematic fact that some datasets are already covered (i.e. used in the training data) by the foundation models training sets, we decided to postpone their evaluation for a future work.

\subsubsection{Diffusion models}
\label{s:diffusion}

Another recent impressing breakthrough in generative modeling, especially in image generation, was the apparition of diffusion models \cite{sohl2015deep, ho2020denoising, song2020score, dhariwal2021diffusion, karras2022elucidating}.
These models learn to iteratively transform random Gaussian noise to a sample from an unknown data distribution. This transformation is defined through a forward and a backward diffusion processes. The first one iteratively adds Gaussian noise to samples, while the second one iteratively “denoises” samples from the Gaussian distribution to obtain samples from the data distribution. 
The transposition of these models to tabular data gave rise to powerful synthesizers:
TableDiffusion~\cite{truda2023generating}, Stasy \cite{kim_stasy_2023}, CoDi~\cite{CoDi}, and TabDDPM \cite{kotelnikov2023tabddpm}.
In TabSyn \cite{zhang2023mixed}, a more recent proposal,  the authors transposed the idea of \cite{vahdat2021score,rombach2022high} to make use of a transformer-based VAE in order to embed the diffusion in a latent space. We selected both {TabDDPM} and {TabSyn} for our benchmark.

\subsection{Non-neural Models}
\label{s:notnn}

The most common statistical approaches are based on \textit{copulas} and \textit{Probabilistic Graphical Models} (PGMs). 
{Copulas}~\cite{Nels06} are functions that join or “couple” multivariate distribution functions to their one-dimensional marginals. They have been widely adopted for tabular data generation because they allow modeling the marginals and the feature inter-dependencies separately~\cite{SDV, li2020sync, kamthe2021copula, Meyer_and_Nagler_2021, Meyer_and_Nagler_and_Hogan_2021}.
Nevertheless, parametric copulas have been shown to perform poorly on high-dimension data synthesis problems \cite{CTGAN, c2st22}.

On the other hand PGMs can model variable dependencies in high-dimension spaces \cite{pmlr-v97-mckenna19a, zhang2021privsyn, fonseca2023tabular, Dahmen2019SynSysAS, kerneladasyn, pmlr-v151-baak22a}. This approach has been shown to be quite efficient in the data privacy community \cite{McKenna_Miklau_Sheldon_2021, tao2022benchmarking, hu2023sok, du2024towards}. {However, it often requires a prior knowledge on the dependency graph because graph inference from data is inefficient in high dimension, especially if the sample size is small~\cite{Jewson2022Graphical}.}

The fact that ensembles of trees remain state of the art for predictive tasks on tabular data \cite{grinsztajn2022why} motivated some interesting attempts to mimic the neural generative approaches with decision trees. It gave rise to adversarial forests \cite{watson2023adversarial}, Forest-Flow, and Forest-VP (a variance-preserving diffusion algorithm) \cite{jolicoeur2023generating}.
 We tested Forest-VP but we encountered a severe scalability issue: contrary to neural diffusion models, introduced in Section~\ref{s:diffusion}, where the noise level is combined with the input as an auxiliary variable, the Forest-VP algorithm trains a different ensemble of trees for each level of noise.

Another simple, but quite efficient way to generate new tabular data is to interpolate between existing instances. This  geometric ``nearest neighbors" approach called \textit{Synthetic Minority Over-sampling Technique} (SMOTE)~\cite{chawla2002smote,smoter,gsmoter} is frequently used to resample instances before training predictive models on unbalanced datasets.
It proceeds by picking a random instance with a fixed target value and finding its $k$ nearest neighbors. New data points are then generated by interpolation between these neighbors.
Although very simple, this model is a solid baseline for tabular data generation as shown in \cite{kotelnikov2023tabddpm}.

\section{Benchmarks Settings}
\label{sec:settings}

We first present the selected challengers and the various evaluation metrics. Then, we present the optimization framework that we developed and we 
 discuss the choices made to wrap the different challengers into this framework. Finally, We present the datasets.

\subsection{Selected Challengers}
We selected five models for our benchmarks, two iterative neural: TVAE, CTGAN; two diffusion models: TabDDPM, and TabSyn; and one non-neural: SMOTE (with a variant unconditional SMOTE, namely ucSMOTE). We followed \cite{c2st22} and used a customized  version of TVAE which allows for an optimized choice of the architecture and of the feature encoder.

For each of these algorithms we had to carefully examinate the code to optimize large scale hyperparameters, features encoding and architecture.
We also reported the trivial baseline which consists in resampling directly the train set, namely ``\textit{Train~Copy}".

\subsection{Evaluation Metrics}
\label{s:metrics}

Our purpose is to assess the 
quality of tabular data generation though multiple facets. These facets can be summarized with four questions:
\begin{enumerate*}[label=(\roman*)]
    \item is synthetic data realistic? does it respect the original distribution's traits?
    \item is it useful? \textit{i.e.} can it be used to train machine learning models?
    \item do synthesis preserve training data anonymity? does it overfit?
    \item what are the model's costs and CO$_2$ impacts?
\end{enumerate*}
Each of these questions is related to specific metrics.

\begin{figure}[htbp]
    \centering
    \includegraphics[width=0.75\textwidth]{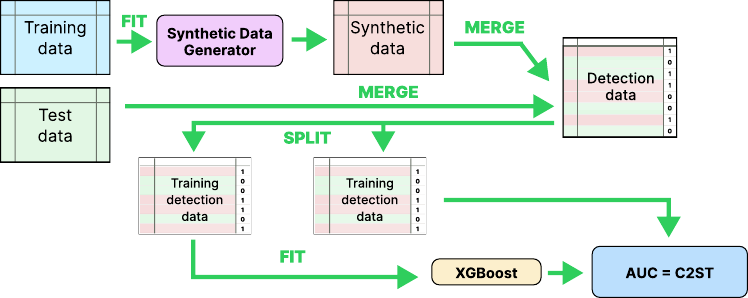}
    \caption{C2ST Metric calculation.}
    \label{fig:c2st}
\end{figure}

\paragraph{The first and most important question is the realism of the generated data}
We rely on \textit{Classifier Two Sample Test}\footnote{Also mentioned as \textit{Detection test} in SDMetrics \url{https://docs.sdv.dev/sdmetrics}} (C2ST) \cite{lopez2016revisiting} as a primary metric to address it. This metric is also the one that we used for hyperparameter optimization.
It evaluates the performance of a classifier at discerning real data from synthetic data\footnote{This is similar to what is done in GANs but with a fresh dataset. An overfitted GAN can indeed have a poor C2ST on test set.}. To compute C2ST we use the same protocol as in \cite{c2st22} where the computed value is the mean ROC-AUC of XGBoost \cite{chen2016xgboost} over three folds. A C2ST around $1/2$ means that XGBoost is unable to discern the test set from the generated set. A high C2ST means on the contrary that XGBoost is able to detect easily the fake data.
The C2ST calculation procedure is summarized in Figure \ref{fig:c2st}.

We also consider two other statistical metrics for data realism: \textit{column-wise similarity} and \textit{pair-wise correlation}. To compute these metrics we used the SDMetrics library \cite{SDV}.
The {column-wise similarity} measures how accurately the synthetic data captures the shape of each column distribution individually. It is reported as "Shape" in the result tables. 
{Pair-wise correlation} on the other hand captures how each column varies with each other. Pair-wise correlation is reported as "Pair" in the result tables.
A naive generator that would assume independence of the columns might have a high "Shape" score but it would have a low "Pair" score.

\begin{figure}[htbp]
    \centering
    \includegraphics[width=0.75\textwidth]{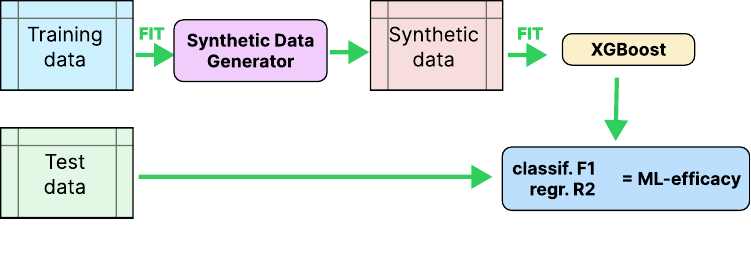}
    \caption{ML-Efficacy Metric calculation.}
    \label{fig:mleff}
\end{figure}

\paragraph{The second important question is the utility of generated data} It is commonly measured by ML-Efficacy which evaluates 
the performance of a predictive model trained on synthetic data. To compute ML-Efficacy we use the same protocol as in \cite{kotelnikov2023tabddpm} where CatBoost \cite{prokhorenkova2018catboost} is used as a predictor. We report the F1 score for classification tasks and the normalized R2 score for regression tasks. The procedure is summarized in 
Figure~\ref{fig:mleff}.
It is important to evaluate the degradation of these scores against the ones obtained when training directly on real data (via \textit{Train~Copy}): a large degradation means a low utility.

\begin{figure}[htbp]
    \centering
    \includegraphics[width=0.75\textwidth]{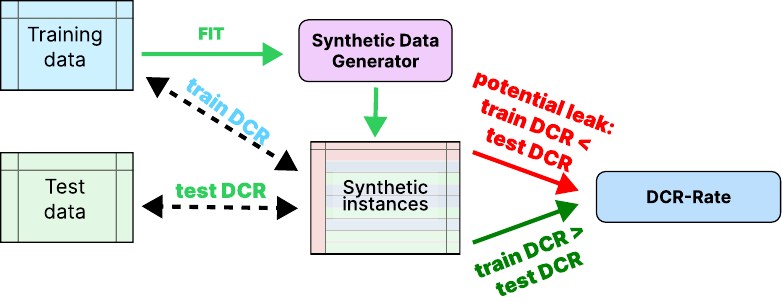}
    \caption{{DCR-Rate} Metric calculation.}
    \label{fig:dcrrate}
\end{figure}

\paragraph{The third question is anonymity} We want to make sure that our generative model will not leak sensitive information by recopying or over-fitting the training instances. To do so we measure respectively the minimum distances of each generated instance to the train and test sets. The \textit{Distance to Closest Record} Rate (DCR-Rate) counts the proportion of generated instances that are closer to train set than test set \cite{platzer2021holdout}. The procedure is summarized in 
Figure~\ref{fig:dcrrate}.

A synthetic dataset is considered safe if it has a {DCR-Rate} that is close to $1/2$. On the other hand, a plain copy of the train set, as \textit{Train Copy} baseline does, would have a {DCR-Rate} of $1$. This metric does not guarantee against all privacy breaches, but it provides a reasonable safeguard and ranking criterion for the models.
It is also informative about overfitting, as an overfitted model would have samples that are systematically closer to the train set than a model with more generalization capabilities.

\paragraph{The fourth question is the models' costs and their carbon impacts}
\label{sec:costest}
It requires an accurate estimate of three cost values: \textit{time},   \textit{energy consumption}, and \textit{CO$_2$ impact}. Ideally, these values should be estimated for the three phases of
\begin{enumerate*}[label=(\roman*)]
    \item training (gradient descent)
    \item sampling, and
    \item hyperparameters search.
\end{enumerate*}
For each dataset, each fold and each optimized model architecture, we ran the training and sampling phases on the exact same hardware and software architecture, a single Tesla V100 32 GB, and we measured accurately the cost values with the \textit{CodeCarbon} library\footnote{\label{fn:codecarbon}https://codecarbon.io/}.
However, due to the massive nature of the experiments, we could not perform the whole hyperparameter search and training phases on such a uniform hardware and software architecture.
We hence estimated the global search costs from the tuning logs by rescaling the training cost measures according to the effective number of steps performed and the number of GPU used (c.f. equation~\eqref{eq:costest}).
\begin{equation}
    \text{total-gpu-cost} \simeq{} \sum\limits_{ t \in \text{trials}} \frac{\text{init-cost}_t + \text{avg-cost-per-step}\times\text{num-steps}_t}{\text{trials-per-gpu}}
    \label{eq:costest}
\end{equation}
This is slightly overestimated since:
\begin{enumerate*}[label=(\roman*)]
    \item \textit{CodeCarbon} assumes a $100\%$ GPUs load while ours was roughly around $95\%$; and
    \item we measured the step costs on already optimized models which are usually slower because they often count more layers and parameters.
\end{enumerate*}
Note that the number of trials that can be parallelized on a single GPU depends on the memory footprint of the model\footnote{For instance we could safely run $10$ TVAE trials on the same V100 GPU while only $4$ TabSyn's trials could fit because of the VAE transformer's   footprint.}.

\subsection{Large Scale Optimization Framework and Implementation Details}
\label{sec:framework}

\begin{figure}
    \begin{center}
        \includegraphics[width=0.7\textwidth]{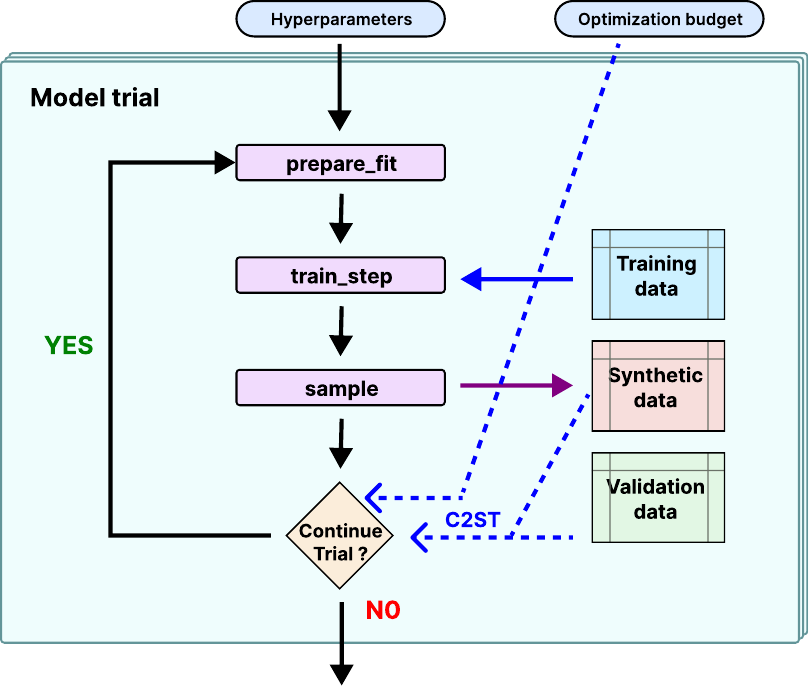}
        \caption{Hyperparameters trial optimization loop.}
        \label{fig:trialloop}
    \end{center}
\end{figure}

Providing a fair and reliable comparison of the different tabular generative models is a tough technical challenge. For this reason most existing benchmarks like \cite{great23, zhang2023mixed} only report the performance of models with their default hyperparameters. A few benchmarks like \cite{kotelnikov2023tabddpm, du2024towards} perform hyperparameters search for all models, but with a simple train/validation/test split, a reduced search space, and a small number of trials (usually from 20 to 50).

We wanted our experiment to be more extensive and more robust, so we decided to deploy it at a large scale on a super-computer\footnote{\href{http://www.idris.fr/eng/jean-zay/jean-zay-presentation-eng.html}{Jean-Zay super-computer \texttt{www.idris.fr}}} equipped with several nodes with 4 GPUs V100 32GB each\footnote{Complementary experiments were also performed on a $2\times$RTX4090 workstation.}.

For this purpose, we used the \textit{Ray tune} distributed library \cite{liaw2018tune} coupled with \textit{Hyperopt} \cite{bergstra2013making} which is based on Tree Parzen Estimators (TPE)~\cite{NIPS2011_86e8f7ab} and Asynchronous Successive Halving Algorithm (ASHA) as a scheduler 
\cite{li2018massively} to optimize hyperparameters and architecture efficiently. Depending on the model's memory footprints, each GPU could host from four to twelve concurrent tuning trials.
For each dataset and each model we performed a strict 3-fold cross validation, which means that for each tuple (dataset, fold, model) we performed an extensive hyperparameters search with 300 trials (except for TabSyn where we reduced this number to 100 for technical reasons explained in Section~\ref{sec:tabsyn_hp_tuning}). We hence obtained a different optimized architecture for each (dataset, fold, model) tuple.

In \cite{kotelnikov2023tabddpm} the parameters were optimized for ML-efficacy, in \cite{du2024towards} the parameters were  optimized for an equal combination of realism, utility and privacy. We chose to optimize for realism as in \cite{c2st22} through XGBoost-based C2ST metric (c.f. Section~\ref{s:metrics}). To obtain reliable evaluations with variance estimates, for each dataset we evaluated the models by averaging all the metrics on the three folds test sets with five synthetic samples for each fold.

As described in Figure~\ref{fig:trialloop}, in order to work with our framework each algorithm has to be wrapped into a generic \textit{Synthesizer} class that provides three methods: \textit{prepare\_fit} which prepares the dataset and the model according to the hyperparameters, \textit{train\_step} which performs a training step roughly equivalent to one or a few epochs, and \textit{sample} which generates synthetic data. After each train step, the model trial was evaluated and it was canceled out by early stopping or by the Ray-Tune scheduler if it performed too poorly or if the time budget was depleted.

\subsubsection{The importance of feature encoding}
\label{sec:feat_encoders}

Table~\ref{table:encoding_schemes} presents the encoding schemes that we used for the different benchmark challengers.
As pointed out in \cite{grinsztajn2022why, zhao2021ctab, c2st22}, categorical variables are not the main weakness of neural networks but numerical feature encoding is critical.
Most recent neural models use a Quantile-based numerical feature encoding and seem to work well with it. However, the original versions of TVAE and CTGAN rely on a specific Cluster-Based normalization \cite{CTGAN}.
We hence explored several encoding policies for these two models through hyperparameters optimization (see Tables \ref{table:hp_tvae} and \ref{table:hp_ctgan} in the appendix section). 

\begin{table}[h!]
    \centering
   
    \begin{tabular}{c|c|c|c}
        \toprule
        Model & Num. Encoder & Cat. Encoder & Num. Target\\
        \hline
        TVAE\_base & Cluster-Based \tablefootnote{\label{fn:cluster_normalizer}https://docs.sdv.dev/rdt/transformers-glossary/numerical/clusterbasednormalizer} & One hot & - \\ \hline
        CTGAN\_base & Cluster-Based\footref{fn:cluster_normalizer} & One hot & - \\ \hline
        TVAE & \textit{\color{blue}Optimized} & One hot & - \\ \hline        
        CTGAN & \textit{\color{blue}Optimized} & One hot & - \\ \hline
        TabDDPM & Quantile & One Hot& Standardize \\ \hline
        TabSyn & Quantile & Embedding & - \\ \hline
        SMOTE & - & One hot & Median cut \\ \hline
        ucSMOTE & - & One hot & Dummy \\ 
        \bottomrule
    \end{tabular}
    \caption{Encoding schemes applied to TVAE and CTGAN.}
    \label{table:encoding_schemes}
\end{table}

We kept the native Cluster-Based\footref{fn:cluster_normalizer} encoder as described in \cite{CTGAN}, along with the ones proposed in~\cite{c2st22}, namely:
\begin{itemize*}[label={}]
\item prototype encoding (PTP)~\cite{c2st22},
\item piece-wise linear encoder (PLE) \cite{gorishniy2022embeddings},
\item continuously distributed residuals (CDF)~\cite{dunn1996randomized, c2st22},
\item hybrid (PLE\_CDF)~\cite{c2st22}.
\end{itemize*}
We also added some standard \textit{scikit-learn} transformers: MinMaxScaler and QuantileTransformer.

Contrary to QuantileTransformer which maps values 
deterministically, the CDF encoding uses randomization to produce continuously distributed residuals even when the original distribution is discrete or partially discrete \cite{dunn1996randomized}.
The PLE encoder \cite{gorishniy2022embeddings} performs a feature binning and normalizes each numerical value depending on the bin it belongs to. The PLE\_CDF applies a CDF to the output of a PLE encoding. The PTP encoding, inspired by prototypical networks~\cite{prototype_networks}, encodes the input as a weighted average of fixed prototypes.

\subsubsection{Model-specific implementation details}
\label{sec:tabsyn_hp_tuning}

Wrapping heterogeneous models within a \textit{Synthesizer} class required some implementation choices from our part, and despite all our efforts to keep a fair comparison, these choices had some impact on the compute time and performance of the models.

The first issue is the discrepancy of the training step's costs.
The usual training step unit for most models is the \textit{epoch} which corresponds to a single pass on all instances of the training set.
However, depending on the hyperparameters some models like CTGAN perform both one pass through the generator and several passes through the discriminator at each training step. Other models such as TabDDPM are randomized and require several quick passes (almost one for each level of noise) for each instance.
We hence had to caliber our wrapper's train\_step functions to perform a compute effort that is roughly equivalent to an epoch.

Another issue is the fact that TabSyn \cite{zhang2023mixed} combines two models and was not designed for hyperparameters tuning.
According to the authors the model does not need hyperparameters tuning\footnote{Our experiment show however (c.f. Section~\ref{sec:results}), that  even if the non-optimized TabSyn is quite good on most datasets, the  hyperparameter tuning 
clearly improves the quality of the data it generates.}.
We decided to train a new transformer-based VAE for each trial because it hosts most of the parameters, compute-time and hyperparameters of TabSyn. Training a diffusion model on an unstable latent space would not be meaningful. We thus considered three technical options:
\begin{enumerate*}[label=(\roman*)]
    \item optimize first the VAE on a proxy metric (for instance the ability of its decoder to generate realistic data from a standard Gaussian), then optimize the denoiser in the latent space;
    \item wrap the VAE training steps into the \textit{train\_step} function and retrain a new denoiser from scratch at each step;
    \item wrap the VAE training phase into the \textit{prepare\_fit} function and loose the ability to prune its training steps.
\end{enumerate*}
The first option is the cheapest and it is probably recommended for most practical applications, but it may be sub-optimal due to the proxy metric. The second option is extremely costly.
We hence opted for the third option although it had a non-negligible cost. Indeed, we followed the recommendation of  \cite{zhang2023mixed} to train the VAE through $4000$ epochs which turns out to be huge knowing that most other models only utilized $400$ epochs in our benchmark.

We also reduced the number of parallel trials per GPU because of the large memory footprint of the VAE's transformers. As a consequence, for TabSyn we had both to reduce the number of trials to 100, and to work on a sample of the largest dataset (\textit{Covertype}) to get the results in a {reasonable} amount of time. It is worth noting that despite these handicaps, the optimized TabSyn remained better than its non-optimized version. To avoid this experimental bias for the second experiment in Section~\ref{sec:reducedtest}, we constrained TabSyn's VAE to use only $10$ minutes for training.
On \textit{Adult} dataset, for instance, it resulted in roughly $560$ epochs.

Finally, the last issue was the different ways the models deal with the target columns. A model conditioned on the target column may improve its ML-efficacy.
TabDDPM and SMOTE implementations are natively conditioned on classification targets, while TVAE, CTGAN and TabSyn are not.

We did not modify TabDDPM, but we considered two variants in our experiment for SMOTE: the first, that we call SMOTE, follows the design of \cite{kotelnikov2023tabddpm}, it uses the train target distribution for classification datasets and a rough median split for regression targets (see Table~\ref{table:encoding_schemes}).
The second, that we call ucSMOTE (for unconditional SMOTE), adds a dummy target filled with zeros to the data before calling the SMOTE oversampling library. By doing so, all columns, including the original target, are considered equally.

\subsection{Datasets}
\label{s:data}
To evaluate the models, we picked datasets with various characteristics to assess their performances  under different scenarios. 
Datasets were chosen to cover various sizes, dimensions, different types of tasks  (regression, binary, and multi-class classification) and various types of  features (numerical, categorical, or mixed).
We also added \textit{Moons} a well known \textit{scikitlearn} synthetic dataset.
The complete list of datasets and their characteristics is presented in Table \ref{tab:datasets}. The \textit{Covertype} dataset size was reduced for TabSyn tuning as follows: 27500 in the training set, 18333 in the validation set, and 9167 in the test set.

\begin{table}[th]
\small
    \centering
    \begin{tabular}{ccccccc}\hline
        Name & Train & Validation & Test & Num & Categ. & Task \\\hline
        Abalone\tablefootnote{\label{fn:dataset_link_openml}\url{https://www.openml.org}} & 2088 & 696&1393&7&2&Binclass\\
        Adult\footref{fn:dataset_link_openml} & 24420 & 8141 & 16281 & 6 &9 &Binclass\\
        Bank Marketing\footref{fn:dataset_link_openml} & 22605 & 7535& 15071& 7& 10& Regression\\
        Black Friday\footref{fn:dataset_link_openml} &83410&27804&55607&6&4&Regression\\
        Bike Sharing\footref{fn:dataset_link_openml}&8689&2897&5793&9&4&Regression\\
        Covertype\footref{fn:dataset_link_openml}& 290505 & 96836&193671&10&45&Multiclass\\
        Cardio\tablefootnote{\label{fn:dataset_link_kaggle}\url{https://www.kaggle.com/datasets}}& 34999& 11667& 23334& 11& 1& Binclass\\
        Churn Modelling\footref{fn:dataset_link_kaggle}& 4999& 1667& 3334& 8& 4& Binclass\\
        Diamonds\footref{fn:dataset_link_openml}& 26970& 8990& 17980& 7& 3& Regression\\
        HELOC\footref{fn:dataset_link_kaggle}& 5229 & 1743&3487&23&1&Binclass\\
        Higgs\footref{fn:dataset_link_openml}& 49024& 16342& 32684& 28& 1& Binclass\\
        House 16H\footref{fn:dataset_link_openml}& 11391& 3798& 7595& 17& 0& Regression\\
        Insurance\footref{fn:dataset_link_kaggle}& 669& 223& 446& 4& 3& Regression\\
        King\footref{fn:dataset_link_kaggle}& 10806& 3602& 7205& 19& 1& Regression\\
        MiniBooNE\footref{fn:dataset_link_openml}& 65031& 21678& 43355& 50& 1& Binclass\\
        Two Moons&19999 &6667&13334&2&1&Binclass\\\hline
    \end{tabular}
    \caption{List of datasets. Direct links to exact versions of datasets used can be found in \ref{appendix:datasets_links}.}
    \label{tab:datasets}
\end{table}

\section{Extensive Benchmark}
\label{sec:results}

\begin{figure}[ht!]
    \centering
    \includegraphics[width=1\textwidth]{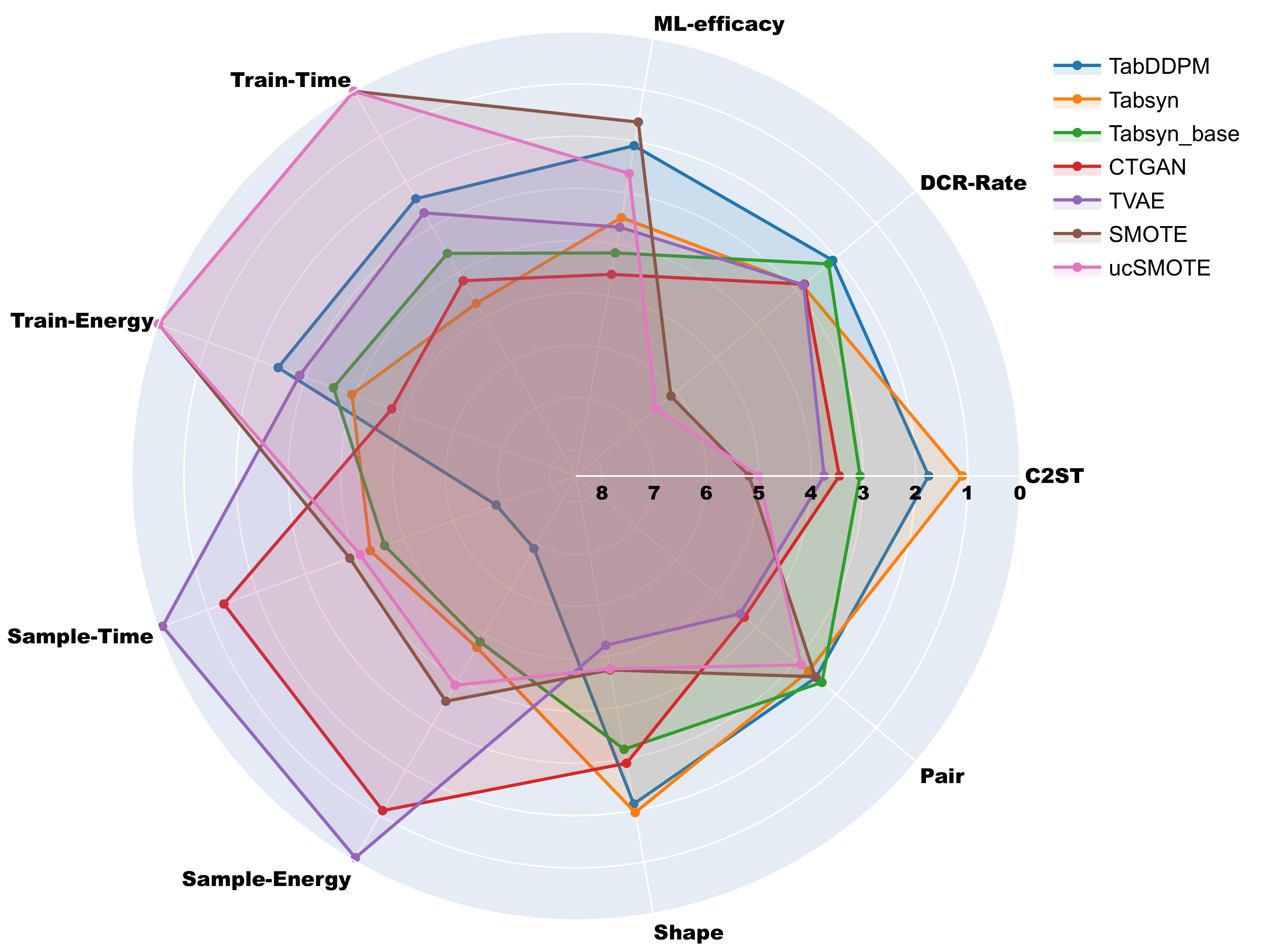}
    \caption{Radar chart of the \textbf{extensive experiment} showing optimized model's average ranking on all datasets across various metrics. The training costs of optimized models include both tuning and gradient descent.}
    \label{fig:radar_rank_all}
\end{figure}

As mentioned in Section~\ref{s:metrics}, 
the evaluation and comparison of tabular generative models is based mainly on four criteria: realism, usefulness, anonymity, and cost. After a global multi-criteria overview of the results, we study and compare the model's behaviour according to each criterion  individually.

\subsection{Multi-criteria Overview} 
\label{sec:result_overview}

Figure \ref{fig:radar_rank_all} shows the average ranking over all datasets and folds of seven model variants over eight metrics (c.f. Section~\ref{s:metrics}).
To complete these rankings, Table~\ref{table:quartiles_extensive_baselines} and Table~\ref{table:quartiles_costs} summarize respectively the quality metric and cost distributions among all datasets and folds.
Recall that the best scores for the C2ST are around $0.50$ (as it means poor AUC for the classifier at telling synthetic data apart from holdout data).

Overall, no model provides the best performance over all considered criteria.
We observe, as expected, a strong correlation between energy and GPU time as well as a strong correlation among ``quality metrics'' (\textit{i.e.} C2ST, \textit{Shape}, and \textit{Pair}).
On the one hand, the diffusion models achieve the best performance in terms of quality metrics, especially the tuned version of TabSyn. As shown in Table~\ref{table:quartiles_extensive_baselines},  this model has a median C2ST value of $0.64$. However, TabSyn is also one of the most expensive models in term of training costs  (\textit{Train-Energy} as well as \textit{Train and Sample Times}), as it requires to train both a transformer-based VAE and a denoiser model for each dataset.

On the other hand, the SMOTE baselines obtain the poorest quality and privacy ranking with a high DCR-Rate. The median DCR-Rate score for SMOTE and ucSMOTE is at $0.97$ which means that most of the samples from these models are very similar to the training set. However, they achieve strong utility in terms of \textit{ML-Efficacy} with quartiles very close to \textit{Train~Copy} and the same median value of $0.73$. As expected for neighborhood-based algorithms, their training cost is negligible, but their deployment requires a neighborhood search which can be costly on large datasets.
Finally the tuned neural push-forward models (CTGAN and TVAE) achieve mitigated results in term of both quality and utility but their deployment is clearly the cheapest.
We also note that all neural models achieve reasonable results in term of privacy preservation. Although TabDDPM is clearly the slowest algorithm for deployment, it is fast at training and it obtains homogeneous results over all other metrics. We note that the tuned version of TabDDPM is performing better than the base version of TabSyn.

For this extensive experiment we only limited mildly the time budget and the number of epochs. However, some models like TabSyn and CTGAN consumed much more GPU time than others, especially TabDDPM which was very quick at performing an equivalent number of epochs (c.f. Section~\ref{sec:tabsyn_hp_tuning}).
It is also important to compare these algorithms with a fair allocation of GPU resource as we do in Section~\ref{sec:hpsens}.

\begin{table}[htbp]
    \centering

    \resizebox{0.8\textwidth}{!}{
    
    \begin{tabular}{|c|c|c|c|c|c|c|}
        \hline
        Model & \scriptsize Percentiles & C2ST $\downarrow$ & DCR-R $\downarrow$ & ML-EF $\uparrow$ & Shape $\uparrow$ & Pair $\uparrow$ \\
        \hline
        \multirow{4}{*} {Train~Copy} & 75\% & 0.50 & 1.00 & 0.90 & 0.99 & 0.99 \\
\cline{2-7} 
 & 50\% & 0.50 & 1.00 & 0.73 & 0.99 & 0.98 \\
\cline{2-7} 
 & 25\% & 0.50 & 1.00 & 0.63 & 0.99 & 0.91 \\
\specialrule{1.5pt}{1pt}{1pt} 
\multirow{4}{*} {TVAE} & 75\% & 0.88 & 0.65 & 0.79 & 0.96 & 0.95 \\
\cline{2-7} 
 & 50\% & 0.81 & 0.63 & 0.71 & 0.95 & 0.92 \\
\cline{2-7} 
 & 25\% & 0.74 & 0.61 & 0.50 & 0.94 & 0.85 \\
\specialrule{1.5pt}{1pt}{1pt}

\multirow{4}{*} {TVAE\_base} & 75\% & 1.00 & 0.62 & 0.70 & 0.92 & 0.90 \\
\cline{2-7} 
 & 50\% & 0.98 & 0.61 & 0.65 & 0.91 & 0.80 \\
\cline{2-7} 
 & 25\% & 0.96 & 0.60 & 0.44 & 0.87 & 0.74 \\
\specialrule{1.5pt}{1pt}{1pt} 

\multirow{4}{*} {CTGAN} & 75\% & 0.90 & 0.64 & 0.71 & 0.98 & 0.96 \\
\cline{2-7} 
 & 50\% & 0.83 & 0.63 & 0.69 & 0.97 & 0.91 \\
\cline{2-7} 
 & 25\% & 0.71 & 0.61 & 0.47 & 0.96 & 0.85 \\
\specialrule{1.5pt}{1pt}{1pt} 

 \multirow{4}{*} {CTGAN\_base} & 75\% & 1.00 & 0.62 & 0.63 & 0.92 & 0.93 \\
\cline{2-7} 
 & 50\% & 0.99 & 0.60 & 0.47 & 0.88 & 0.83 \\
\cline{2-7} 
 & 25\% & 0.93 & 0.60 & 0.30 & 0.87 & 0.76 \\

\specialrule{1.5pt}{1pt}{1pt} 
\multirow{4}{*} {TabDDPM} & 75\% & 0.78 & 0.64 & 0.80 & 0.98 & 0.98 \\
\cline{2-7} 
 & 50\% & 0.67 & 0.62 & 0.69 & 0.98 & 0.93 \\
\cline{2-7} 
 & 25\% & 0.63 & 0.61 & 0.58 & 0.95 & 0.84 \\
\specialrule{1.5pt}{1pt}{1pt} 

\multirow{4}{*} {TabDDPM\_base} & 75\% & 0.86 & 0.64 & 0.75 & 0.99 & 0.96 \\
\cline{2-7} 
 & 50\% & 0.77 & 0.62 & 0.68 & 0.98 & 0.92 \\
\cline{2-7} 
 & 25\% & 0.65 & 0.61 & 0.44 & 0.94 & 0.74 \\
 \specialrule{1.5pt}{1pt}{1pt} 

\multirow{4}{*} {TabSyn} & 75\% & 0.80 & 0.63 & 0.76 & 0.99 & 0.97 \\
\cline{2-7} 
 & 50\% & 0.64 & 0.62 & 0.68 & 0.97 & 0.93 \\
\cline{2-7} 
 & 25\% & 0.59 & 0.61 & 0.46 & 0.96 & 0.74 \\
\specialrule{1.5pt}{1pt}{1pt} 
\multirow{4}{*} {TabSyn\_base} & 75\% & 0.86 & 0.64 & 0.75 & 0.98 & 0.97 \\
\cline{2-7} 
 & 50\% & 0.71 & 0.62 & 0.57 & 0.97 & 0.95 \\
\cline{2-7} 
 & 25\% & 0.63 & 0.61 & 0.29 & 0.95 & 0.88 \\
\specialrule{1.5pt}{1pt}{1pt} 
\multirow{4}{*} {SMOTE} & 75\% & 0.97 & 0.98 & 0.86 & 0.97 & 0.99 \\
\cline{2-7} 
 & 50\% & 0.90 & 0.97 & 0.73 & 0.95 & 0.95 \\
\cline{2-7} 
 & 25\% & 0.80 & 0.86 & 0.60 & 0.93 & 0.85 \\
\specialrule{1.5pt}{1pt}{1pt} 
\multirow{4}{*} {UC-SMOTE} & 75\% & 0.95 & 0.98 & 0.87 & 0.97 & 0.98 \\
\cline{2-7} 
 & 50\% & 0.88 & 0.97 & 0.73 & 0.95 & 0.95 \\
\cline{2-7} 
 & 25\% & 0.81 & 0.91 & 0.59 & 0.93 & 0.87 \\
\hline 

    \end{tabular}

    } 
    \caption{Summary results from the \textit{\textbf{extensive}} benchmark and the \textit{\textbf{base}} models (i.e. with default hyperparameters), DCR-R stands for DCR-rate and  ML-EF for ML-Efficacy.}
    \label{table:quartiles_extensive_baselines}
\end{table}

\subsection{Detailed Analysis} 
\label{sec:1b1analysis}

In this section we study and compare the model's behaviour according to each criterion taken individually.
For each dataset we computed the performance metrics over 3 folds and 5 synthetic samples per fold to provide a stable central tendency and dispersion estimate. The full dataset-level results are provided in Appendix Table~\ref{table:results}. We summarize these results among all datasets and folds in Table~\ref{table:quartiles_extensive_baselines} and Table~\ref{table:quartiles_costs}.
For quality metrics we compare the models through critical difference diagrams.

\subsubsection{Are synthetic data realistic ?} 
\label{sec:realism_result}

\begin{figure}[t!]
\begin{center}
    \subfigure[C2ST]{\includegraphics[width=0.9\textwidth]{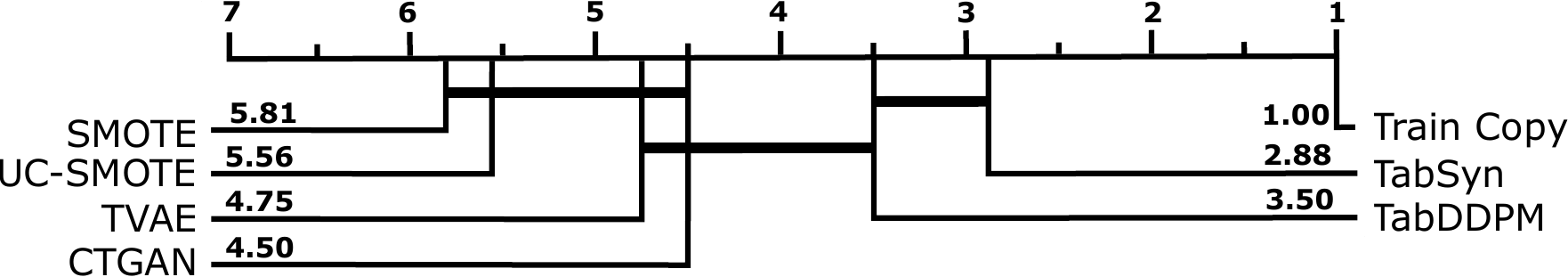} \label{fig:cdd_c2st}
    }

    \subfigure[Column-wise shape similarity]{\includegraphics[width=0.9\textwidth]{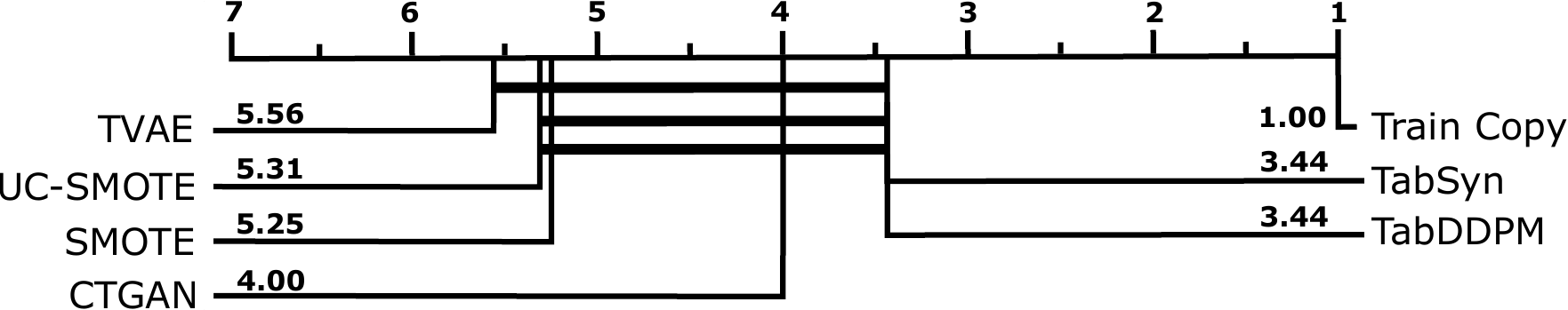}
    \label{fig:cdd_shape}
    }

    \subfigure[Pair-wise correlation]{\includegraphics[width=0.9 
 \textwidth]{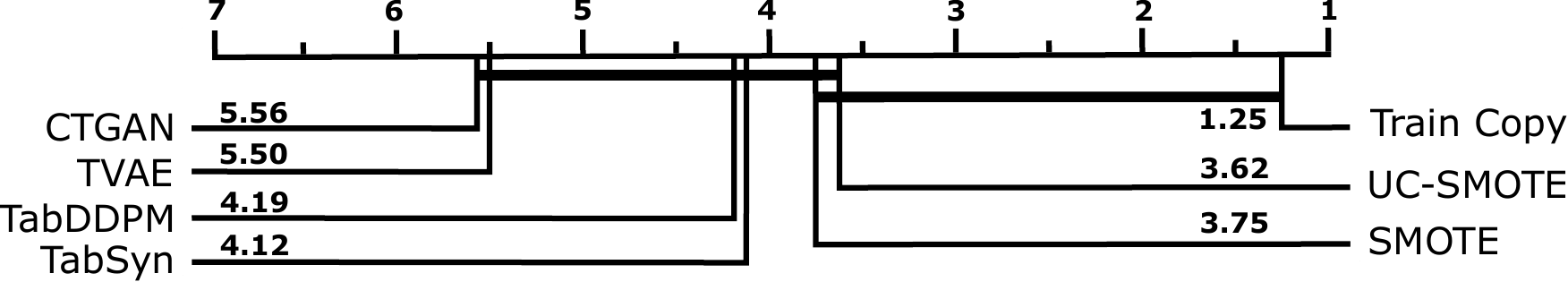}
    \label{fig:cdd_pair}
    }

    \caption{Models ranking with critical difference diagrams for C2ST, pair-wise correlations, and column-wise metrics over all datasets.}
    \label{fig:cdd}
\end{center}    
\end{figure}

We show in Figure \ref{fig:cdd} the \textit{critical difference diagrams} \cite{demvsar2006statistical} of all tuned models respectively for C2ST, pair-wise correlation and column-wise similarity. These diagrams were obtained by aggregating the ranks of the seven models over all datasets and folds. A thick horizontal line groups the set of models for which the pairwise ``\textit{no significant difference}" test hypothesis could not be rejected.

If we except the trivial \textit{Train~Copy} policy which is by construction the most realistic generator, we note that TabSyn is significantly better in terms of C2ST than all other models except TabDDPM. On the other side, the SMOTE baselines are significantly worse than TabDDPM and TabSyn.
The absolute C2ST values in Table~\ref{table:quartiles_extensive_baselines} corroborate these ranking results with three \textit{sets} of models:
\begin{enumerate*}[label=(\roman*)]
\item diffusion-based models TabSyn and TabDDPM with a C2ST lower than $0.80$ on most datasets and median around  $0.65$;    
\item push-forward models CTGAN and TVAE with a median C2ST around $0.82$;
\item  SMOTE algorithms with a C2ST higher than $0.80$ on most datasets.
\end{enumerate*}

The rankings obtained according to the \textit{column-wise similarity} are broadly the same as the one obtained for C2ST. In addition to the rank shown in Figure \ref{fig:cdd_shape}, we can see that all the models obtain good scores (usually around $0.96$) as compared to the \textit{Train~Copy} baseline ($0.99$). This result suggests that all the models succeed at capturing univariate distributions. 
For \textit{pair-wise correlation}, we note surprisingly high values for SMOTE and ucSMOTE baselines while the neural network models obtain broadly the same ranking as for C2ST but the gaps between models are less marked.

A side takeaway from this result is that XGBoost-based C2ST provides a stronger discriminative power than \textit{column-wise similarity} and \textit{pair-wise correlation} metrics.

\subsubsection{Can the synthetic data be used to train a machine learning model ?}

\begin{figure}[htbp]
    \centering
    \includegraphics[width=0.9\textwidth]{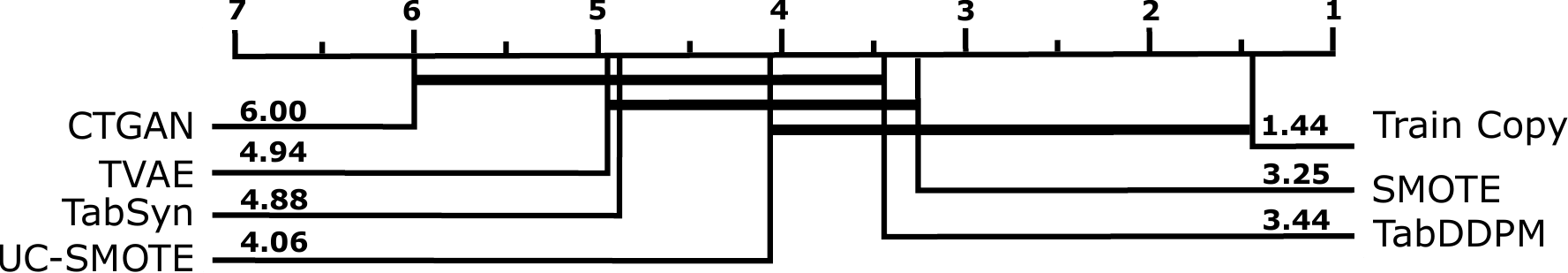}
    \caption{Models ranking with critical difference diagram for Catboost ML-Efficacy over all datasets.}
    \label{fig:mle_cdd}
\end{figure}

According to the \textit{machine learning efficacy} metric (ML-Efficacy), the most useful generators are the ones that are conditioned on their targets (namely SMOTE and TabDDPM) with median values respectively at $0.73$ and $0.69$ (against $0.73$ for \textit{Train~Copy}).

TabDDPM outperforms both base and tuned versions of TabSyn for this metric. It is also safer than SMOTE and it performs its training iterations faster than the other evaluated models.
If ML-Efficacy is of importance, it is advisable to use this model. 
As expected, all models are far from the performance obtained on real data (\textit{Train~Copy}).

\subsubsection{Does synthetic data preserve anonymity?}
\label{sec:privacy}

\begin{figure}[ht]
    \centering
    \includegraphics[width=0.9\textwidth]{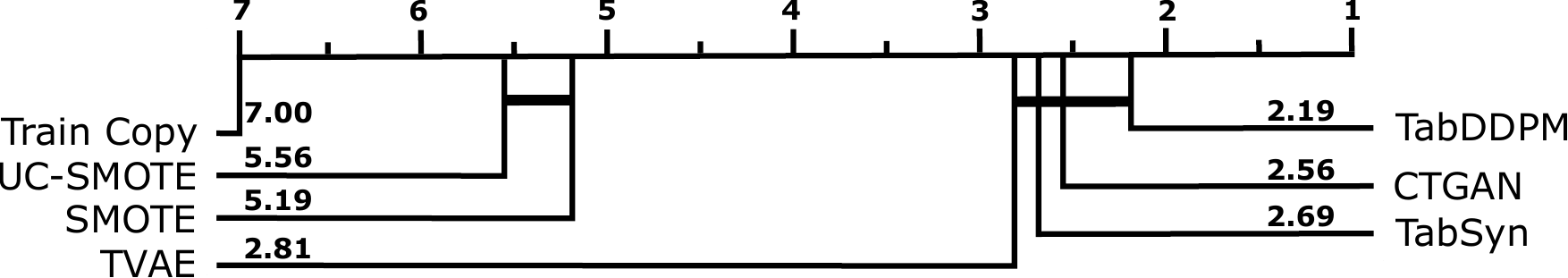}
    \caption{Models ranking with a critical difference diagram for the DCR-Rate metric over all datasets.}
    \label{fig:cdd_dcr}
\end{figure}

A data synthesizer that would only copy its training set would be of little value. If it generates new instances that are too close to its training set, it would obtain a good C2ST score, but it would be prone to over-fitting and it would leak private information from the training set.
We assess the ability of a model to generate new data through the DCR-Rate metric (c.f. Section~\ref{s:metrics}).

On Figure \ref{fig:cdd_dcr} we observe two significantly distinct {groups} of models. On the left-hand side a "leaky" group that contains both SMOTE and ucSMOTE, and on the right-hand side, a "safe" group that contains all neural algorithms. 
The poor performance of SMOTE is mainly due to the way it generates new data points by interpolating between existing ones. Therefore, these models cannot be considered safe concerning data protection.
By taking a look at Table~\ref{table:quartiles_extensive_baselines}, the DCR-Rate of the two SMOTE variants is almost always above $0.86$. On the other hand, if we exclude the tiny "Insurance" dataset where CTGAN and TVAE overfitted, the DCR-Rate values of the "safe" group are quite uniform around $0.62$ and almost always below $0.65$: these models can be considered safe.

\subsubsection{What are the models' costs?}
\label{sec:models_cost}

 \begin{table}[htbp]
    \centering

    \resizebox{0.75\textwidth}{!}{
    
    \begin{tabular}{|c|c|c|c|c|c|c|}
        \hline
        Model &  \scriptsize Percentiles & \makecell{Duration \\ (HH:MM)} $\downarrow$ & \makecell{Emission \\ (Kg)} $\downarrow$ & \makecell{Energy \\ (kWh)} $\downarrow$ \\
        \hline
        \multirow{4}{*} {TVAE} & 75\% & 01:00 & 1.05 & 15.57  \\
\cline{2-5} 
 & 50\% & 00:24 & 0.42 & 6.22  \\
\cline{2-5} 
 & 25\% & 00:12 & 0.21 & 3.08  \\
\specialrule{1.5pt}{1pt}{1pt} 
\multirow{4}{*} {CTGAN} & 75\% & 02:20 & 2.45 & 36.36  \\
\cline{2-5} 
 & 50\% & 01:37 & 1.67 & 24.84  \\
\cline{2-5} 
 & 25\% & 01:06 & 1.16 & 17.14  \\
\specialrule{1.5pt}{1pt}{1pt} 
\multirow{4}{*} {TabDDPM} & 75\% & 00:26 & 0.36 & 5.30  \\
\cline{2-5} 
 & 50\% & 00:18 & 0.29 & 4.37  \\
\cline{2-5} 
 & 25\% & 00:12 & 0.19 & 2.82  \\
\specialrule{1.5pt}{1pt}{1pt} 
\multirow{4}{*} {TabSyn} & 75\% & 03:25 & 1.92 & 22.65  \\
\cline{2-5} 
 & 50\% & 01:59 & 1.03 & 14.20  \\
\cline{2-5} 
 & 25\% & 01:29 & 0.72 & 9.09  \\
\specialrule{1.5pt}{1pt}{1pt} 
\multirow{4}{*} {SMOTE} & 75\% & 00:02 & 0.00 & 0.01  \\
\cline{2-5} 
 & 50\% & 00:00 & 0.00 & 0.00  \\
\cline{2-5} 
 & 25\% & 00:00 & 0.00 & 0.00  \\
\specialrule{1.5pt}{1pt}{1pt} 
\multirow{4}{*} {UC-SMOTE} & 75\% & 00:03 & 0.00 & 0.01  \\
\cline{2-5} 
 & 50\% & 00:00 & 0.00 & 0.00  \\
\cline{2-5} 
 & 25\% & 00:00 & 0.00 & 0.00  \\
\hline 

    \end{tabular}

    } 
    \caption{Summary of the \textit{costs} of the \textit{\textbf{extensive}} benchmark. \textit{Costs} dispersion of the 25th, 50th, and 75th percentiles are provided over all datasets and folds. }
    \label{table:quartiles_costs}
\end{table}

Model training, optimization, and data sampling have a cost and an environmental impact that varies greatly from one model to another.
We hence measured and estimated time, energy consumption, and CO$_2$ impact for each model during three phases:
\begin{enumerate*}[label=(\roman*)]
    \item training (measured), 
    \item hyperparameters search (estimated), and
    \item sampling (measured).
\end{enumerate*}
These results are reported fully in \ref{appendix:tuning_cost}.
We summarize the tuning and training process cost distributions over all datasets and fold in Table~\ref{table:quartiles_costs}.
As mentioned in Section~\ref{s:metrics}, these values are estimated from the tuning logs by taking into account the effective number of training steps, the number of trials, and the average GPU resource usage per training step as reported in Table~\ref{table:emissions_estimation_normalized}.

 With a median tuning time around $18$ minutes as shown in Table~\ref{table:quartiles_costs}, TabDDPM is the fastest neural model. As a result, it also consumes less energy and it has less emissions at the training stage. 
As shown in Figure~\ref{fig:radar_rank_all}, considering its other performance metrics, it is a suitable choice to achieve {good} results at a relatively low training cost.

As expected, Figure~\ref{fig:radar_rank_all} shows that the push models CTGAN and TVAE are the fastest at sampling stage. We notice that although CTGAN achieves slightly better quality results than TVAE, it is also one of the costliest models at the training stage as shown in Table \ref{table:quartiles_costs}.
In order to achieve reasonable performance while reducing training costs, TVAE is hence an option to consider prior to CTGAN.

The training and tuning of TabSyn is one of the most demanding (with a median time of of $2$ hours for tuning). 
In the end, however, it delivers the best performance in terms of quality metrics. This model also has the advantage of providing a set of default hyperparameters that can have a \textit{reasonable} performance although, if we follow the author's recommendation of $4000$ VAE epochs, its  training remains costly by comparison to other models. Indeed, Figure~\ref{fig:radar_rank_all} show that, even if we consider the whole tuning+training pipeline TabDDPM and TVAE remain cheaper to train than TabSyn\_base.

In terms of sampling cost, TabDDPM is the worst-performing model. It takes longer than the other models (c.f. Figure~\ref{fig:radar_rank_all}) and hence consumes more energy with more CO$_2$ emissions at this step. TabSyn reduces the number of denoising steps by using VAE embedding and linear noises to reduce its sampling time~\cite{zhang2023mixed}. It hence  achieves better performance at inference than TabDDPM. 

The two baselines SMOTE and ucSMOTE being based on neighbourhood interpolation their training and tuning cost is negligible. However, as shown in Appendix Table~\ref{table:results}, their sampling process requiring a nearest neighbor search, it is slower on large datasets than push-forward models like TVAE or CTGAN.

\subsection{Is it worth optimizing the hyperparameters for all models ?}
\label{sec:tune_vs_base}

\begin{figure}[!htb]
     \centering

    \begin{adjustbox}{minipage=\linewidth,scale=0.88}

    \subfigure[C2ST]{\includegraphics[width=0.9\textwidth]{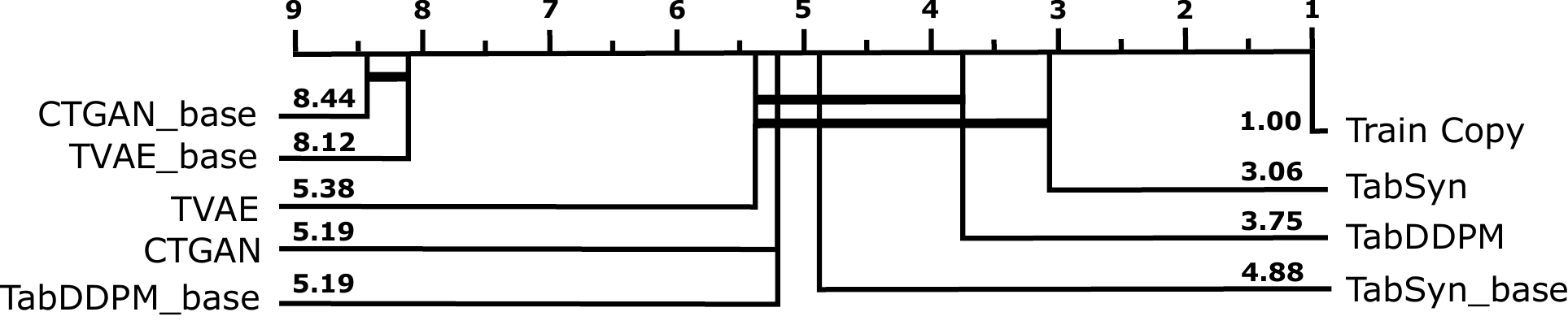}
    \label{fig:cdd_c2st_tune_vs_base}
    }

        \subfigure[DCR-Rate]{\includegraphics[width=0.9 
 \textwidth]{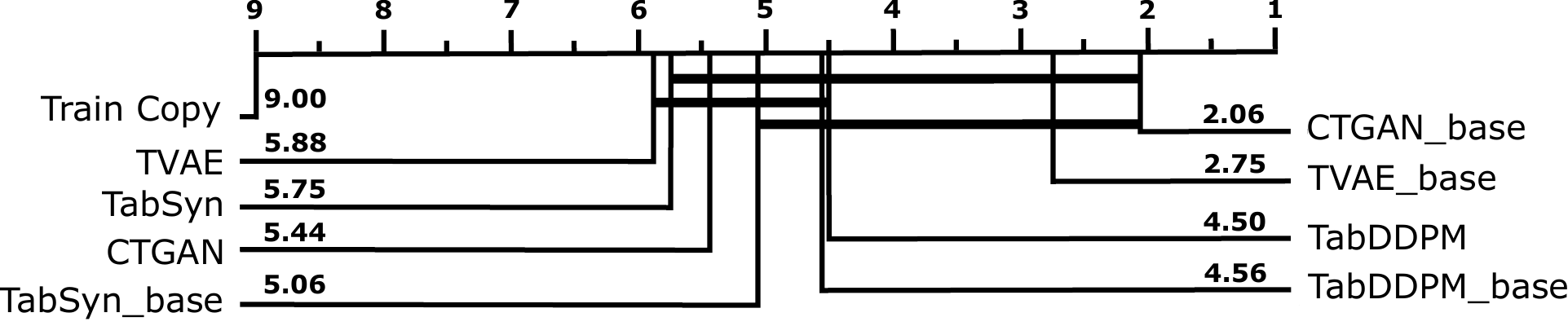}
    \label{fig:cdd_dcr_rate_tune_vs_base}
    }
    
        \subfigure[ML-Efficacy]{\includegraphics[width=0.9 
 \textwidth]{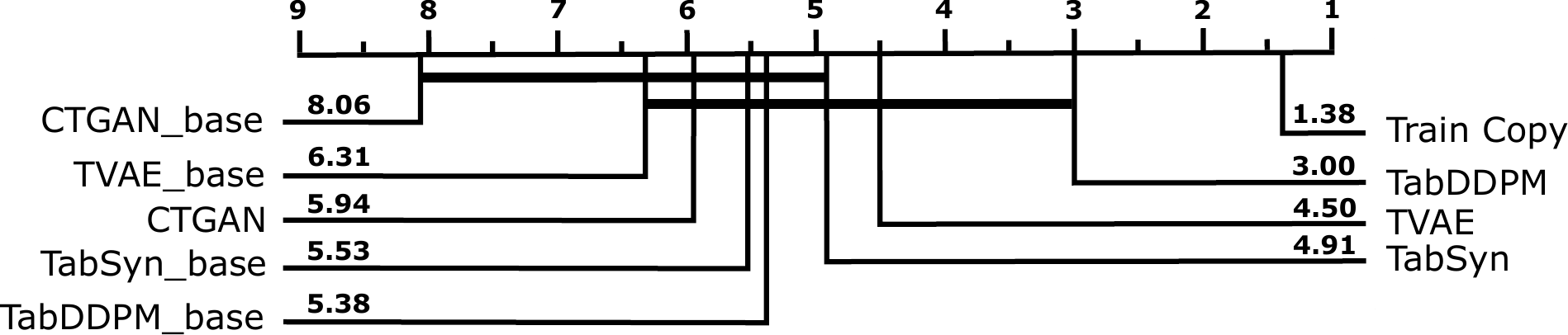}
    \label{fig:cdd_mle_tune_vs_base}
    }
    
    \end{adjustbox}

    \caption{Models' ranking with critical difference diagrams on C2ST, DCR-Rate, and ML-Efficacy metrics over all datasets: base models (i.e. with default hyperparameters) versus extensively tuned models.}
\end{figure}

As mentioned in the previous section, even with the help of sophisticated search algorithms, hyperparameter tuning is costly and the performance-versus-tuning-budget curve is following a \emph{diminishing returns law}.
We were hence interested in comparing heavily tuned models against non-tuned models.

To assess this, we trained the neural models using the hyperparameters provided by the authors in the original papers. These results are detailed in \ref{appendix:base_vs_tuned}.

In Figure~\ref{fig:cdd_c2st_tune_vs_base} we can see a, sometimes huge, C2ST performance improvement of all models when tuned.
This improvement is statistically significant between optimized TVAE and CTGAN and their base versions. Looking at the absolute C2ST values in Table~\ref{table:quartiles_extensive_baselines}, it confirms that these models should not be used with their default hyperparameters.

For TabDDPM and TabSyn, we also notice a 10 points improvement on the median C2ST but this gap is not large enough to provide statistical guarantees.

Although it was not the main target for tuning, we also observe an ML-Efficacy  gain for all models. This gain is marked at the $25^{th}$ percentile (\textit{i.e.} for the hardest datasets).

We can therefore conclude that for all neural tabular generative models that we considered (including TabSyn), it is worth optimizing the hyperparameters specifically for each dataset if we want to improve performance.
But the trade-off between the optimization cost and the performance gain is highly correlated to the size and design of the hyperparameter's space (c.f. \ref{appendix:hp_search_space}).
In the next section we propose
a reduced search space and study the impact of a "light" hyperparameters optimization with a limited budget.

\section{Limited-Budget Benchmark}
\label{sec:hpsens}

As underlined in Section~\ref{sec:tune_vs_base}, optimizing the hyperparameters for each dataset can significantly improve the quality of the generators. But a large-scale optimization like the one we performed is technically difficult, costly, and it has a non-negligible carbon impact (c.f. Table~\ref{table:quartiles_costs}).
Researchers and practitioners may hence be interested in reducing this cost without deteriorating too much the model's performance.

In this section, we leverage the results of our extensive tuning experiment to:
\begin{enumerate*}[label=(\roman*)]
    \item suggest reduced search spaces achieving reasonable performance at a much lower cost;
    \item assess and compare the performance of the models when tuned and trained with the same limited budget.
\end{enumerate*}
By comparing the model's performance after this limited-budget tuning/training with our previous results (respectively with heavy tuning or without tuning at all), we gain new insights into the models.

\subsection{Hyperparameters Search Space Reduction}

We carried out this experiment for the most hyperparameter-heavy models, namely: TVAE, CTGAN, TabDDPM, and TabSyn. To reduce their search space, we independently considered each  hyperparameter/architecture configuration variable and kept only the values that were the most frequently selected during the large-scale tuning phase. For discrete variables we kept the $80\%$ most frequent values, and for continuous variables we kept the value ranging between the 10th and 90th percentiles.

For instance, on CTGAN the large-scale tuning included six encoder options: CDF, PLE\_CDF, PTP, Quantile, MinMax, and CBN. But only CDF and PLE\_CDF were selected on most datasets. We could also drastically reduce TabSyn VAE's learning rate range from $(10^{-5},10^{-2})$ to $(10^{-3},7\cdot{}10^{-3})$. We present all these reduced search spaces in \ref{appendix:hp_search_space}.

\subsection{Putting the Reduced Search Spaces to the Test}
\label{sec:reducedtest}

To evaluate the reduced search spaces, we ran a new tuning on all datasets with only $50$ trials and a strict limit of $20$ minutes per trial. Since TabSyn's training is done in two steps, we allocated $10$ minutes for the VAE training and $10$ minutes for the denoiser. Each hyperparameter search was performed with the same 3-fold splits and the same methodology as in the extensive experiment.

\subsubsection{A Cost-Performance Trade-off}
\label{section:light_experiment_results}

\begin{figure}[ht!]
    \centering
    \includegraphics[width=1\textwidth]{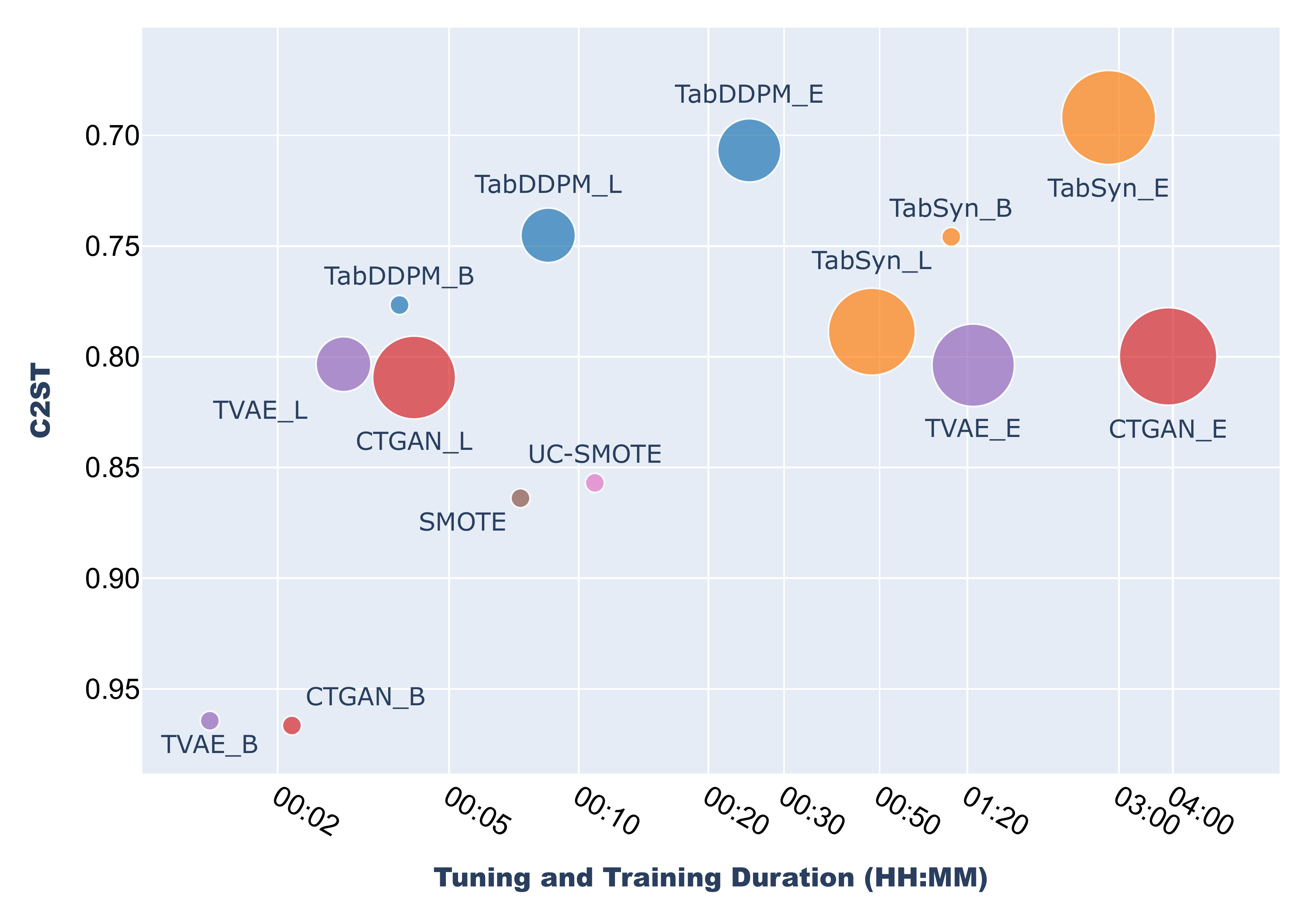}
    \caption{Average C2ST performance of the models under various setups (base models are appended with "\_B", limited-budget with "\_L" and extensive with "\_E"). The C2ST axis shows the best models at the top, the dot diameter indicates the complexity of the search space. The duration axis and dot diameters are log-scaled for better visualization.
    } 
    \label{fig:scatter_performance_comparison}
\end{figure}

In Figure \ref{fig:scatter_performance_comparison} we summarize the performance of the models after limited-budget hyperparameter search against the ones obtained with the base models and the ones obtained after the extensive tuning of Section~\ref{sec:framework}.
On the x-axis we display the total GPU-time (search+optimization) needed to obtain the corresponding model and  on the y-axis we show the C2ST performance. The diameters of the dots represents the number of possible configurations of the  hyperparameter search space as described in~\ref{appendix:hp_search_space}. 

If we compare against the non-tuned base models, the C2ST performance of all models except TabSyn is clearly improved by the limited-budget hyperparameter tuning.
For CTGAN, and TVAE the gain is huge: after only a few minutes of tuning with the reduced search space we reach the same performance as the one obtained after hours of heavy tuning.
For TabDDPM, we also improved the base performance by running the limited-budget hyperparameter tuning. However, we did not reach the same performance as we did with an extensive search.

In the limited-budget experiment, the TabSyn's performance degradation against the base model reveals 
the importance of a well-trained VAE  which leads to better samples from the diffusion process. Indeed, on Adult dataset for instance, within its $10$ minutes budget, it was trained with less than $600$ epochs against $4000$ for the base model.
This suggests to increase the number of epochs or the training time for TabSyn's VAE in further experiments.

Finally, as shown in
Figure~\ref{fig:scatter_performance_comparison}, the costs induced by the heavily optimized TabSyn model are high compared to the base one. The performance gain and the gap in cost should be considered depending on the task and constraints as the base model already delivers \textit{decent} performance.

It is worth noting that, due to its quick implementation, TabDDPM performed sometimes more training steps within its $20$ minutes time budget on the limited-budget experiment than it did on the extensive search where the number of steps was bounded. But on the other hand, its performance was also affected by the reduced number of trials ($50$ instead of $300$).

\subsubsection{Multi-criteria Analysis of the Limited-Budget Tuning}

  \begin{table}[htbp]
    \centering
    
    \resizebox{0.8\textwidth}{!}{
    
    \begin{tabular}{|c|c|c|c|c|c|c|}
        \hline
        Model &  \scriptsize Percentiles & C2ST $\downarrow$ & DCR-R $\downarrow$ & ML-EF $\uparrow$ & Shape $\uparrow$ & Pair $\uparrow$ \\
        \hline
        \multirow{4}{*} {Train Copy} & 75\% & 0.50 & 1.00 & 0.90 & 0.99 & 0.99 \\
\cline{2-7} 
 & 50\% & 0.50 & 1.00 & 0.73 & 0.99 & 0.98 \\
\cline{2-7} 
 & 25\% & 0.50 & 1.00 & 0.63 & 0.99 & 0.91 \\
\specialrule{1.5pt}{1pt}{1pt} 
\multirow{4}{*} {TVAE} & 75\% & 0.92 & 0.65 & 0.80 & 0.96 & 0.95 \\
\cline{2-7} 
 & 50\% & 0.80 & 0.63 & 0.70 & 0.95 & 0.93 \\
\cline{2-7} 
 & 25\% & 0.70 & 0.61 & 0.51 & 0.94 & 0.87 \\
\specialrule{1.5pt}{1pt}{1pt} 
\multirow{4}{*} {CTGAN} & 75\% & 0.89 & 0.64 & 0.71 & 0.98 & 0.96 \\
\cline{2-7} 
 & 50\% & 0.85 & 0.63 & 0.68 & 0.97 & 0.92 \\
\cline{2-7} 
 & 25\% & 0.70 & 0.62 & 0.45 & 0.95 & 0.85 \\
\specialrule{1.5pt}{1pt}{1pt} 
\multirow{4}{*} {TabDDPM} & 75\% & 0.83 & 0.67 & 0.83 & 0.97 & 0.95 \\
\cline{2-7} 
 & 50\% & 0.72 & 0.63 & 0.69 & 0.96 & 0.91 \\
\cline{2-7} 
 & 25\% & 0.63 & 0.62 & 0.50 & 0.95 & 0.81 \\
\specialrule{1.5pt}{1pt}{1pt} 
\multirow{4}{*} {TabSyn} & 75\% & 0.88 & 0.63 & 0.71 & 0.97 & 0.96 \\
\cline{2-7} 
 & 50\% & 0.81 & 0.62 & 0.65 & 0.95 & 0.93 \\
\cline{2-7} 
 & 25\% & 0.72 & 0.61 & 0.34 & 0.94 & 0.82 \\
\hline 

    \end{tabular}
}
    
    \caption{Performance dispersion across all datasets in the \textit{\textbf{limited-budget}} benchmark. DCR-R and ML-EF stand for DCR-Rate and ML-Efficacy respectively.}
    \label{table:quartiles_light}
\end{table}

\begin{figure}[ht!]
    \centering
    \includegraphics[width=.9\textwidth]{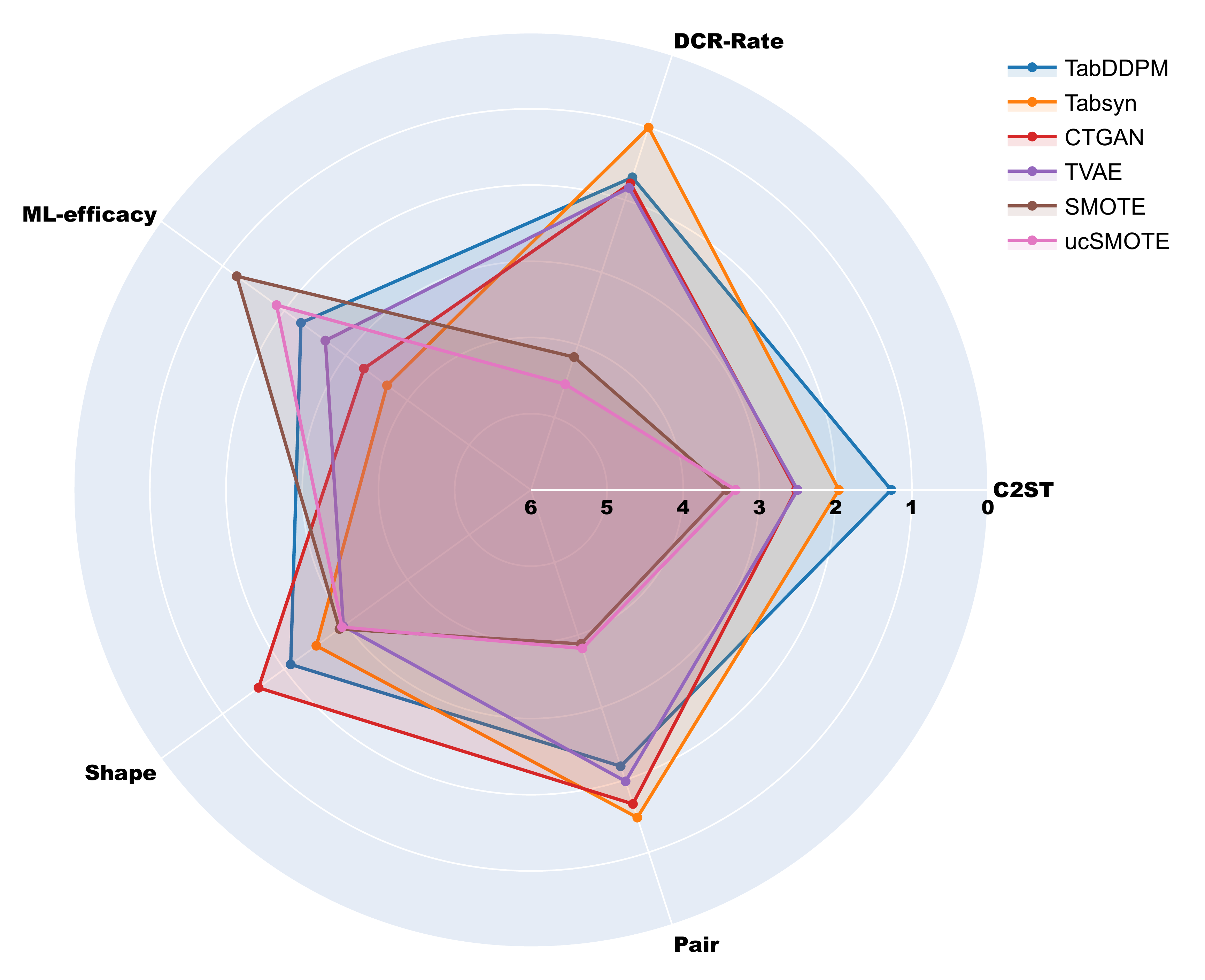}
    \caption{Radar chart of the optimized models in the limited-budget benchmark across various metrics. "Shape" stands for \textit{column-wise similarity}, and "Pair" stands for \textit{pair-wise correlation}.
    } 
    \label{fig:radar_rank_all_light}
\end{figure}

In Figure~\ref{fig:radar_rank_all_light} we summarize the model's relative performances according to the five metrics for the limited-budget tuning experiment.
To complete this point of view, Table~\ref{table:quartiles_light} summarizes 
the quality metric distributions among all datasets and folds. The full results are detailed in \ref{appendix:light_experiment}.

A first remark about Figure~\ref{fig:radar_rank_all_light} is that it is more balanced than  Figure~\ref{fig:radar_rank_all}: with a fair and limited GPU budget allocation, 
the models tends to perform similarly among all criteria.

There is no more clear leading model 
although TabDDPM obtains the best C2ST performance and TabSyn slightly dominates in terms of privacy (DCR-Rate).
We however have a safe median DCR-Rate score of about $0.63$ for this experiment on all neural models (against $0.62$ for the extensive experiment).
For ML-Efficacy, TabDDPM remains slightly the best but there is no clear leader either.

The reduced search space that we provide is hence a good starting point to perform
a quick dataset-specific hyperparameters optimization that will fit on a  medium-size workstation.

The relatively balanced results of this new  experiment confirms that the superiority of a tabular data generator is not only due to its model but also to the whole tuning and training compute effort.

Overall, TabDDPM provides a good balance between realism, privacy, utility, and cost. TVAE is a viable alternative if the utility constraint can be relaxed. It can be tuned using the limited-budget experiment search space to quickly achieve good performance. If resources are available and cost is not a priority, it is advisable to tune TabSyn, which achieves good realism results but at a higher cost. Finally, for quick results when privacy is not a concern and when the main focus is  utility, SMOTE is the recommended approach.

\section{Conclusion}
\label{sec:conclu}

\label{sec:concl}
We proposed a typology of state-of-art models for tabular data generation and we benchmarked five of them on $16$ datasets with a strict $3$-fold cross-validation procedure.
The number of folds was limited to  3 because increasing it would have dramatically increased the costs and carbon footprint.
We performed extensive large-scale tuning on a super computer from which we derived a reduced search-space.
Moreover, we performed a limited-budget benchmark that fits on a  medium-size workstation.
Leveraging these benchmarks we were able to provide several insights on the models while answering to three technical questions:
\begin{enumerate*}[label=(\roman*)]
    \item is it worth optimizing the hyperparameters/preprocessing specifically for each dataset?
    \item can we propose a reduced search space that fits well for all datasets?
    \item is there a clear trade-off between training/sampling costs, and synthetic data quality?
\end{enumerate*}

For the two first questions, Figure~\ref{fig:scatter_performance_comparison} is certainly the best summary: most models, including TabSyn, benefit greatly from a dataset-specific tuning.
However, the whole tuning process is costly in terms of time, money, energy, and CO$_2$ emissions, and there is clearly a "diminishing return" effect.
We conclude that, even for TabSyn, a quick dataset-specific model tuning based on the reduced search spaces we provide in~\ref{appendix:hp_search_space} is enough to get most of the performance at the scale of a medium-size workstation.  
We also recommend using a subset of
the dataset for hyperparameter tuning. The obtained hyperparameters can then be used to train the
model at full-scale.

Regarding the trade-off question on the multiple considered criteria: realism, privacy, utility and costs, Figure~\ref{fig:radar_rank_all} and
Figure~\ref{fig:radar_rank_all_light} confirm that if we do not limit the compute power, the two diffusion-based models TabDDPM and TabSyn are the most recommended solutions for tabular data generation.
Nevertheless, the performance gaps between models become quite narrow with a limited compute power.

\section{Future Work}
\label{sec:future}

Our study focuses on models trained, tuned, and evaluated on a single table and can be extended to evaluate cross-table models that fits with multiple table structures. This is an important step towards foundation models for tabular data generation. Models such as GReaT~\cite{great23} offers an interesting approach for a cross-table approach by converting the data generation process into a text generation one. However, when evaluated on a single tables, it struggle against models like CTGAN~\cite{CTGAN} for joint probability distribution~\cite{zhang2023mixed, du2024towards}. Working on a cross-table foundation model also introduces additional considerations towards data poisoning or adversarial attacks as foundation models are proned to being poisoned~\cite{WanPoisonLLM2023}. 

Another important research direction is the models' interpretability. Understanding the generation procedure is also important to build user's trust for decision making. Such interpretability study can be performed starting from a tree-based model such as Forest-Flow~\cite{jolicoeur2023generating}.

\section{Ackowlegment}
This work was granted access to the HPC resources of IDRIS under the allocations 2023-AD011014381, 2023-AD011011407R3, and 2023-AD011012220R2 made by GENCI.   
\bibliographystyle{elsarticle-num-names} 
\bibliography{biblioCharbel}
\label{sec:biblio}





\appendix
\label{sec:appendix}

\section{Benchmark challengers Hyperparameters}
\label{appendix:hp_search_space}

Hyperparameters search space of TVAE is in Table \ref{table:hp_tvae}, for CTGAN in \ref{table:hp_ctgan}, for TabSyn's VAE and MLP in \ref{table:hp_tsyn}, for TabDDPM in \ref{table:hp_tabddpm}, for smote and ucsmote in \ref{table:hp_smote}. These tables also present the reduced search spaces suggested and applied for the experiment of Section~\ref{sec:hpsens}.

\begin{table}[ht]
    \centering
    \resizebox{\textwidth}{!}{\begin{tabular}{|l|l|l|}
    \hline
        \multirow{2}{*}{\textbf{Parameter}} & \multicolumn{2}{c|}{\textbf{Possible values}} \\ 
        \cline{2-3}
        & \textbf{Extensive} & \textbf{Reduced} \\
        \hline
        Learning rate & qLogUniform(1e-4, 1e-2, 1e-4) & qLogUniform(1e-4, 7.3e-03, 1e-4)\\ \hline
        Batch size & [100, 500, 2000] & [100] \\ 
        \hline
        Embedding dim. & [16, 32, 64, 128, 256, 512] & [16, 32, 64]\\ \hline
        Encoder dim. & [64, 128, 256, 512] & [256, 512] \\ \hline
        Encoder depth & [2, 3, 4] & [2]\\ \hline
        Decoder dim. & [64, 128, 256, 512] & [256, 512]\\ \hline
        Decoder depth & [2, 3, 4] & [2, 4]\\ \hline
        Loss factor & [3, 2, 1, 0.5] & [3, 2] \\ \hline
        L2 scale & qLogUniform(1e-5, 1e-4, 1e-5) & qLogUniform(1e-5, 6.3e-5, 1e-6)\\ 
        \hline
        Numerical encoder & \makecell{[CDF, PLE\_CDF, PTP,\\ QuantileTransformer, MinMaxScaler, CBN\tablefootnote{\label{footnote_cbn}ClusterBasedNormalizer}]} & [CDF] \\ \hline
        Categorical encoder & [one-hot-encoder] & [one-hot-encoder] \\ 
        \hline
        Epochs & [400] & $\infty$ \\
        \specialrule{1.5pt}{1pt}{1pt}
        Number of trials per fold & 300 & 50 \\ \hline
    \end{tabular}}
    \caption{Hyperparameter search spaces of TVAE: extensive and limited-budget benchmarks.}
    \label{table:hp_tvae}
\end{table}

\begin{table}[ht]
    \centering
    \resizebox{\textwidth}{!}{\begin{tabular}{|l|l|l|}
    \hline
        \multirow{2}{*}{\textbf{Parameter}} & \multicolumn{2}{c|}{\textbf{Possible values}} \\ 
        \cline{2-3}
        & \textbf{Extensive} & \textbf{Reduced} \\
        \hline
        Discriminator learning rate & qLogUniform(5e-5, 1e-2, 5e-5) & qLogUniform(4e-4, 2.1e-03, 5e-5) \\ \hline
        Generator learning rate & qLogUniform(5e-5, 1e-2, 5e-5) & qLogUniform(5e-5, 1.3e-3, 5e-5) \\ \hline
        Batch size & [50, 100, 250, 500, 1000] & [100, 500, 1000]\\ 
        \hline
        Embedding dim. & [32, 64, 128, 256] & [32, 128] \\ \hline
        Generator dim. & [128, 256] & [128] \\ \hline
        Generator depth & [2, 3, 4] & [3, 4]\\ \hline
        Discriminator dim. & [128, 256] & [256] \\ \hline
        Discriminator depth & [2, 3] & [2, 3] \\ \hline
        Generator decay & qLogUniform(1e-6, 1e-5, 1e-6) & qLogUniform(1e-6, 6.4e-6, 1e-7) \\ \hline
        Discriminator decay & qLogUniform(1e-6,1e-5, 1e-6) & qLogUniform(1e-6, 8e-6, 1e-6) \\ \hline
        Log frequency & [False, True] & [False, True] \\ 
        \hline
        Numerical encoder & \makecell{[CDF, PLE\_CDF, PTP,\\ QuantileTransformer, MinMaxScaler, CBN\textsuperscript{\ref{footnote_cbn}}]} & [CDF, PLE\_CDF] \\ \hline
        Categorical encoder & [one-hot-encoder] & [one-hot-encoder] \\ 
        \hline
        Epochs & [400] & $\infty$ \\
        \specialrule{1.5pt}{1pt}{1pt}
        Number of trials per fold & 300 & 50 \\ 
        \hline
    \end{tabular}}
    \caption{Hyperparameter search spaces of CTGAN for the extensive and limited-budget benchmarks. }
    \label{table:hp_ctgan}
\end{table}

\begin{table}[ht]
    \centering
    \resizebox{\textwidth}{!}{\begin{tabular}{|l|l|l|l|}
    \hline
        \multirow{2}{*}{\textbf{Model}} & \multirow{2}{*}{\textbf{Parameter}} & \multicolumn{2}{c|}{\textbf{Possible values}} \\ 
        \cline{3-4}
        && \textbf{Extensive} & \textbf{Reduced} \\
        \hline
        \multirow{3}{*}{VAE} & Learning rate & qLogUniform(5e-5, 1e-2, 5e-5) & qLogUniform(1.1e-3, 7.2e-3, 5e-5) \\ 
        \cline{2-4}
        & Batch size & [1024, 2048, 4096] & [1024, 2048] \\ 
        \cline{2-4}
        & Weight decay & qLogUniform(1e-6, 1e-5, 1e-6) & qLogUniform(1e-6, 5e-6, 1e-6) \\ 
        \cline{2-4}
        & Token dim. & [2, 4] & [4] \\ 
        \cline{2-4}
        & Number of head & [1, 2] & [1, 2] \\ 
        \cline{2-4}
        & Factor & [8, 16, 32, 64] & [8, 16, 64] \\ 
        \cline{2-4}
        & Number of layers & [1, 2, 3, 4] & [1, 2, 4] \\ 
        \cline{2-4}
        & Max. beta & [1e-2] & [1e-2] \\ 
        \cline{2-4}
        & Min. beta & [1e-5] & [1e-5] \\ 
        \cline{2-4}
        & Lambda & [0.7, 0.8, 0.85, 0.9, 0.95] & [0.8, 0.85, 0.9, 0.95] \\ 
        \cline{2-4}
        & Epochs & [4000] & $\infty$ \\
        \cline{1-4}
        \multirow{3}{*}{MLP } & Learning Rate & qLogUniform(5e-5, 1e-2, 5e-5) & qLogUniform(7.7e-4, 2.5e-3, 1e-5) \\ 
        \cline{2-4}
        & Weight Decay & qLogUniform(1e-6, 1e-5, 1e-6) & qLogUniform(1e-6, 3.3e-6, 1e-7) \\ 
        \cline{2-4}
        & Batch size & [1024, 2048, 4096] & [1024, 4096] \\ 
        \cline{2-4}
        
        & MLP’s hidden dimension & [512, 1024] & [1024] \\ 
        \cline{2-4}
        & Epochs & [2000] & $\infty$ \\
        \specialrule{1.5pt}{1pt}{1pt}
        \multicolumn{2}{|c|}{Number of trials per fold} & 100 & 50 \\ 
                \hline

    \end{tabular}}
    \caption{Hyperparameter search spaces of  TabSyn's VAE and MLP for the extensive and limited-budget benchmarks. In the second benchmark, the number of epochs was bounded by a time budget (10 minutes for the VAE, 10 minutes for the denoiser).}
    \label{table:hp_tsyn}
\end{table}

\begin{table}[ht]
    \centering
    \resizebox{\textwidth}{!}{\begin{tabular}{|l|l|l|}
    \hline
        \multirow{2}{*}{\textbf{Parameter}} & \multicolumn{2}{c|}{\textbf{Possible values}} \\ 
        \cline{2-3}
        & \textbf{Extensive} & \textbf{Reduced} \\
        \hline
        Batch size & [256, 4096] & [4096] \\ \hline
        Dropout & [0.0] & [0.0] \\ \hline
        Number of timesteps & [1000] & [1000] \\ \hline
        Learning rate & qLogUniform(1e-5, 1e-3, 1e-5) & qLogUniform(3.5e-4, 9.2e-4, 1e-5) \\ \hline
        Number of layers & [2, 4, 6, 8] & [2, 4, 6] \\ \hline
        First layer's dim. & [128, 256, 512, 1024] & [256, 512, 1024] \\ \hline
        Middle layer's dim. & [128, 256, 512, 1024] & [512, 1024] \\ \hline
        Last layer's dim. & [128, 256, 512, 1024] & [256, 512, 1024] \\ 
        \hline
        Training iterations & [20000] & $\infty$ \\
        \specialrule{1.5pt}{1pt}{1pt}
        Number of trials per fold & 300 & 50 \\ 
        \hline

    \end{tabular}}
    \caption{Hyperparameter search spaces of  TabDDPM for the extensive and limited-budget benchmarks.}
    \label{table:hp_tabddpm}
\end{table}

\begin{table}[htbp]
    \centering
    \begin{tabular}{|l|l|}
    \hline
        Parameter & Possible values \\ \hline
        K-Neighbors & Grid search in range [2, 20] \\ 
        \specialrule{1.5pt}{1pt}{1pt}
        Number of trials per fold & 38 \\ 
        \hline

    \end{tabular}
    \caption{Hyperparameter search space of SMOTE and ucSMOTE.}
    \label{table:hp_smote}
\end{table}

\section{Datasets Links}
\label{appendix:datasets_links}
We provide the list of links toward the datasets that were used in this paper in Table~\ref{table:dataset_links} below.

\noindent

\begin{table}[h!]
    \centering
\begin{tabular}{ll@{\quad{}}l}
\hline
    Domains & {Dataset}   &  {URL}\\
\hline  
    Environment & \small Abalone &\footnotesize\href{https://www.openml.org/d/183}{www.openml.org/d/183}\\
    Social & \small Adult&\footnotesize\href{https://www.openml.org/d/1590}{www.openml.org/d/1590}\\
    Finance / Marketing & \small Bank&\footnotesize\href{https://www.openml.org/d/1461}{www.openml.org/d/1461}\\
    Marketing & \small Black Friday&\footnotesize\href{https://www.openml.org/d/41540}{www.openml.org/d/41540}\\
    Social / Environment & \small Bike Sharing&\footnotesize\href{https://www.openml.org/d/42712}{www.openml.org/d/42712}\\
    Environment & \small Covertype&\footnotesize\href{https://www.openml.org/d/150}{www.openml.org/d/150}\\
    Health & \small Cardio&\footnotesize\href{https://www.kaggle.com/datasets/sulianova/cardiovascular-disease-dataset}{www.kaggle.com/sulianova/datasets}\\
    Finance / Marketing & \small Churn&\footnotesize\href{https://www.kaggle.com/datasets/shrutimechlearn/churn-modelling}{www.kaggle.com/datasets/shrutimechlearn}\\
    Finance & \small Diamonds&\footnotesize\href{https://www.openml.org/d/42225}{www.openml.org/d/42225}\\
    Finance / Social & \small HELOC&\footnotesize\href{https://www.kaggle.com/datasets/averkiyoliabev/heloc-nnsmall}{www.kaggle.com/averkiyoliabev/datasets}\\
    Physics & \small Higgs&\footnotesize\href{https://www.openml.org/d/4532}{www.openml.org/d/4532}\\
    Finance / Social & \small House 16H&\footnotesize\href{https://www.openml.org/d/574}{www.openml.org/d/574}\\
    Health / Insurance & \small Insurance&\footnotesize\href{https://www.kaggle.com/datasets/mirichoi0218/insurance}{www.kaggle.com/datasets/mirichoi0218/insurance}\\
    Finance / Social & \small King&\footnotesize\href{https://www.kaggle.com/datasets/harlfoxem/housesalesprediction}{www.kaggle.com/datasets/harlfoxem/housesalesprediction}\\
    Physics & \small MiniBooNE&\footnotesize\href{https://www.openml.org/d/41150}{www.openml.org/d/41150}\\
    Synthetic & \small Moons&\footnotesize\href{https://scikit-learn.org/stable/modules/generated/sklearn.datasets.make_moons.html#sklearn.datasets.make_moons}{scikit-learn.org/stable/modules/classes.html}\\
\hline     
\end{tabular}
    \caption{Domains and links to the datasets}
    \label{table:dataset_links}
\end{table}

\section{Dataset-Level Results for Large-Scale Experiment}
\label{appendix:results}

Table~\ref{table:results} present the per-dataset performance according to the metrics described in Section \ref{s:metrics} averaged on $3$-folds with $5$ samples per-fold.


            

\begin{tiny}\setlength\tabcolsep{4pt}\setstretch{1.5} %
            \begin{longtable}{|c|l|*{7}{l|}}
            
            \hline
            \multirow{2}{*}{Dataset} & \multirow{2}{*}{Model} & \multicolumn{7}{c|}{Metrics} \\
                                            \cline{3-9}
                                            && C2ST $\downarrow$ & DCR-Rate $\downarrow$ & ML-Efficacy $\uparrow$ & Shape $\uparrow$ & Pair $\uparrow$ & Train time $\downarrow$ & Sample time $\downarrow$ \\
            
            \hline
            \endfirsthead
            \caption*{Results per datasets and models under diverse metrics for the extensive search (continued).}\\
            \multirow{2}{*}{Dataset} & \multirow{2}{*}{Model} & \multicolumn{7}{c|}{Metrics} \\
                                            \cline{3-9}
                                            && C2ST $\downarrow$ & DCR-Rate $\downarrow$ & ML-Efficacy $\uparrow$ & Shape $\uparrow$ & Pair $\uparrow$ & Train time $\downarrow$ & Sample time $\downarrow$ \\
            
             \hline
            \endhead
            
            \multirow{3}{*}{Abalone} & Train Copy & $0.51 \pm 0.00$ & $1.00 \pm 0.00$ & $0.23 \pm 0.01$ & $0.96 \pm 0.01$ & $0.88 \pm 0.01$ & - & - \\  
 \cline{2-9} 
 & CTGAN & \textcolor{ForestGreen}{\bm{$0.73 \pm 0.03$} } & \textcolor{ForestGreen}{\bm{$0.63 \pm 0.02$} } & \textcolor{red}{$0.17 \pm 0.01$ } & $0.93 \pm 0.01$ & $0.86 \pm 0.03$ & $685 \pm 398.05$ & $00 \pm 0.01$ \\  
 \cline{2-9} 
 & TVAE & $0.75 \pm 0.07$ & $0.64 \pm 0.05$ & $0.23 \pm 0.02$ & $0.91 \pm 0.02$ & $0.86 \pm 0.04$ & \textcolor{ForestGreen}{\bm{$113 \pm 3.57$} } & \textcolor{ForestGreen}{\bm{$00 \pm 0.00$} } \\  
 \cline{2-9} 
 & TabDDPM & $0.78 \pm 0.01$ & $0.67 \pm 0.01$ & $0.23 \pm 0.01$ & $0.94 \pm 0.00$ & $0.85 \pm 0.03$ & $169 \pm 28.82$ & \textcolor{red}{$03 \pm 0.62$ } \\  
 \cline{2-9} 
 & TabSyn & $0.78 \pm 0.01$ & $0.63 \pm 0.01$ & $0.22 \pm 0.01$ & $0.94 \pm 0.02$ & $0.87 \pm 0.00$ & \textcolor{red}{$1056 \pm 183.20$ } & $00 \pm 0.08$ \\  
 \cline{2-9} 
 & SMOTE & \textcolor{red}{$1.00 \pm 0.00$ } & $0.85 \pm 0.02$ & $0.50 \pm 0.02$ & \textcolor{red}{$0.85 \pm 0.01$ } & \textcolor{red}{$0.72 \pm 0.01$ } & - & $00 \pm 0.01$ \\  
 \cline{2-9} 
 & UC-SMOTE & $0.89 \pm 0.01$ & \textcolor{red}{$0.92 \pm 0.02$ } & \textcolor{ForestGreen}{\bm{$0.51 \pm 0.03$} } & \textcolor{ForestGreen}{\bm{$0.95 \pm 0.01$} } & \textcolor{ForestGreen}{\bm{$0.88 \pm 0.03$} } & - & $00 \pm 0.00$ \\  
 \hline 
\multirow{3}{*}{Adult} & Train Copy & $0.50 \pm 0.00$ & $1.00 \pm 0.00$ & $0.71 \pm 0.01$ & $0.99 \pm 0.00$ & $0.98 \pm 0.00$ & - & - \\  
 \cline{2-9} 
 & CTGAN & $0.74 \pm 0.02$ & $0.72 \pm 0.00$ & $0.67 \pm 0.01$ & $0.97 \pm 0.00$ & \textcolor{red}{$0.88 \pm 0.01$ } & $2755 \pm 920.97$ & $00 \pm 0.01$ \\  
 \cline{2-9} 
 & TVAE & $0.77 \pm 0.01$ & $0.72 \pm 0.00$ & \textcolor{red}{$0.63 \pm 0.03$ } & $0.96 \pm 0.00$ & $0.91 \pm 0.02$ & $1404 \pm 2.70$ & \textcolor{ForestGreen}{\bm{$00 \pm 0.00$} } \\  
 \cline{2-9} 
 & TabDDPM & $0.65 \pm 0.00$ & \textcolor{ForestGreen}{\bm{$0.62 \pm 0.01$} } & $0.67 \pm 0.01$ & \textcolor{ForestGreen}{\bm{$0.98 \pm 0.00$} } & $0.95 \pm 0.01$ & \textcolor{ForestGreen}{\bm{$443 \pm 44.97$} } & \textcolor{red}{$10 \pm 2.22$ } \\  
 \cline{2-9} 
 & TabSyn & \textcolor{ForestGreen}{\bm{$0.64 \pm 0.01$} } & $0.62 \pm 0.00$ & $0.66 \pm 0.01$ & $0.98 \pm 0.00$ & \textcolor{ForestGreen}{\bm{$0.96 \pm 0.01$} } & \textcolor{red}{$5042 \pm 2119.47$ } & $01 \pm 1.06$ \\  
 \cline{2-9} 
 & SMOTE & \textcolor{red}{$0.93 \pm 0.00$ } & $0.85 \pm 0.00$ & \textcolor{ForestGreen}{\bm{$0.69 \pm 0.01$} } & \textcolor{red}{$0.95 \pm 0.00$ } & $0.90 \pm 0.01$ & - & $06 \pm 0.11$ \\  
 \cline{2-9} 
 & UC-SMOTE & $0.93 \pm 0.00$ & \textcolor{red}{$0.86 \pm 0.01$ } & $0.67 \pm 0.01$ & $0.95 \pm 0.00$ & $0.90 \pm 0.01$ & - & $10 \pm 0.03$ \\  
 \hline 
\multirow{3}{*}{\shortstack{Bank\\ marketing}} & Train Copy & $0.50 \pm 0.00$ & $1.00 \pm 0.00$ & $0.54 \pm 0.01$ & $0.99 \pm 0.00$ & $0.98 \pm 0.01$ & - & - \\  
 \cline{2-9} 
 & CTGAN & $0.72 \pm 0.01$ & $0.63 \pm 0.00$ & $0.47 \pm 0.02$ & $0.98 \pm 0.00$ & $0.95 \pm 0.01$ & \textcolor{red}{$3159 \pm 65.25$ } & $00 \pm 0.01$ \\  
 \cline{2-9} 
 & TVAE & \textcolor{red}{$0.81 \pm 0.04$ } & $0.63 \pm 0.00$ & \textcolor{red}{$0.45 \pm 0.08$ } & \textcolor{red}{$0.96 \pm 0.01$ } & \textcolor{red}{$0.92 \pm 0.02$ } & $1089 \pm 628.11$ & \textcolor{ForestGreen}{\bm{$00 \pm 0.02$} } \\  
 \cline{2-9} 
 & TabDDPM & $0.65 \pm 0.01$ & \textcolor{ForestGreen}{\bm{$0.61 \pm 0.00$} } & $0.52 \pm 0.01$ & \textcolor{ForestGreen}{\bm{$0.99 \pm 0.01$} } & $0.96 \pm 0.01$ & \textcolor{ForestGreen}{\bm{$461 \pm 42.63$} } & \textcolor{red}{$10 \pm 1.98$ } \\  
 \cline{2-9} 
 & TabSyn & \textcolor{ForestGreen}{\bm{$0.61 \pm 0.02$} } & $0.62 \pm 0.01$ & $0.49 \pm 0.02$ & $0.99 \pm 0.01$ & \textcolor{ForestGreen}{\bm{$0.97 \pm 0.01$} } & $2783 \pm 471.56$ & $02 \pm 0.03$ \\  
 \cline{2-9} 
 & SMOTE & $0.79 \pm 0.01$ & \textcolor{red}{$0.98 \pm 0.00$ } & \textcolor{ForestGreen}{\bm{$0.53 \pm 0.02$} } & $0.97 \pm 0.00$ & $0.95 \pm 0.00$ & - & $09 \pm 0.16$ \\  
 \cline{2-9} 
 & UC-SMOTE & $0.79 \pm 0.00$ & $0.98 \pm 0.00$ & $0.46 \pm 0.02$ & $0.97 \pm 0.00$ & $0.95 \pm 0.00$ & - & $09 \pm 0.03$ \\  
 \hline 
\multirow{3}{*}{\shortstack{Bike\\ sharing}} & Train Copy & $0.50 \pm 0.01$ & $1.00 \pm 0.00$ & $0.94 \pm 0.00$ & $0.99 \pm 0.00$ & $0.53 \pm 0.00$ & - & - \\  
 \cline{2-9} 
 & CTGAN & $0.90 \pm 0.01$ & \textcolor{ForestGreen}{\bm{$0.61 \pm 0.00$} } & $0.68 \pm 0.02$ & \textcolor{ForestGreen}{\bm{$0.97 \pm 0.01$} } & \textcolor{red}{$0.53 \pm 0.00$ } & \textcolor{red}{$3772 \pm 2301.33$ } & $00 \pm 0.05$ \\  
 \cline{2-9} 
 & TVAE & $0.81 \pm 0.01$ & $0.61 \pm 0.00$ & $0.78 \pm 0.03$ & $0.96 \pm 0.01$ & $0.53 \pm 0.00$ & $540 \pm 15.46$ & \textcolor{ForestGreen}{\bm{$00 \pm 0.00$} } \\  
 \cline{2-9} 
 & TabDDPM & \textcolor{ForestGreen}{\bm{$0.81 \pm 0.01$} } & $0.62 \pm 0.00$ & $0.79 \pm 0.01$ & $0.97 \pm 0.01$ & $0.53 \pm 0.00$ & \textcolor{ForestGreen}{\bm{$332 \pm 13.26$} } & \textcolor{red}{$07 \pm 0.90$ } \\  
 \cline{2-9} 
 & TabSyn & $0.86 \pm 0.02$ & $0.62 \pm 0.01$ & \textcolor{red}{$0.51 \pm 0.07$ } & $0.96 \pm 0.00$ & $0.53 \pm 0.00$ & $1843 \pm 556.82$ & $00 \pm 0.01$ \\  
 \cline{2-9} 
 & SMOTE & \textcolor{red}{$0.98 \pm 0.00$ } & \textcolor{red}{$0.99 \pm 0.01$ } & $0.85 \pm 0.00$ & $0.95 \pm 0.00$ & \textcolor{ForestGreen}{\bm{$0.67 \pm 0.08$} } & - & $01 \pm 0.06$ \\  
 \cline{2-9} 
 & UC-SMOTE & $0.98 \pm 0.00$ & $0.98 \pm 0.00$ & \textcolor{ForestGreen}{\bm{$0.86 \pm 0.01$} } & \textcolor{red}{$0.95 \pm 0.01$ } & $0.67 \pm 0.08$ & - & $00 \pm 0.09$ \\  
 \hline 
\multirow{3}{*}{\shortstack{Black\\ friday}} & Train Copy & $0.50 \pm 0.00$ & $1.00 \pm 0.00$ & $0.53 \pm 0.01$ & $1.00 \pm 0.00$ & $0.99 \pm 0.00$ & - & - \\  
 \cline{2-9} 
 & CTGAN & $0.82 \pm 0.02$ & $0.66 \pm 0.01$ & $0.46 \pm 0.02$ & $0.97 \pm 0.01$ & $0.87 \pm 0.00$ & \textcolor{red}{$10111 \pm 8081.87$ } & $00 \pm 0.12$ \\  
 \cline{2-9} 
 & TVAE & $0.87 \pm 0.01$ & $0.68 \pm 0.01$ & $0.42 \pm 0.02$ & $0.95 \pm 0.01$ & $0.91 \pm 0.02$ & $4978 \pm 1.32$ & \textcolor{ForestGreen}{\bm{$00 \pm 0.02$} } \\  
 \cline{2-9} 
 & TabDDPM & \textcolor{red}{$0.87 \pm 0.01$ } & \textcolor{ForestGreen}{\bm{$0.64 \pm 0.00$} } & $0.47 \pm 0.02$ & $0.99 \pm 0.01$ & \textcolor{ForestGreen}{\bm{$0.98 \pm 0.01$} } & \textcolor{ForestGreen}{\bm{$366 \pm 13.15$} } & $39 \pm 14.32$ \\  
 \cline{2-9} 
 & TabSyn & $0.87 \pm 0.01$ & $0.68 \pm 0.00$ & \textcolor{red}{$0.17 \pm 0.01$ } & \textcolor{ForestGreen}{\bm{$0.99 \pm 0.00$} } & \textcolor{red}{$0.46 \pm 0.00$ } & $7533 \pm 563.20$ & $06 \pm 3.54$ \\  
 \cline{2-9} 
 & SMOTE & \textcolor{ForestGreen}{\bm{$0.80 \pm 0.00$} } & \textcolor{red}{$0.97 \pm 0.00$ } & \textcolor{ForestGreen}{\bm{$0.50 \pm 0.01$} } & $0.95 \pm 0.00$ & $0.94 \pm 0.00$ & - & $31 \pm 0.16$ \\  
 \cline{2-9} 
 & UC-SMOTE & $0.81 \pm 0.01$ & $0.97 \pm 0.00$ & $0.49 \pm 0.01$ & \textcolor{red}{$0.94 \pm 0.00$ } & $0.93 \pm 0.00$ & - & \textcolor{red}{$106 \pm 0.07$ } \\  
 \hline 
\multirow{3}{*}{Cardio} & Train Copy & $0.50 \pm 0.00$ & $1.00 \pm 0.00$ & $0.72 \pm 0.01$ & $1.00 \pm 0.00$ & $0.98 \pm 0.01$ & - & - \\  
 \cline{2-9} 
 & CTGAN & $0.62 \pm 0.01$ & $0.64 \pm 0.00$ & \textcolor{red}{$0.70 \pm 0.02$ } & $0.99 \pm 0.01$ & $0.96 \pm 0.01$ & \textcolor{red}{$4678 \pm 59.79$ } & $00 \pm 0.01$ \\  
 \cline{2-9} 
 & TVAE & $0.72 \pm 0.02$ & $0.64 \pm 0.01$ & $0.72 \pm 0.01$ & $0.97 \pm 0.01$ & \textcolor{red}{$0.95 \pm 0.02$ } & $1708 \pm 984.53$ & \textcolor{ForestGreen}{\bm{$00 \pm 0.02$} } \\  
 \cline{2-9} 
 & TabDDPM & \textcolor{ForestGreen}{\bm{$0.55 \pm 0.01$} } & \textcolor{ForestGreen}{\bm{$0.62 \pm 0.00$} } & $0.72 \pm 0.01$ & \textcolor{ForestGreen}{\bm{$0.99 \pm 0.00$} } & \textcolor{ForestGreen}{\bm{$0.98 \pm 0.01$} } & \textcolor{ForestGreen}{\bm{$185 \pm 73.66$} } & \textcolor{red}{$13 \pm 8.85$ } \\  
 \cline{2-9} 
 & TabSyn & $0.56 \pm 0.00$ & $0.64 \pm 0.00$ & $0.72 \pm 0.00$ & $0.99 \pm 0.00$ & $0.98 \pm 0.01$ & $2960 \pm 578.82$ & $03 \pm 0.01$ \\  
 \cline{2-9} 
 & SMOTE & \textcolor{red}{$0.95 \pm 0.00$ } & \textcolor{red}{$0.98 \pm 0.00$ } & \textcolor{ForestGreen}{\bm{$0.73 \pm 0.00$} } & \textcolor{red}{$0.93 \pm 0.00$ } & $0.97 \pm 0.01$ & - & $00 \pm 0.02$ \\  
 \cline{2-9} 
 & UC-SMOTE & $0.94 \pm 0.00$ & $0.98 \pm 0.00$ & $0.72 \pm 0.01$ & $0.93 \pm 0.01$ & $0.97 \pm 0.01$ & - & $01 \pm 0.05$ \\  
 \hline 
\multirow{3}{*}{Churn} & Train Copy & $0.50 \pm 0.01$ & $1.00 \pm 0.00$ & $0.59 \pm 0.02$ & $0.95 \pm 0.00$ & $0.87 \pm 0.01$ & - & - \\*  
 \cline{2-9} 
 & CTGAN & $0.63 \pm 0.02$ & $0.64 \pm 0.01$ & $0.36 \pm 0.03$ & $0.93 \pm 0.01$ & $0.84 \pm 0.02$ & \textcolor{red}{$2505 \pm 953.49$ } & $00 \pm 0.03$ \\*  
 \cline{2-9} 
 & TVAE & $0.64 \pm 0.01$ & $0.63 \pm 0.00$ & $0.51 \pm 0.00$ & $0.92 \pm 0.01$ & \textcolor{ForestGreen}{\bm{$0.84 \pm 0.01$} } & \textcolor{ForestGreen}{\bm{$294 \pm 6.20$} } & \textcolor{ForestGreen}{\bm{$00 \pm 0.01$} } \\*  
 \cline{2-9} 
 & TabDDPM & $0.66 \pm 0.07$ & $0.64 \pm 0.05$ & $0.50 \pm 0.05$ & $0.89 \pm 0.06$ & $0.82 \pm 0.06$ & $618 \pm 35.47$ & \textcolor{red}{$25 \pm 2.36$ } \\*  
 \cline{2-9} 
 & TabSyn & \textcolor{ForestGreen}{\bm{$0.58 \pm 0.02$} } & \textcolor{ForestGreen}{\bm{$0.61 \pm 0.01$} } & \textcolor{ForestGreen}{\bm{$0.56 \pm 0.03$} } & \textcolor{ForestGreen}{\bm{$0.93 \pm 0.01$} } & \textcolor{red}{$0.48 \pm 0.00$ } & $1627 \pm 340.95$ & $00 \pm 0.22$ \\*  
 \cline{2-9} 
 & SMOTE & $0.76 \pm 0.01$ & $0.86 \pm 0.03$ & $0.50 \pm 0.02$ & $0.87 \pm 0.01$ & $0.81 \pm 0.02$ & - & $01 \pm 0.06$ \\*  
 \cline{2-9} 
 & UC-SMOTE & \textcolor{red}{$0.78 \pm 0.02$ } & \textcolor{red}{$0.93 \pm 0.02$ } & \textcolor{red}{$0.12 \pm 0.09$ } & \textcolor{red}{$0.87 \pm 0.01$ } & $0.80 \pm 0.02$ & - & $00 \pm 0.05$ \\  
 \hline
 \pagebreak
\nopagebreak\multirow{3}{*}{Covertype} & Train Copy & $0.50 \pm 0.00$ & $1.00 \pm 0.00$ & $0.90 \pm 0.00$ & $1.00 \pm 0.00$ & $1.00 \pm 0.01$ & - & - \\*  
 \nopagebreak\cline{2-9} \nopagebreak
 & CTGAN & $0.97 \pm 0.01$ & $0.60 \pm 0.01$ & $0.70 \pm 0.01$ & $0.98 \pm 0.00$ & $0.96 \pm 0.01$ & \textcolor{red}{$51644 \pm 17257.41$ } & $04 \pm 0.60$ \\*  
 \nopagebreak\cline{2-9} \nopagebreak
 & TVAE & $0.90 \pm 0.01$ & $0.60 \pm 0.00$ & $0.77 \pm 0.01$ & $0.98 \pm 0.00$ & $0.96 \pm 0.01$ & $23219 \pm 18238.02$ & \textcolor{ForestGreen}{\bm{$01 \pm 0.44$} } \\*  
 \nopagebreak\cline{2-9} 
 \nopagebreak& TabDDPM & $0.94 \pm 0.01$ & \textcolor{ForestGreen}{\bm{$0.59 \pm 0.04$} } & \textcolor{red}{$0.66 \pm 0.04$ } & \textcolor{red}{$0.95 \pm 0.01$ } & $0.91 \pm 0.01$ & \textcolor{ForestGreen}{\bm{$868 \pm 30.61$} } & \textcolor{red}{$249 \pm 25.67$ }\\*
 \nopagebreak\cline{2-9} 
 \nopagebreak& TabSyn & \textcolor{ForestGreen}{\bm{$0.63 \pm 0.02$} } & $0.62 \pm 0.01$ & $0.75 \pm 0.02$ & $0.99 \pm 0.00$ & \textcolor{red}{$0.69 \pm 0.00$ } & $4937 \pm 1556.02$ & $03 \pm 0.05$ \\*  
 \nopagebreak\cline{2-9}
 \nopagebreak& SMOTE & \textcolor{red}{$0.97 \pm 0.00$ } & \textcolor{red}{$0.97 \pm 0.00$ } & \textcolor{ForestGreen}{\bm{$0.90 \pm 0.00$} } & \textcolor{ForestGreen}{\bm{$1.00 \pm 0.00$} } & \textcolor{ForestGreen}{\bm{$1.00 \pm 0.01$} } & - & $125 \pm 0.69$ \\*  
 \cline{2-9} \nopagebreak  
 & UC-SMOTE & $0.97 \pm 0.00$ & $0.97 \pm 0.00$ & $0.90 \pm 0.00$ & $1.00 \pm 0.00$ & $0.99 \pm 0.01$ & - & $138 \pm 0.71$ \\  
 \hline 
\multirow{3}{*}{Diamonds} & Train Copy & $0.50 \pm 0.00$ & $1.00 \pm 0.00$ & $0.98 \pm 0.00$ & $0.99 \pm 0.00$ & $0.77 \pm 0.01$ & - & - \\* 
 \cline{2-9} 
 & CTGAN & $0.86 \pm 0.01$ & $0.65 \pm 0.01$ & $0.94 \pm 0.01$ & $0.97 \pm 0.01$ & $0.80 \pm 0.04$ & \textcolor{red}{$3389 \pm 2.57$ } & $00 \pm 0.01$ \\*  
 \cline{2-9} 
 & TVAE & $0.79 \pm 0.03$ & $0.64 \pm 0.00$ & $0.96 \pm 0.01$ & \textcolor{red}{$0.94 \pm 0.01$ } & $0.74 \pm 0.04$ & $1104 \pm 637.85$ & \textcolor{ForestGreen}{\bm{$00 \pm 0.02$} } \\  
 \cline{2-9} 
 & TabDDPM & \textcolor{ForestGreen}{\bm{$0.71 \pm 0.01$} } & \textcolor{ForestGreen}{\bm{$0.61 \pm 0.00$} } & \textcolor{ForestGreen}{\bm{$0.97 \pm 0.00$} } & \textcolor{ForestGreen}{\bm{$0.98 \pm 0.01$} } & \textcolor{red}{$0.70 \pm 0.02$ } & \textcolor{ForestGreen}{\bm{$374 \pm 13.27$} } & \textcolor{red}{$16 \pm 6.28$ } \\  
 \cline{2-9} 
 & TabSyn & $0.87 \pm 0.01$ & $0.65 \pm 0.00$ & \textcolor{red}{$0.79 \pm 0.10$ } & $0.98 \pm 0.01$ & $0.76 \pm 0.04$ & $3255 \pm 360.89$ & $02 \pm 1.12$ \\  
 \cline{2-9} 
 & SMOTE & \textcolor{red}{$0.97 \pm 0.00$ } & $0.93 \pm 0.01$ & $0.92 \pm 0.01$ & $0.97 \pm 0.00$ & $0.77 \pm 0.02$ & - & $03 \pm 0.02$ \\  
 \cline{2-9} 
 & UC-SMOTE & $0.97 \pm 0.00$ & \textcolor{red}{$0.94 \pm 0.01$ } & $0.93 \pm 0.01$ & $0.97 \pm 0.00$ & \textcolor{ForestGreen}{\bm{$0.80 \pm 0.04$} } & - & $01 \pm 0.04$ \\  
 \hline 
\multirow{3}{*}{Heloc} & Train Copy & $0.50 \pm 0.01$ & $1.00 \pm 0.00$ & $0.70 \pm 0.01$ & $0.99 \pm 0.00$ & $0.97 \pm 0.01$ & - & - \\  
 \cline{2-9} 
 & CTGAN & $0.92 \pm 0.01$ & \textcolor{ForestGreen}{\bm{$0.64 \pm 0.01$} } & \textcolor{red}{$0.66 \pm 0.05$ } & \textcolor{ForestGreen}{\bm{$0.97 \pm 0.01$} } & \textcolor{red}{$0.94 \pm 0.02$ } & \textcolor{red}{$4330 \pm 3393.17$ } & $00 \pm 0.10$ \\  
 \cline{2-9} 
 & TVAE & $0.87 \pm 0.00$ & $0.66 \pm 0.01$ & $0.70 \pm 0.01$ & $0.95 \pm 0.00$ & $0.94 \pm 0.01$ & $421 \pm 249.85$ & \textcolor{ForestGreen}{\bm{$00 \pm 0.00$} } \\  
 \cline{2-9} 
 & TabDDPM & $0.72 \pm 0.01$ & $0.64 \pm 0.01$ & \textcolor{ForestGreen}{\bm{$0.70 \pm 0.01$} } & $0.97 \pm 0.00$ & $0.95 \pm 0.01$ & \textcolor{ForestGreen}{\bm{$80 \pm 9.69$} } & \textcolor{red}{$01 \pm 0.53$ } \\  
 \cline{2-9} 
 & TabSyn & \textcolor{ForestGreen}{\bm{$0.70 \pm 0.02$} } & $0.66 \pm 0.01$ & $0.70 \pm 0.02$ & $0.97 \pm 0.00$ & $0.95 \pm 0.01$ & $1870 \pm 302.36$ & $00 \pm 0.00$ \\  
 \cline{2-9} 
 & SMOTE & \textcolor{red}{$0.92 \pm 0.01$ } & \textcolor{red}{$1.00 \pm 0.00$ } & $0.69 \pm 0.01$ & \textcolor{red}{$0.94 \pm 0.00$ } & \textcolor{ForestGreen}{\bm{$0.95 \pm 0.01$} } & - & $00 \pm 0.04$ \\  
 \cline{2-9} 
 & UC-SMOTE & $0.92 \pm 0.01$ & $1.00 \pm 0.00$ & $0.69 \pm 0.01$ & $0.94 \pm 0.00$ & $0.95 \pm 0.02$ & - & $00 \pm 0.02$ \\  
 \hline 
\multirow{3}{*}{Higgs} & Train Copy & $0.50 \pm 0.00$ & $1.00 \pm 0.00$ & $0.74 \pm 0.00$ & $0.99 \pm 0.00$ & $0.99 \pm 0.00$ & - & - \\  
 \cline{2-9} 
 & CTGAN & $0.85 \pm 0.02$ & \textcolor{ForestGreen}{\bm{$0.60 \pm 0.00$} } & \textcolor{red}{$0.70 \pm 0.01$ } & $0.99 \pm 0.01$ & $0.98 \pm 0.00$ & \textcolor{red}{$12696 \pm 4197.49$ } & $00 \pm 0.13$ \\  
 \cline{2-9} 
 & TVAE & \textcolor{red}{$0.92 \pm 0.01$ } & $0.60 \pm 0.00$ & $0.70 \pm 0.02$ & $0.94 \pm 0.01$ & \textcolor{red}{$0.98 \pm 0.01$ } & $5752 \pm 116.50$ & \textcolor{ForestGreen}{\bm{$00 \pm 0.08$} } \\  
 \cline{2-9} 
 & TabDDPM & $0.57 \pm 0.00$ & $0.63 \pm 0.00$ & $0.73 \pm 0.00$ & \textcolor{ForestGreen}{\bm{$0.99 \pm 0.00$} } & \textcolor{ForestGreen}{\bm{$0.99 \pm 0.00$} } & \textcolor{ForestGreen}{\bm{$223 \pm 10.78$} } & \textcolor{red}{$22 \pm 2.25$ } \\  
 \cline{2-9} 
 & TabSyn & \textcolor{ForestGreen}{\bm{$0.57 \pm 0.01$} } & $0.61 \pm 0.00$ & $0.73 \pm 0.00$ & \textcolor{red}{$0.94 \pm 0.00$ } & $0.99 \pm 0.00$ & $7503 \pm 2581.02$ & $04 \pm 2.17$ \\  
 \cline{2-9} 
 & SMOTE & $0.84 \pm 0.00$ & \textcolor{red}{$1.00 \pm 0.00$ } & \textcolor{ForestGreen}{\bm{$0.73 \pm 0.00$} } & $0.96 \pm 0.00$ & $0.98 \pm 0.01$ & - & $00 \pm 0.05$ \\  
 \cline{2-9} 
 & UC-SMOTE & $0.83 \pm 0.00$ & $1.00 \pm 0.00$ & $0.73 \pm 0.00$ & $0.96 \pm 0.00$ & $0.98 \pm 0.01$ & - & $01 \pm 0.04$ \\  
 \hline 
\multirow{3}{*}{\shortstack{House\\ 16h}} & Train Copy & $0.50 \pm 0.01$ & $1.00 \pm 0.00$ & $0.64 \pm 0.01$ & $0.99 \pm 0.01$ & $0.99 \pm 0.00$ & - & - \\  
 \cline{2-9} 
 & CTGAN & $0.84 \pm 0.01$ & $0.62 \pm 0.00$ & $0.47 \pm 0.01$ & $0.98 \pm 0.01$ & \textcolor{red}{$0.97 \pm 0.01$ } & $983 \pm 559.75$ & \textcolor{ForestGreen}{\bm{$00 \pm 0.01$} } \\  
 \cline{2-9} 
 & TVAE & $0.82 \pm 0.03$ & $0.61 \pm 0.01$ & $0.48 \pm 0.06$ & $0.95 \pm 0.01$ & $0.98 \pm 0.01$ & $922 \pm 6.68$ & $00 \pm 0.01$ \\  
 \cline{2-9} 
 & TabDDPM & \textcolor{ForestGreen}{\bm{$0.60 \pm 0.01$} } & \textcolor{ForestGreen}{\bm{$0.61 \pm 0.00$} } & $0.61 \pm 0.02$ & \textcolor{ForestGreen}{\bm{$0.98 \pm 0.00$} } & $0.99 \pm 0.01$ & \textcolor{ForestGreen}{\bm{$95 \pm 21.99$} } & \textcolor{red}{$02 \pm 0.88$ } \\  
 \cline{2-9} 
 & TabSyn & $0.74 \pm 0.01$ & $0.62 \pm 0.01$ & \textcolor{red}{$0.40 \pm 0.02$ } & $0.97 \pm 0.00$ & $0.99 \pm 0.00$ & \textcolor{red}{$2370 \pm 810.03$ } & $01 \pm 0.01$ \\  
 \cline{2-9} 
 & SMOTE & $0.87 \pm 0.00$ & \textcolor{red}{$0.86 \pm 0.05$ } & \textcolor{ForestGreen}{\bm{$0.62 \pm 0.01$} } & \textcolor{red}{$0.92 \pm 0.01$ } & \textcolor{ForestGreen}{\bm{$0.99 \pm 0.00$} } & - & $00 \pm 0.03$ \\  
 \cline{2-9} 
 & UC-SMOTE & \textcolor{red}{$0.87 \pm 0.01$ } & $0.83 \pm 0.05$ & $0.62 \pm 0.02$ & $0.92 \pm 0.01$ & $0.99 \pm 0.00$ & - & $00 \pm 0.02$ \\  
 \hline 
\multirow{3}{*}{Insurance} & Train Copy & $0.48 \pm 0.01$ & $1.00 \pm 0.00$ & $0.85 \pm 0.03$ & $0.96 \pm 0.01$ & $0.91 \pm 0.01$ & - & - \\  
 \cline{2-9} 
 & CTGAN & $0.67 \pm 0.03$ & $0.92 \pm 0.01$ & \textcolor{red}{$0.71 \pm 0.01$ } & \textcolor{ForestGreen}{\bm{$0.94 \pm 0.00$} } & $0.87 \pm 0.02$ & $265 \pm 23.11$ & $00 \pm 0.00$ \\  
 \cline{2-9} 
 & TVAE & $0.67 \pm 0.02$ & \textcolor{red}{$0.92 \pm 0.01$ } & $0.77 \pm 0.01$ & \textcolor{red}{$0.91 \pm 0.01$ } & \textcolor{red}{$0.86 \pm 0.02$ } & \textcolor{ForestGreen}{\bm{$32 \pm 4.50$} } & \textcolor{ForestGreen}{\bm{$00 \pm 0.00$} } \\  
 \cline{2-9} 
 & TabDDPM & $0.68 \pm 0.02$ & $0.62 \pm 0.01$ & \textcolor{ForestGreen}{\bm{$0.83 \pm 0.03$} } & $0.93 \pm 0.02$ & \textcolor{ForestGreen}{\bm{$0.89 \pm 0.02$} } & $181 \pm 18.27$ & \textcolor{red}{$03 \pm 1.01$ } \\  
 \cline{2-9} 
 & TabSyn & \textcolor{ForestGreen}{\bm{$0.60 \pm 0.02$} } & \textcolor{ForestGreen}{\bm{$0.61 \pm 0.02$} } & $0.82 \pm 0.00$ & $0.94 \pm 0.01$ & $0.88 \pm 0.00$ & \textcolor{red}{$1156 \pm 74.54$ } & $00 \pm 0.03$ \\  
 \cline{2-9} 
 & SMOTE & \textcolor{red}{$0.68 \pm 0.01$ } & $0.80 \pm 0.03$ & $0.81 \pm 0.01$ & $0.94 \pm 0.01$ & $0.87 \pm 0.01$ & - & $00 \pm 0.01$ \\  
 \cline{2-9} 
 & UC-SMOTE & $0.67 \pm 0.01$ & $0.85 \pm 0.02$ & $0.80 \pm 0.02$ & $0.93 \pm 0.02$ & $0.87 \pm 0.02$ & - & $00 \pm 0.00$ \\  
 \hline
 \pagebreak
\multirow{3}{*}{King} & Train Copy & $0.50 \pm 0.01$ & $1.00 \pm 0.00$ & $0.87 \pm 0.00$ & $0.98 \pm 0.01$ & $0.97 \pm 0.01$ & - & - \\  
 \cline{2-9} 
 & CTGAN & \textcolor{ForestGreen}{\bm{$0.93 \pm 0.01$} } & $0.62 \pm 0.00$ & $0.72 \pm 0.04$ & \textcolor{ForestGreen}{\bm{$0.97 \pm 0.01$} } & $0.94 \pm 0.01$ & $1805 \pm 8.66$ & $00 \pm 0.01$ \\  
 \cline{2-9} 
 & TVAE & $0.94 \pm 0.01$ & \textcolor{ForestGreen}{\bm{$0.61 \pm 0.01$} } & $0.81 \pm 0.02$ & $0.94 \pm 0.01$ & $0.94 \pm 0.01$ & $990 \pm 6.97$ & \textcolor{ForestGreen}{\bm{$00 \pm 0.00$} } \\  
 \cline{2-9} 
 & TabDDPM & $0.97 \pm 0.01$ & $0.73 \pm 0.06$ & $0.64 \pm 0.20$ & \textcolor{red}{$0.71 \pm 0.06$ } & \textcolor{red}{$0.83 \pm 0.01$ } & \textcolor{ForestGreen}{\bm{$244 \pm 14.03$} } & \textcolor{red}{$06 \pm 0.61$ } \\  
 \cline{2-9} 
 & TabSyn & $0.93 \pm 0.01$ & $0.62 \pm 0.00$ & \textcolor{red}{$0.12 \pm 0.03$ } & $0.96 \pm 0.01$ & $0.94 \pm 0.00$ & \textcolor{red}{$2717 \pm 648.93$ } & $01 \pm 0.01$ \\  
 \cline{2-9} 
 & SMOTE & \textcolor{red}{$0.98 \pm 0.00$ } & \textcolor{red}{$0.97 \pm 0.00$ } & \textcolor{ForestGreen}{\bm{$0.83 \pm 0.01$} } & $0.92 \pm 0.01$ & $0.95 \pm 0.01$ & - & $02 \pm 0.23$ \\  
 \cline{2-9} 
 & UC-SMOTE & $0.98 \pm 0.00$ & $0.96 \pm 0.01$ & $0.82 \pm 0.03$ & $0.92 \pm 0.01$ & \textcolor{ForestGreen}{\bm{$0.96 \pm 0.01$} } & - & $00 \pm 0.03$ \\  
 \hline 
\multirow{3}{*}{\shortstack{Miniboo\\ ne}} & Train Copy & $0.50 \pm 0.00$ & $1.00 \pm 0.00$ & $0.89 \pm 0.01$ & $0.99 \pm 0.00$ & $0.99 \pm 0.01$ & - & - \\  
 \cline{2-9} 
 & CTGAN & $0.91 \pm 0.02$ & $0.60 \pm 0.01$ & \textcolor{red}{$0.83 \pm 0.03$ } & $0.96 \pm 0.02$ & \textcolor{red}{$0.58 \pm 0.01$ } & $6879 \pm 2388.04$ & $00 \pm 0.06$ \\  
 \cline{2-9} 
 & TVAE & \textcolor{red}{$0.95 \pm 0.03$ } & \textcolor{ForestGreen}{\bm{$0.60 \pm 0.00$} } & $0.86 \pm 0.00$ & \textcolor{red}{$0.93 \pm 0.01$ } & $0.59 \pm 0.01$ & $4876 \pm 4730.59$ & \textcolor{ForestGreen}{\bm{$00 \pm 0.05$} } \\  
 \cline{2-9} 
 & TabDDPM & $0.64 \pm 0.02$ & $0.61 \pm 0.01$ & $0.89 \pm 0.00$ & \textcolor{ForestGreen}{\bm{$0.99 \pm 0.00$} } & $0.91 \pm 0.03$ & \textcolor{ForestGreen}{\bm{$193 \pm 46.01$} } & \textcolor{red}{$23 \pm 8.73$ } \\  
 \cline{2-9} 
 & TabSyn & \textcolor{ForestGreen}{\bm{$0.61 \pm 0.01$} } & $0.61 \pm 0.00$ & $0.89 \pm 0.00$ & $0.99 \pm 0.01$ & $0.91 \pm 0.03$ & \textcolor{red}{$7861 \pm 1700.48$ } & $07 \pm 0.08$ \\  
 \cline{2-9} 
 & SMOTE & $0.83 \pm 0.00$ & \textcolor{red}{$1.00 \pm 0.00$ } & \textcolor{ForestGreen}{\bm{$0.89 \pm 0.00$} } & $0.97 \pm 0.01$ & \textcolor{ForestGreen}{\bm{$0.99 \pm 0.01$} } & - & $01 \pm 0.10$ \\  
 \cline{2-9} 
 & UC-SMOTE & $0.83 \pm 0.00$ & $1.00 \pm 0.00$ & $0.89 \pm 0.00$ & $0.97 \pm 0.01$ & $0.98 \pm 0.01$ & - & $03 \pm 0.04$ \\  
 \hline 
\multirow{3}{*}{Moons} & Train Copy & $0.50 \pm 0.00$ & $1.00 \pm 0.00$ & $1.00 \pm 0.00$ & $0.99 \pm 0.00$ & $0.99 \pm 0.00$ & - & - \\  
 \cline{2-9} 
 & CTGAN & \textcolor{red}{$0.68 \pm 0.01$ } & $0.61 \pm 0.01$ & \textcolor{ForestGreen}{\bm{$1.00 \pm 0.00$} } & \textcolor{red}{$0.96 \pm 0.01$ } & \textcolor{red}{$0.97 \pm 0.01$ } & $2571 \pm 967.05$ & $00 \pm 0.02$ \\  
 \cline{2-9} 
 & TVAE & $0.63 \pm 0.01$ & $0.61 \pm 0.02$ & $1.00 \pm 0.00$ & $0.98 \pm 0.00$ & $0.97 \pm 0.00$ & $752 \pm 51.10$ & \textcolor{ForestGreen}{\bm{$00 \pm 0.01$} } \\  
 \cline{2-9} 
 & TabDDPM & \textcolor{ForestGreen}{\bm{$0.52 \pm 0.01$} } & $0.61 \pm 0.01$ & $1.00 \pm 0.00$ & $0.99 \pm 0.00$ & $0.99 \pm 0.00$ & \textcolor{ForestGreen}{\bm{$183 \pm 92.44$} } & \textcolor{red}{$09 \pm 6.74$ } \\  
 \cline{2-9} 
 & TabSyn & $0.54 \pm 0.01$ & \textcolor{ForestGreen}{\bm{$0.61 \pm 0.01$} } & $1.00 \pm 0.00$ & $0.99 \pm 0.00$ & $0.99 \pm 0.01$ & \textcolor{red}{$2647 \pm 656.79$ } & $01 \pm 0.81$ \\  
 \cline{2-9} 
 & SMOTE & $0.54 \pm 0.00$ & \textcolor{red}{$0.84 \pm 0.05$ } & $1.00 \pm 0.00$ & \textcolor{ForestGreen}{\bm{$0.99 \pm 0.00$} } & \textcolor{ForestGreen}{\bm{$0.99 \pm 0.00$} } & - & $00 \pm 0.02$ \\  
 \cline{2-9} 
 & UC-SMOTE & $0.54 \pm 0.01$ & $0.80 \pm 0.03$ & $1.00 \pm 0.00$ & $0.99 \pm 0.00$ & $0.99 \pm 0.01$ & - & $00 \pm 0.03$ \\  
 \hline

 \caption{Results per dataset and model under diverse metrics. The training times (gradient descent) and sampling times are given in seconds.
 The sampling times are given for 5 samples.
 The best values per metric are formatted in \textcolor{ForestGreen}{\textbf{bold green}} and the worse values are in \textcolor{red}{red}.}
 \label{table:results}
 \end{longtable}
\end{tiny}

\section{Training and Sampling  Impact}
\label{appendix:tuning_cost}

We performed large-scale experiments including a quite costly hyperparameter tuning step. Note that the term "\textit{costs}" in this section refers to the duration, energy consumption, and CO$_2$ emissions. We evaluate the training and sampling costs of the models with their optimized hyperparameters, as well as the full hyperparameter tuning costs.

\subsection{Training and Sampling Costs}  
\label{appendix:impact}

Due to the massive nature of the experiments, the hyperparameter search could not all be run on the same hardware. We hence estimated the training (gradient descent) cost by running all models on a single Tesla V100-SXM2 32 GB.
Table~\ref{table:emissions_estimation_normalized} provides the raw training and sampling energy consumption and emissions for reference.
All of the TabSyn costs shown are a combination of the VAE cost and the denoiser cost, which are estimated separately and then added together. Also, all TabSyn costs on the \textit{Covertype} dataset are estimated considering the complete dataset size.

            

    \begin{tiny}\setlength\tabcolsep{4pt}\setstretch{1.5} %
    \begin{longtable}{|c|l|*{4}{l|}}
    
    \hline
    Dataset & Model & Emissions Training $\downarrow$ & Energy Training $\downarrow$ & Emissions Sampling $\downarrow$ & Energy Sampling $\downarrow$ \\
    \hline
    \endfirsthead
    
    \caption*{Tuning costs estimation for the best models (continued).}\\
    
    \hline
    Dataset & Model & Emissions Training $\downarrow$ & Energy Training $\downarrow$ & Emissions Sampling $\downarrow$ & Energy Sampling $\downarrow$ \\
     \hline
    \endhead
    
    \multirow{3}{*}{Abalone} & CTGAN & \textcolor{red}{$3.05\!\pm\!1.77\ \cdot{}10^{-3}$ } & \textcolor{red}{$4.53\!\pm\!2.62\ \cdot{}10^{-2}$ } & $9.74\!\pm\!2.37\ \cdot{}10^{-8}$ & $1.45\!\pm\!0.35\ \cdot{}10^{-6}$ \\  
 \cline{2-6} 
 & TabDDPM & \textcolor{ForestGreen}{ \bm{$9.16\!\pm\!1.42\ \cdot{}10^{-6}$} } & \textcolor{ForestGreen}{ \bm{$1.36\!\pm\!0.21\ \cdot{}10^{-4}$} } & \textcolor{red}{$2.90\!\pm\!0.50\ \cdot{}10^{-5}$ } & \textcolor{red}{$4.31\!\pm\!0.74\ \cdot{}10^{-4}$ } \\  
 \cline{2-6} 
 & TabSyn & $3.18\!\pm\!0.32\ \cdot{}10^{-4}$ & $4.27\!\pm\!0.32\ \cdot{}10^{-3}$ & $1.49\!\pm\!0.78\ \cdot{}10^{-6}$ & $2.21\!\pm\!1.16\ \cdot{}10^{-5}$ \\  
 \cline{2-6} 
 & TVAE & $5.06\!\pm\!0.13\ \cdot{}10^{-4}$ & $7.51\!\pm\!0.19\ \cdot{}10^{-3}$ & \textcolor{ForestGreen}{ \bm{$6.23\!\pm\!2.14\ \cdot{}10^{-8}$} } & \textcolor{ForestGreen}{ \bm{$9.24\!\pm\!3.17\ \cdot{}10^{-7}$} } \\  
 \cline{2-6} 
 & SMOTE & - & - & $4.04\!\pm\!0.35\ \cdot{}10^{-7}$ & $5.99\!\pm\!0.51\ \cdot{}10^{-6}$ \\  
 \cline{2-6} 
 & UC-SMOTE & - & - & $1.37\!\pm\!0.07\ \cdot{}10^{-7}$ & $2.03\!\pm\!0.11\ \cdot{}10^{-6}$ \\  
 \hline 
\multirow{3}{*}{Adult} & CTGAN & \textcolor{red}{$1.24\!\pm\!0.41\ \cdot{}10^{-2}$ } & \textcolor{red}{$1.84\!\pm\!0.61\ \cdot{}10^{-1}$ } & $9.67\!\pm\!0.41\ \cdot{}10^{-7}$ & $1.43\!\pm\!0.06\ \cdot{}10^{-5}$ \\  
 \cline{2-6} 
 & TabDDPM & \textcolor{ForestGreen}{ \bm{$2.68\!\pm\!0.35\ \cdot{}10^{-4}$} } & \textcolor{ForestGreen}{ \bm{$3.98\!\pm\!0.51\ \cdot{}10^{-3}$} } & \textcolor{red}{$8.41\!\pm\!2.01\ \cdot{}10^{-5}$ } & \textcolor{red}{$1.25\!\pm\!0.30\ \cdot{}10^{-3}$ } \\  
 \cline{2-6} 
 & TabSyn & $1.68\!\pm\!0.72\ \cdot{}10^{-3}$ & $1.42\!\pm\!0.36\ \cdot{}10^{-2}$ & $1.56\!\pm\!0.85\ \cdot{}10^{-5}$ & $2.31\!\pm\!1.26\ \cdot{}10^{-4}$ \\  
 \cline{2-6} 
 & TVAE & $6.30\!\pm\!0.01\ \cdot{}10^{-3}$ & $9.34\!\pm\!0.01\ \cdot{}10^{-2}$ & \textcolor{ForestGreen}{ \bm{$3.54\!\pm\!0.07\ \cdot{}10^{-7}$} } & \textcolor{ForestGreen}{ \bm{$5.25\!\pm\!0.11\ \cdot{}10^{-6}$} } \\  
 \cline{2-6} 
 & SMOTE & - & - & $2.46\!\pm\!0.04\ \cdot{}10^{-5}$ & $3.66\!\pm\!0.06\ \cdot{}10^{-4}$ \\  
 \cline{2-6} 
 & UC-SMOTE & - & - & $3.82\!\pm\!0.01\ \cdot{}10^{-5}$ & $5.67\!\pm\!0.01\ \cdot{}10^{-4}$ \\  
 \hline 
\multirow{3}{*}{\shortstack{Bank\\ marketing}} & CTGAN & \textcolor{red}{$1.41\!\pm\!0.03\ \cdot{}10^{-2}$ } & \textcolor{red}{$2.09\!\pm\!0.04\ \cdot{}10^{-1}$ } & $8.88\!\pm\!0.32\ \cdot{}10^{-7}$ & $1.32\!\pm\!0.05\ \cdot{}10^{-5}$ \\  
 \cline{2-6} 
 & TabDDPM & \textcolor{ForestGreen}{ \bm{$2.50\!\pm\!0.22\ \cdot{}10^{-4}$} } & \textcolor{ForestGreen}{ \bm{$3.71\!\pm\!0.33\ \cdot{}10^{-3}$} } & \textcolor{red}{$7.81\!\pm\!1.91\ \cdot{}10^{-5}$ } & \textcolor{red}{$1.16\!\pm\!0.28\ \cdot{}10^{-3}$ } \\  
 \cline{2-6} 
 & TabSyn & $1.06\!\pm\!0.08\ \cdot{}10^{-3}$ & $1.46\!\pm\!0.04\ \cdot{}10^{-2}$ & $1.80\!\pm\!0.02\ \cdot{}10^{-5}$ & $2.67\!\pm\!0.03\ \cdot{}10^{-4}$ \\  
 \cline{2-6} 
 & TVAE & $4.90\!\pm\!2.83\ \cdot{}10^{-3}$ & $7.27\!\pm\!4.19\ \cdot{}10^{-2}$ & \textcolor{ForestGreen}{ \bm{$2.84\!\pm\!0.47\ \cdot{}10^{-7}$} } & \textcolor{ForestGreen}{ \bm{$4.22\!\pm\!0.69\ \cdot{}10^{-6}$} } \\  
 \cline{2-6} 
 & SMOTE & - & - & $3.46\!\pm\!0.05\ \cdot{}10^{-5}$ & $5.13\!\pm\!0.08\ \cdot{}10^{-4}$ \\  
 \cline{2-6} 
 & UC-SMOTE & - & - & $3.32\!\pm\!0.01\ \cdot{}10^{-5}$ & $4.92\!\pm\!0.02\ \cdot{}10^{-4}$ \\  
 \hline 
\multirow{3}{*}{\shortstack{Bike\\ sharing}} & CTGAN & \textcolor{red}{$1.68\!\pm\!1.03\ \cdot{}10^{-2}$ } & \textcolor{red}{$2.50\!\pm\!1.52\ \cdot{}10^{-1}$ } & $5.79\!\pm\!2.48\ \cdot{}10^{-7}$ & $8.59\!\pm\!3.68\ \cdot{}10^{-6}$ \\  
 \cline{2-6} 
 & TabDDPM & \textcolor{ForestGreen}{ \bm{$8.73\!\pm\!0.70\ \cdot{}10^{-5}$} } & \textcolor{ForestGreen}{ \bm{$1.30\!\pm\!0.10\ \cdot{}10^{-3}$} } & \textcolor{red}{$6.04\!\pm\!0.72\ \cdot{}10^{-5}$ } & \textcolor{red}{$8.96\!\pm\!1.07\ \cdot{}10^{-4}$ } \\  
 \cline{2-6} 
 & TabSyn & $6.29\!\pm\!1.72\ \cdot{}10^{-4}$ & $8.11\!\pm\!0.27\ \cdot{}10^{-3}$ & $7.02\!\pm\!0.03\ \cdot{}10^{-6}$ & $1.04\!\pm\!0.00\ \cdot{}10^{-4}$ \\  
 \cline{2-6} 
 & TVAE & $2.42\!\pm\!0.07\ \cdot{}10^{-3}$ & $3.59\!\pm\!0.11\ \cdot{}10^{-2}$ & \textcolor{ForestGreen}{ \bm{$1.37\!\pm\!0.27\ \cdot{}10^{-7}$} } & \textcolor{ForestGreen}{ \bm{$2.03\!\pm\!0.40\ \cdot{}10^{-6}$} } \\  
 \cline{2-6} 
 & SMOTE & - & - & $4.53\!\pm\!0.22\ \cdot{}10^{-6}$ & $6.72\!\pm\!0.33\ \cdot{}10^{-5}$ \\  
 \cline{2-6} 
 & UC-SMOTE & - & - & $1.48\!\pm\!0.32\ \cdot{}10^{-6}$ & $2.20\!\pm\!0.48\ \cdot{}10^{-5}$ \\  
 \hline 
\multirow{3}{*}{\shortstack{Black\\ friday}} & CTGAN & \textcolor{red}{$4.55\!\pm\!3.60\ \cdot{}10^{-2}$ } & \textcolor{red}{$6.75\!\pm\!5.34\ \cdot{}10^{-1}$ } & $2.70\!\pm\!0.56\ \cdot{}10^{-6}$ & $4.01\!\pm\!0.83\ \cdot{}10^{-5}$ \\  
 \cline{2-6} 
 & TabDDPM & \textcolor{ForestGreen}{ \bm{$8.97\!\pm\!1.75\ \cdot{}10^{-4}$} } & \textcolor{ForestGreen}{ \bm{$1.33\!\pm\!0.26\ \cdot{}10^{-2}$} } & $2.98\!\pm\!1.09\ \cdot{}10^{-4}$ & $4.42\!\pm\!1.62\ \cdot{}10^{-3}$ \\  
 \cline{2-6} 
 & TabSyn & $2.78\!\pm\!0.26\ \cdot{}10^{-3}$ & $2.58\!\pm\!0.90\ \cdot{}10^{-2}$ & $5.18\!\pm\!2.81\ \cdot{}10^{-5}$ & $7.69\!\pm\!4.17\ \cdot{}10^{-4}$ \\  
 \cline{2-6} 
 & TVAE & $2.23\!\pm\!0.01\ \cdot{}10^{-2}$ & $3.31\!\pm\!0.01\ \cdot{}10^{-1}$ & \textcolor{ForestGreen}{ \bm{$8.59\!\pm\!0.65\ \cdot{}10^{-7}$} } & \textcolor{ForestGreen}{ \bm{$1.27\!\pm\!0.10\ \cdot{}10^{-5}$} } \\  
 \cline{2-6} 
 & SMOTE & - & - & $1.14\!\pm\!0.01\ \cdot{}10^{-4}$ & $1.68\!\pm\!0.01\ \cdot{}10^{-3}$ \\  
 \cline{2-6} 
 & UC-SMOTE & - & - & \textcolor{red}{$3.84\!\pm\!0.02\ \cdot{}10^{-4}$ } & \textcolor{red}{$5.70\!\pm\!0.03\ \cdot{}10^{-3}$ } \\  
 \hline 
\multirow{3}{*}{Cardio} & CTGAN & \textcolor{red}{$2.11\!\pm\!0.03\ \cdot{}10^{-2}$ } & \textcolor{red}{$3.13\!\pm\!0.04\ \cdot{}10^{-1}$ } & $1.30\!\pm\!0.03\ \cdot{}10^{-6}$ & $1.93\!\pm\!0.04\ \cdot{}10^{-5}$ \\  
 \cline{2-6} 
 & TabDDPM & \textcolor{ForestGreen}{ \bm{$2.22\!\pm\!1.13\ \cdot{}10^{-4}$} } & \textcolor{ForestGreen}{ \bm{$3.30\!\pm\!1.68\ \cdot{}10^{-3}$} } & \textcolor{red}{$1.05\!\pm\!0.70\ \cdot{}10^{-4}$ } & \textcolor{red}{$1.56\!\pm\!1.04\ \cdot{}10^{-3}$ } \\  
 \cline{2-6} 
 & TabSyn & $1.15\!\pm\!0.16\ \cdot{}10^{-3}$ & $1.67\!\pm\!0.03\ \cdot{}10^{-2}$ & $2.89\!\pm\!0.01\ \cdot{}10^{-5}$ & $4.29\!\pm\!0.02\ \cdot{}10^{-4}$ \\  
 \cline{2-6} 
 & TVAE & $7.64\!\pm\!4.39\ \cdot{}10^{-3}$ & $1.13\!\pm\!0.65\ \cdot{}10^{-1}$ & \textcolor{ForestGreen}{ \bm{$2.70\!\pm\!0.94\ \cdot{}10^{-7}$} } & \textcolor{ForestGreen}{ \bm{$4.00\!\pm\!1.40\ \cdot{}10^{-6}$} } \\  
 \cline{2-6} 
 & SMOTE & - & - & $2.27\!\pm\!0.05\ \cdot{}10^{-6}$ & $3.37\!\pm\!0.07\ \cdot{}10^{-5}$ \\  
 \cline{2-6} 
 & UC-SMOTE & - & - & $4.27\!\pm\!0.16\ \cdot{}10^{-6}$ & $6.34\!\pm\!0.24\ \cdot{}10^{-5}$ \\  
 \hline 
\multirow{3}{*}{Churn} & CTGAN & \textcolor{red}{$1.15\!\pm\!0.43\ \cdot{}10^{-2}$ } & \textcolor{red}{$1.71\!\pm\!0.63\ \cdot{}10^{-1}$ } & $6.61\!\pm\!1.15\ \cdot{}10^{-7}$ & $9.81\!\pm\!1.71\ \cdot{}10^{-6}$ \\  
 \cline{2-6} 
 & TabDDPM & \textcolor{ForestGreen}{ \bm{$1.10\!\pm\!0.07\ \cdot{}10^{-4}$} } & \textcolor{ForestGreen}{ \bm{$1.63\!\pm\!0.10\ \cdot{}10^{-3}$} } & \textcolor{red}{$1.96\!\pm\!0.17\ \cdot{}10^{-4}$ } & \textcolor{red}{$2.90\!\pm\!0.25\ \cdot{}10^{-3}$ } \\  
 \cline{2-6} 
 & TabSyn & $4.97\!\pm\!0.79\ \cdot{}10^{-4}$ & $6.25\!\pm\!0.62\ \cdot{}10^{-3}$ & $3.43\!\pm\!1.81\ \cdot{}10^{-6}$ & $5.09\!\pm\!2.68\ \cdot{}10^{-5}$ \\  
 \cline{2-6} 
 & TVAE & $1.34\!\pm\!0.03\ \cdot{}10^{-3}$ & $1.98\!\pm\!0.05\ \cdot{}10^{-2}$ & \textcolor{ForestGreen}{ \bm{$3.08\!\pm\!0.29\ \cdot{}10^{-7}$} } & \textcolor{ForestGreen}{ \bm{$4.57\!\pm\!0.42\ \cdot{}10^{-6}$} } \\  
 \cline{2-6} 
 & SMOTE & - & - & $4.13\!\pm\!0.20\ \cdot{}10^{-6}$ & $6.12\!\pm\!0.29\ \cdot{}10^{-5}$ \\  
 \cline{2-6} 
 & UC-SMOTE & - & - & $3.53\!\pm\!0.19\ \cdot{}10^{-6}$ & $5.24\!\pm\!0.28\ \cdot{}10^{-5}$ \\  
 \hline 
 \pagebreak
\multirow{3}{*}{Covertype} & CTGAN & \textcolor{red}{$2.33\!\pm\!0.78\ \cdot{}10^{-1}$ } & \textcolor{red}{$3.46\!\pm\!1.15$ } & $1.92\!\pm\!0.27\ \cdot{}10^{-5}$ & $2.85\!\pm\!0.41\ \cdot{}10^{-4}$ \\  
 \cline{2-6} 
 & TabDDPM & $6.22\!\pm\!0.35\ \cdot{}10^{-3}$ & $9.23\!\pm\!0.53\ \cdot{}10^{-2}$ & \textcolor{red}{$1.80\!\pm\!0.26\ \cdot{}10^{-3}$ } & \textcolor{red}{$2.67\!\pm\!0.38\ \cdot{}10^{-2}$ } \\  
 \cline{2-6} 
 & TabSyn & \textcolor{ForestGreen}{ \bm{$1.82\!\pm\!0.68\ \cdot{}10^{-3}$} } & \textcolor{ForestGreen}{ \bm{$1.26\!\pm\!0.72\ \cdot{}10^{-2}$} } & $2.37\!\pm\!0.04\ \cdot{}10^{-5}$ & $3.52\!\pm\!0.06\ \cdot{}10^{-4}$ \\  
 \cline{2-6} 
 & TVAE & $1.04\!\pm\!0.81\ \cdot{}10^{-1}$ & $1.54\!\pm\!1.21$ & \textcolor{ForestGreen}{ \bm{$8.00\!\pm\!1.93\ \cdot{}10^{-6}$} } & \textcolor{ForestGreen}{ \bm{$1.19\!\pm\!0.29\ \cdot{}10^{-4}$} } \\  
 \cline{2-6} 
 & SMOTE & - & - & $4.52\!\pm\!0.02\ \cdot{}10^{-4}$ & $6.71\!\pm\!0.04\ \cdot{}10^{-3}$ \\  
 \cline{2-6} 
 & UC-SMOTE & - & - & $5.01\!\pm\!0.03\ \cdot{}10^{-4}$ & $7.43\!\pm\!0.04\ \cdot{}10^{-3}$ \\  
 \hline 
\multirow{3}{*}{Diamonds} & CTGAN & \textcolor{red}{$1.52\!\pm\!0.00\ \cdot{}10^{-2}$ } & \textcolor{red}{$2.26\!\pm\!0.01\ \cdot{}10^{-1}$ } & $8.44\!\pm\!0.13\ \cdot{}10^{-7}$ & $1.25\!\pm\!0.02\ \cdot{}10^{-5}$ \\  
 \cline{2-6} 
 & TabDDPM & \textcolor{ForestGreen}{ \bm{$3.33\!\pm\!0.71\ \cdot{}10^{-4}$} } & \textcolor{ForestGreen}{ \bm{$4.93\!\pm\!1.06\ \cdot{}10^{-3}$} } & \textcolor{red}{$1.30\!\pm\!0.47\ \cdot{}10^{-4}$ } & \textcolor{red}{$1.92\!\pm\!0.70\ \cdot{}10^{-3}$ } \\  
 \cline{2-6} 
 & TabSyn & $1.21\!\pm\!0.18\ \cdot{}10^{-3}$ & $1.36\!\pm\!0.49\ \cdot{}10^{-2}$ & $1.67\!\pm\!0.89\ \cdot{}10^{-5}$ & $2.48\!\pm\!1.32\ \cdot{}10^{-4}$ \\  
 \cline{2-6} 
 & TVAE & $4.93\!\pm\!2.84\ \cdot{}10^{-3}$ & $7.31\!\pm\!4.21\ \cdot{}10^{-2}$ & \textcolor{ForestGreen}{ \bm{$2.42\!\pm\!0.76\ \cdot{}10^{-7}$} } & \textcolor{ForestGreen}{ \bm{$3.59\!\pm\!1.12\ \cdot{}10^{-6}$} } \\  
 \cline{2-6} 
 & SMOTE & - & - & $1.32\!\pm\!0.01\ \cdot{}10^{-5}$ & $1.96\!\pm\!0.01\ \cdot{}10^{-4}$ \\  
 \cline{2-6} 
 & UC-SMOTE & - & - & $3.67\!\pm\!0.14\ \cdot{}10^{-6}$ & $5.44\!\pm\!0.21\ \cdot{}10^{-5}$ \\  
 \hline 
\multirow{3}{*}{Heloc} & CTGAN & \textcolor{red}{$1.93\!\pm\!1.51\ \cdot{}10^{-2}$ } & \textcolor{red}{$2.87\!\pm\!2.24\ \cdot{}10^{-1}$ } & $6.03\!\pm\!4.19\ \cdot{}10^{-7}$ & $8.95\!\pm\!6.22\ \cdot{}10^{-6}$ \\  
 \cline{2-6} 
 & TabDDPM & \textcolor{ForestGreen}{ \bm{$1.21\!\pm\!0.29\ \cdot{}10^{-5}$} } & \textcolor{ForestGreen}{ \bm{$1.80\!\pm\!0.43\ \cdot{}10^{-4}$} } & \textcolor{red}{$1.50\!\pm\!0.42\ \cdot{}10^{-5}$ } & \textcolor{red}{$2.23\!\pm\!0.63\ \cdot{}10^{-4}$ } \\  
 \cline{2-6} 
 & TabSyn & $5.78\!\pm\!0.66\ \cdot{}10^{-4}$ & $6.91\!\pm\!0.18\ \cdot{}10^{-3}$ & $4.60\!\pm\!0.11\ \cdot{}10^{-6}$ & $6.83\!\pm\!0.16\ \cdot{}10^{-5}$ \\  
 \cline{2-6} 
 & TVAE & $1.88\!\pm\!1.11\ \cdot{}10^{-3}$ & $2.80\!\pm\!1.65\ \cdot{}10^{-2}$ & \textcolor{ForestGreen}{ \bm{$1.15\!\pm\!0.43\ \cdot{}10^{-7}$} } & \textcolor{ForestGreen}{ \bm{$1.71\!\pm\!0.63\ \cdot{}10^{-6}$} } \\  
 \cline{2-6} 
 & SMOTE & - & - & $9.32\!\pm\!1.28\ \cdot{}10^{-7}$ & $1.38\!\pm\!0.19\ \cdot{}10^{-5}$ \\  
 \cline{2-6} 
 & UC-SMOTE & - & - & $2.02\!\pm\!0.07\ \cdot{}10^{-6}$ & $3.00\!\pm\!0.10\ \cdot{}10^{-5}$ \\  
 \hline 
\multirow{3}{*}{Higgs} & CTGAN & \textcolor{red}{$5.71\!\pm\!1.88\ \cdot{}10^{-2}$ } & \textcolor{red}{$8.47\!\pm\!2.79\ \cdot{}10^{-1}$ } & $2.92\!\pm\!0.56\ \cdot{}10^{-6}$ & $4.33\!\pm\!0.84\ \cdot{}10^{-5}$ \\  
 \cline{2-6} 
 & TabDDPM & \textcolor{ForestGreen}{ \bm{$4.01\!\pm\!0.25\ \cdot{}10^{-4}$} } & \textcolor{ForestGreen}{ \bm{$5.96\!\pm\!0.37\ \cdot{}10^{-3}$} } & \textcolor{red}{$1.77\!\pm\!0.18\ \cdot{}10^{-4}$ } & \textcolor{red}{$2.63\!\pm\!0.27\ \cdot{}10^{-3}$ } \\  
 \cline{2-6} 
 & TabSyn & $2.88\!\pm\!1.12\ \cdot{}10^{-3}$ & $2.07\!\pm\!0.74\ \cdot{}10^{-2}$ & $3.20\!\pm\!1.72\ \cdot{}10^{-5}$ & $4.75\!\pm\!2.55\ \cdot{}10^{-4}$ \\  
 \cline{2-6} 
 & TVAE & $2.57\!\pm\!0.05\ \cdot{}10^{-2}$ & $3.81\!\pm\!0.08\ \cdot{}10^{-1}$ & \textcolor{ForestGreen}{ \bm{$1.06\!\pm\!0.34\ \cdot{}10^{-6}$} } & \textcolor{ForestGreen}{ \bm{$1.58\!\pm\!0.51\ \cdot{}10^{-5}$} } \\  
 \cline{2-6} 
 & SMOTE & - & - & $2.01\!\pm\!0.19\ \cdot{}10^{-6}$ & $2.98\!\pm\!0.28\ \cdot{}10^{-5}$ \\  
 \cline{2-6} 
 & UC-SMOTE & - & - & $6.19\!\pm\!0.13\ \cdot{}10^{-6}$ & $9.18\!\pm\!0.20\ \cdot{}10^{-5}$ \\  
 \hline 
\multirow{3}{*}{\shortstack{House\\ 16h}} & CTGAN & \textcolor{red}{$4.42\!\pm\!2.48\ \cdot{}10^{-3}$ } & \textcolor{red}{$6.55\!\pm\!3.68\ \cdot{}10^{-2}$ } & \textcolor{ForestGreen}{ \bm{$1.24\!\pm\!0.16\ \cdot{}10^{-7}$} } & \textcolor{ForestGreen}{ \bm{$1.85\!\pm\!0.24\ \cdot{}10^{-6}$} } \\  
 \cline{2-6} 
 & TabDDPM & \textcolor{ForestGreen}{ \bm{$3.81\!\pm\!1.14\ \cdot{}10^{-5}$} } & \textcolor{ForestGreen}{ \bm{$5.66\!\pm\!1.69\ \cdot{}10^{-4}$} } & \textcolor{red}{$2.03\!\pm\!0.70\ \cdot{}10^{-5}$ } & \textcolor{red}{$3.01\!\pm\!1.04\ \cdot{}10^{-4}$ } \\  
 \cline{2-6} 
 & TabSyn & $8.56\!\pm\!2.73\ \cdot{}10^{-4}$ & $9.40\!\pm\!1.12\ \cdot{}10^{-3}$ & $9.68\!\pm\!0.08\ \cdot{}10^{-6}$ & $1.44\!\pm\!0.01\ \cdot{}10^{-4}$ \\  
 \cline{2-6} 
 & TVAE & $4.12\!\pm\!0.02\ \cdot{}10^{-3}$ & $6.11\!\pm\!0.03\ \cdot{}10^{-2}$ & $1.40\!\pm\!0.17\ \cdot{}10^{-7}$ & $2.08\!\pm\!0.26\ \cdot{}10^{-6}$ \\  
 \cline{2-6} 
 & SMOTE & - & - & $2.92\!\pm\!1.19\ \cdot{}10^{-7}$ & $4.34\!\pm\!1.77\ \cdot{}10^{-6}$ \\  
 \cline{2-6} 
 & UC-SMOTE & - & - & $4.04\!\pm\!0.48\ \cdot{}10^{-7}$ & $5.99\!\pm\!0.72\ \cdot{}10^{-6}$ \\  
 \hline 
\multirow{3}{*}{Insurance} & CTGAN & \textcolor{red}{$1.18\!\pm\!0.10\ \cdot{}10^{-3}$ } & \textcolor{red}{$1.75\!\pm\!0.15\ \cdot{}10^{-2}$ } & $7.40\!\pm\!1.62\ \cdot{}10^{-8}$ & $1.10\!\pm\!0.24\ \cdot{}10^{-6}$ \\  
 \cline{2-6} 
 & TabDDPM & \textcolor{ForestGreen}{ \bm{$2.83\!\pm\!0.29\ \cdot{}10^{-6}$} } & \textcolor{ForestGreen}{ \bm{$4.20\!\pm\!0.43\ \cdot{}10^{-5}$} } & \textcolor{red}{$2.65\!\pm\!0.94\ \cdot{}10^{-5}$ } & \textcolor{red}{$3.93\!\pm\!1.40\ \cdot{}10^{-4}$ } \\  
 \cline{2-6} 
 & TabSyn & $3.45\!\pm\!0.04\ \cdot{}10^{-4}$ & $4.64\!\pm\!0.36\ \cdot{}10^{-3}$ & $5.87\!\pm\!2.03\ \cdot{}10^{-7}$ & $8.71\!\pm\!3.01\ \cdot{}10^{-6}$ \\  
 \cline{2-6} 
 & TVAE & $1.44\!\pm\!0.20\ \cdot{}10^{-4}$ & $2.14\!\pm\!0.29\ \cdot{}10^{-3}$ & \textcolor{ForestGreen}{ \bm{$4.70\!\pm\!1.02\ \cdot{}10^{-8}$} } & \textcolor{ForestGreen}{ \bm{$6.98\!\pm\!1.51\ \cdot{}10^{-7}$} } \\  
 \cline{2-6} 
 & SMOTE & - & - & $1.83\!\pm\!0.26\ \cdot{}10^{-7}$ & $2.71\!\pm\!0.39\ \cdot{}10^{-6}$ \\  
 \cline{2-6} 
 & UC-SMOTE & - & - & $1.02\!\pm\!0.04\ \cdot{}10^{-7}$ & $1.51\!\pm\!0.06\ \cdot{}10^{-6}$ \\  
 \hline 
\multirow{3}{*}{King} & CTGAN & \textcolor{red}{$8.15\!\pm\!0.04\ \cdot{}10^{-3}$ } & \textcolor{red}{$1.21\!\pm\!0.01\ \cdot{}10^{-1}$ } & $5.72\!\pm\!0.34\ \cdot{}10^{-7}$ & $8.49\!\pm\!0.50\ \cdot{}10^{-6}$ \\  
 \cline{2-6} 
 & TabDDPM & \textcolor{ForestGreen}{ \bm{$9.27\!\pm\!0.56\ \cdot{}10^{-5}$} } & \textcolor{ForestGreen}{ \bm{$1.38\!\pm\!0.08\ \cdot{}10^{-3}$} } & \textcolor{red}{$5.08\!\pm\!0.49\ \cdot{}10^{-5}$ } & \textcolor{red}{$7.55\!\pm\!0.73\ \cdot{}10^{-4}$ } \\  
 \cline{2-6} 
 & TabSyn & $9.40\!\pm\!1.26\ \cdot{}10^{-4}$ & $9.57\!\pm\!0.35\ \cdot{}10^{-3}$ & $9.15\!\pm\!0.02\ \cdot{}10^{-6}$ & $1.36\!\pm\!0.00\ \cdot{}10^{-4}$ \\  
 \cline{2-6} 
 & TVAE & $4.43\!\pm\!0.04\ \cdot{}10^{-3}$ & $6.57\!\pm\!0.07\ \cdot{}10^{-2}$ & \textcolor{ForestGreen}{ \bm{$2.59\!\pm\!0.18\ \cdot{}10^{-7}$} } & \textcolor{ForestGreen}{ \bm{$3.85\!\pm\!0.26\ \cdot{}10^{-6}$} } \\  
 \cline{2-6} 
 & SMOTE & - & - & $8.77\!\pm\!0.84\ \cdot{}10^{-6}$ & $1.30\!\pm\!0.12\ \cdot{}10^{-4}$ \\  
 \cline{2-6} 
 & UC-SMOTE & - & - & $2.25\!\pm\!0.10\ \cdot{}10^{-6}$ & $3.34\!\pm\!0.14\ \cdot{}10^{-5}$ \\  
 \hline 
 \pagebreak
\multirow{3}{*}{\shortstack{Miniboo\\ ne}} & CTGAN & \textcolor{red}{$3.16\!\pm\!1.09\ \cdot{}10^{-2}$ } & \textcolor{red}{$4.69\!\pm\!1.62\ \cdot{}10^{-1}$ } & $3.50\!\pm\!0.24\ \cdot{}10^{-6}$ & $5.20\!\pm\!0.35\ \cdot{}10^{-5}$ \\  
 \cline{2-6} 
 & TabDDPM & \textcolor{ForestGreen}{ \bm{$4.50\!\pm\!1.33\ \cdot{}10^{-4}$} } & \textcolor{ForestGreen}{ \bm{$6.68\!\pm\!1.97\ \cdot{}10^{-3}$} } & \textcolor{red}{$1.84\!\pm\!0.69\ \cdot{}10^{-4}$ } & \textcolor{red}{$2.74\!\pm\!1.03\ \cdot{}10^{-3}$ } \\  
 \cline{2-6} 
 & TabSyn & $3.41\!\pm\!0.61\ \cdot{}10^{-3}$ & $3.18\!\pm\!0.79\ \cdot{}10^{-2}$ & $5.57\!\pm\!0.07\ \cdot{}10^{-5}$ & $8.26\!\pm\!0.11\ \cdot{}10^{-4}$ \\  
 \cline{2-6} 
 & TVAE & $2.18\!\pm\!2.12\ \cdot{}10^{-2}$ & $3.24\!\pm\!3.14\ \cdot{}10^{-1}$ & \textcolor{ForestGreen}{ \bm{$1.94\!\pm\!0.22\ \cdot{}10^{-6}$} } & \textcolor{ForestGreen}{ \bm{$2.88\!\pm\!0.32\ \cdot{}10^{-5}$} } \\  
 \cline{2-6} 
 & SMOTE & - & - & $4.39\!\pm\!0.34\ \cdot{}10^{-6}$ & $6.51\!\pm\!0.51\ \cdot{}10^{-5}$ \\  
 \cline{2-6} 
 & UC-SMOTE & - & - & $1.32\!\pm\!0.01\ \cdot{}10^{-5}$ & $1.95\!\pm\!0.02\ \cdot{}10^{-4}$ \\  
 \hline 
\multirow{3}{*}{Moons} & CTGAN & \textcolor{red}{$1.15\!\pm\!0.43\ \cdot{}10^{-2}$ } & \textcolor{red}{$1.71\!\pm\!0.64\ \cdot{}10^{-1}$ } & $5.02\!\pm\!1.07\ \cdot{}10^{-7}$ & $7.45\!\pm\!1.59\ \cdot{}10^{-6}$ \\  
 \cline{2-6} 
 & TabDDPM & \textcolor{ForestGreen}{ \bm{$1.31\!\pm\!0.77\ \cdot{}10^{-4}$} } & \textcolor{ForestGreen}{ \bm{$1.94\!\pm\!1.14\ \cdot{}10^{-3}$} } & \textcolor{red}{$7.39\!\pm\!5.33\ \cdot{}10^{-5}$ } & \textcolor{red}{$1.10\!\pm\!0.79\ \cdot{}10^{-3}$ } \\  
 \cline{2-6} 
 & TabSyn & $9.26\!\pm\!2.41\ \cdot{}10^{-4}$ & $1.14\!\pm\!0.36\ \cdot{}10^{-2}$ & $1.22\!\pm\!0.65\ \cdot{}10^{-5}$ & $1.81\!\pm\!0.96\ \cdot{}10^{-4}$ \\  
 \cline{2-6} 
 & TVAE & $3.37\!\pm\!0.24\ \cdot{}10^{-3}$ & $5.00\!\pm\!0.36\ \cdot{}10^{-2}$ & \textcolor{ForestGreen}{ \bm{$1.78\!\pm\!0.43\ \cdot{}10^{-7}$} } & \textcolor{ForestGreen}{ \bm{$2.65\!\pm\!0.64\ \cdot{}10^{-6}$} } \\  
 \cline{2-6} 
 & SMOTE & - & - & $2.44\!\pm\!0.52\ \cdot{}10^{-7}$ & $3.63\!\pm\!0.78\ \cdot{}10^{-6}$ \\  
 \cline{2-6} 
 & UC-SMOTE & - & - & $1.03\!\pm\!0.12\ \cdot{}10^{-6}$ & $1.53\!\pm\!0.18\ \cdot{}10^{-5}$ \\  
 \hline

 \caption{CO$_{2}$ emissions (in Kg) and Energy Consumption (in kWh) of the benchmark challengers. The energy consumption is obtained by summing the CPU, GPU and RAM energy. Training costs are given for all models on the same basis of $400$ epochs. Sampling costs are given for 5 samples. The best values are formatted in \textcolor{ForestGreen}{\textbf{bold green}} and the worse are in \textcolor{red}{red}.}
 \label{table:emissions_estimation_normalized}
 \end{longtable}
\end{tiny}

\subsection{Whole Tuning Cost Estimation}    
\label{sec:tuning_cost}

As mentioned in Section~\ref{sec:costest} we could not perform the hyperparameter search and training phases on a uniform hardware and software architecture and we estimated the tuning cost with Equation~\eqref{eq:costest}.

Each trial can be stopped based on three conditions: an early stopping decided by the model, a poor C2ST performance or a time limit. Therefore, to get an accurate estimate of the init-cost and the cost-per-step from the single-GPU mentioned in \ref{appendix:impact}, we needed to extract from our logs the effective number of training steps performed per trial.

In addition, we also considered the parallelization scheme applied during the tuning procedure. One issue with TabSyn  was that we observed a general slowdown when the model was parallelized too heavily. This issue was even more marked on datasets with a large number of columns.
We hence reduced the number trials per GPU and the global number of trials to $100$ to fall back to a reasonable time for this model.

Finally, we measured cost on a typical configuration used during our tuning experiment: 8 Tesla V100-SXM2 32 GB. %
Considering those 8 GPUs, we used the following parallel allocation of trials: $64$ for TVAE and CTGAN, $40$ for TabDDPM, and $16$ for TabSyn. A model that can be easily parallelized during the hyperparameter tuning phase offers a cost advantage. It is hence important to consider this aspect during the evaluation process. The results are presented in Table \ref{table:tuning_cost}.

            

            \begin{tiny}\setlength\tabcolsep{4pt}\setstretch{1.5} %
            \begin{longtable}{|c|l|*{3}{l|}}
            
            \hline
            \multirow{2}{*}{Dataset} & \multirow{2}{*}{Model} & \multicolumn{3}{c|}{Metrics} \\
                                            \cline{3-5}
                                            && \makecell{Energy \\ (kWh)} $\downarrow$ & \makecell{Emissions \\ (Kg)} $\downarrow$ & \makecell{ Duration \\ (HH:MM)} $\downarrow$ \\
            \hline
            \endfirsthead
            
            \caption*{Tuning costs estimation (continued).}\\
            
            \hline
            \multirow{2}{*}{Dataset} & \multirow{2}{*}{Model} & \multicolumn{3}{c|}{Metrics} \\
                                            \cline{3-5}
                                            && \makecell{Energy \\ (kWh)} $\downarrow$ & \makecell{Emissions \\ (Kg)} $\downarrow$ & \makecell{ Duration \\ (HH:MM)} $\downarrow$ \\
             \hline
            \endhead
                \multirow{3}{*}{Abalone} & TVAE & 0.00 & 0.01 & 00:02 \\  
 \cline{2-5} 
 & CTGAN & 0.01 & 0.08 & 00:19 \\  
 \cline{2-5} 
 & TabDDPM & \textcolor{ForestGreen}{\textbf{0.00} } & 0.05 & 00:10 \\  
 \cline{2-5} 
 & TabSyn & \textcolor{red}{0.03 } & \textcolor{red}{0.40 } & \textcolor{red}{01:31 } \\  
 \cline{2-5} 
 & SMOTE & 0.00 & \textcolor{ForestGreen}{\textbf{0.00} } & 00:00 \\  
 \cline{2-5} 
 & UC-SMOTE & 0.00 & 0.00 & \textcolor{ForestGreen}{\textbf{00:00} } \\  
 \hline 
\multirow{3}{*}{Adult} & TVAE & 0.01 & 0.17 & 00:42 \\  
 \cline{2-5} 
 & CTGAN & 0.02 & 0.30 & 01:15 \\  
 \cline{2-5} 
 & TabDDPM & 0.01 & 0.12 & 00:25 \\  
 \cline{2-5} 
 & TabSyn & \textcolor{red}{0.08 } & \textcolor{red}{0.91 } & \textcolor{red}{03:27 } \\  
 \cline{2-5} 
 & SMOTE & \textcolor{ForestGreen}{\textbf{0.00} } & \textcolor{ForestGreen}{\textbf{0.01} } & \textcolor{ForestGreen}{\textbf{00:04} } \\  
 \cline{2-5} 
 & UC-SMOTE & 0.00 & 0.02 & 00:06 \\  
 \hline 
\multirow{3}{*}{\shortstack{Bank\\ marketing}} & TVAE & 0.01 & 0.10 & 00:25 \\  
 \cline{2-5} 
 & CTGAN & 0.03 & 0.38 & 01:35 \\  
 \cline{2-5} 
 & TabDDPM & 0.01 & 0.13 & 00:29 \\  
 \cline{2-5} 
 & TabSyn & \textcolor{red}{0.08 } & \textcolor{red}{0.96 } & \textcolor{red}{03:27 } \\  
 \cline{2-5} 
 & SMOTE & \textcolor{ForestGreen}{\textbf{0.00} } & \textcolor{ForestGreen}{\textbf{0.02} } & 00:06 \\  
 \cline{2-5} 
 & UC-SMOTE & 0.00 & 0.02 & \textcolor{ForestGreen}{\textbf{00:05} } \\  
 \hline 
\multirow{3}{*}{\shortstack{Bike\\ sharing}} & TVAE & \textcolor{ForestGreen}{\textbf{0.00} } & 0.05 & 00:12 \\  
 \cline{2-5} 
 & CTGAN & 0.03 & 0.40 & 01:41 \\  
 \cline{2-5} 
 & TabDDPM & 0.01 & 0.12 & 00:22 \\  
 \cline{2-5} 
 & TabSyn & \textcolor{red}{0.05 } & \textcolor{red}{0.61 } & \textcolor{red}{02:18 } \\  
 \cline{2-5} 
 & SMOTE & 0.00 & \textcolor{ForestGreen}{\textbf{0.00} } & 00:00 \\  
 \cline{2-5} 
 & UC-SMOTE & 0.00 & 0.00 & \textcolor{ForestGreen}{\textbf{00:00} } \\  
 \hline 
\multirow{3}{*}{Black friday} & TVAE & 0.03 & 0.46 & 01:55 \\  
 \cline{2-5} 
 & CTGAN & 0.08 & 1.18 & 04:54 \\  
 \cline{2-5} 
 & TabDDPM & 0.01 & 0.13 & 00:23 \\  
 \cline{2-5} 
 & TabSyn & \textcolor{red}{0.21 } & \textcolor{red}{2.40 } & \textcolor{red}{09:19 } \\  
 \cline{2-5} 
 & SMOTE & \textcolor{ForestGreen}{\textbf{0.00} } & \textcolor{ForestGreen}{\textbf{0.06} } & \textcolor{ForestGreen}{\textbf{00:19} } \\  
 \cline{2-5} 
 & UC-SMOTE & 0.01 & 0.22 & 01:07 \\  
 \hline 
\multirow{3}{*}{Cardio} & TVAE & 0.01 & 0.16 & 00:41 \\  
 \cline{2-5} 
 & CTGAN & 0.03 & 0.51 & 02:07 \\  
 \cline{2-5} 
 & TabDDPM & 0.01 & 0.08 & 00:12 \\  
 \cline{2-5} 
 & TabSyn & \textcolor{red}{0.10 } & \textcolor{red}{1.38 } & \textcolor{red}{04:13 } \\  
 \cline{2-5} 
 & SMOTE & \textcolor{ForestGreen}{\textbf{0.00} } & \textcolor{ForestGreen}{\textbf{0.00} } & \textcolor{ForestGreen}{\textbf{00:00} } \\  
 \cline{2-5} 
 & UC-SMOTE & 0.00 & 0.00 & 00:00 \\  
 \hline 
\multirow{3}{*}{Churn} & TVAE & \textcolor{ForestGreen}{\textbf{0.00} } & 0.03 & 00:07 \\  
 \cline{2-5} 
 & CTGAN & 0.02 & 0.29 & 01:11 \\  
 \cline{2-5} 
 & TabDDPM & 0.02 & 0.31 & 00:48 \\  
 \cline{2-5} 
 & TabSyn & \textcolor{red}{0.04 } & \textcolor{red}{0.53 } & \textcolor{red}{02:02 } \\  
 \cline{2-5} 
 & SMOTE & 0.00 & \textcolor{ForestGreen}{\textbf{0.00} } & 00:00 \\  
 \cline{2-5} 
 & UC-SMOTE & 0.00 & 0.00 & \textcolor{ForestGreen}{\textbf{00:00} } \\  
 \hline
 \pagebreak
\multirow{3}{*}{Covertype} & TVAE & 0.19 & 2.77 & 11:34 \\  
 \cline{2-5} 
 & CTGAN & \textcolor{red}{0.54 } & \textcolor{red}{8.04 } & \textcolor{red}{33:17 } \\  
 \cline{2-5} 
 & TabDDPM & 0.03 & 0.49 & 01:48 \\  
 \cline{2-5} 
 & TabSyn & 0.19 & 1.53 & 07:54 \\  
 \cline{2-5} 
 & SMOTE & \textcolor{ForestGreen}{\textbf{0.02} } & \textcolor{ForestGreen}{\textbf{0.26} } & \textcolor{ForestGreen}{\textbf{01:19} } \\  
 \cline{2-5} 
 & UC-SMOTE & 0.02 & 0.28 & 01:27 \\  
 \hline 
\multirow{3}{*}{Diamonds} & TVAE & 0.01 & 0.10 & 00:25 \\  
 \cline{2-5} 
 & CTGAN & 0.03 & 0.40 & 01:39 \\  
 \cline{2-5} 
 & TabDDPM & 0.01 & 0.15 & 00:24 \\  
 \cline{2-5} 
 & TabSyn & \textcolor{red}{0.08 } & \textcolor{red}{1.01 } & \textcolor{red}{03:22 } \\  
 \cline{2-5} 
 & SMOTE & \textcolor{ForestGreen}{\textbf{0.00} } & 0.01 & 00:02 \\  
 \cline{2-5} 
 & UC-SMOTE & 0.00 & \textcolor{ForestGreen}{\textbf{0.00} } & \textcolor{ForestGreen}{\textbf{00:00} } \\  
 \hline 
\multirow{3}{*}{Heloc} & TVAE & 0.00 & 0.04 & 00:10 \\  
 \cline{2-5} 
 & CTGAN & 0.03 & 0.49 & 02:02 \\  
 \cline{2-5} 
 & TabDDPM & \textcolor{ForestGreen}{\textbf{0.00} } & 0.03 & 00:05 \\  
 \cline{2-5} 
 & TabSyn & \textcolor{red}{0.04 } & \textcolor{red}{0.54 } & \textcolor{red}{02:08 } \\  
 \cline{2-5} 
 & SMOTE & 0.00 & \textcolor{ForestGreen}{\textbf{0.00} } & \textcolor{ForestGreen}{\textbf{00:00} } \\  
 \cline{2-5} 
 & UC-SMOTE & 0.00 & 0.00 & 00:00 \\  
 \hline 
\multirow{3}{*}{Higgs} & TVAE & 0.04 & 0.53 & 02:13 \\  
 \cline{2-5} 
 & CTGAN & 0.11 & \textcolor{red}{1.61 } & 06:42 \\  
 \cline{2-5} 
 & TabDDPM & 0.01 & 0.10 & 00:15 \\  
 \cline{2-5} 
 & TabSyn & \textcolor{red}{0.18 } & 1.58 & \textcolor{red}{07:50 } \\  
 \cline{2-5} 
 & SMOTE & \textcolor{ForestGreen}{\textbf{0.00} } & \textcolor{ForestGreen}{\textbf{0.00} } & \textcolor{ForestGreen}{\textbf{00:00} } \\  
 \cline{2-5} 
 & UC-SMOTE & 0.00 & 0.00 & 00:01 \\  
 \hline 
\multirow{3}{*}{\shortstack{House\\ 16h}} & TVAE & 0.01 & 0.09 & 00:23 \\  
 \cline{2-5} 
 & CTGAN & 0.01 & 0.10 & 00:25 \\  
 \cline{2-5} 
 & TabDDPM & \textcolor{ForestGreen}{\textbf{0.00} } & 0.04 & 00:06 \\  
 \cline{2-5} 
 & TabSyn & \textcolor{red}{0.05 } & \textcolor{red}{0.58 } & \textcolor{red}{02:16 } \\  
 \cline{2-5} 
 & SMOTE & 0.00 & \textcolor{ForestGreen}{\textbf{0.00} } & \textcolor{ForestGreen}{\textbf{00:00} } \\  
 \cline{2-5} 
 & UC-SMOTE & 0.00 & 0.00 & 00:00 \\  
 \hline 
\multirow{3}{*}{Insurance} & TVAE & 0.00 & \textcolor{ForestGreen}{\textbf{0.00} } & 00:00 \\  
 \cline{2-5} 
 & CTGAN & \textcolor{ForestGreen}{\textbf{0.00} } & 0.03 & 00:06 \\  
 \cline{2-5} 
 & TabDDPM & 0.00 & 0.05 & 00:12 \\  
 \cline{2-5} 
 & TabSyn & \textcolor{red}{0.03 } & \textcolor{red}{0.35 } & \textcolor{red}{01:25 } \\  
 \cline{2-5} 
 & SMOTE & 0.00 & 0.00 & 00:00 \\  
 \cline{2-5} 
 & UC-SMOTE & 0.00 & 0.00 & \textcolor{ForestGreen}{\textbf{00:00} } \\  
 \hline 
\multirow{3}{*}{King} & TVAE & 0.01 & 0.09 & 00:23 \\  
 \cline{2-5} 
 & CTGAN & 0.01 & 0.21 & 00:51 \\  
 \cline{2-5} 
 & TabDDPM & 0.01 & 0.17 & 00:27 \\  
 \cline{2-5} 
 & TabSyn & \textcolor{red}{0.05 } & \textcolor{red}{0.66 } & \textcolor{red}{02:26 } \\  
 \cline{2-5} 
 & SMOTE & \textcolor{ForestGreen}{\textbf{0.00} } & \textcolor{ForestGreen}{\textbf{0.00} } & 00:01 \\  
 \cline{2-5} 
 & UC-SMOTE & 0.00 & 0.00 & \textcolor{ForestGreen}{\textbf{00:00} } \\  
 \hline 
 \pagebreak
\multirow{3}{*}{\shortstack{Miniboo\\ ne}} & TVAE & 0.04 & 0.56 & 02:19 \\  
 \cline{2-5} 
 & CTGAN & 0.05 & 0.73 & 02:59 \\  
 \cline{2-5} 
 & TabDDPM & 0.01 & 0.10 & 00:14 \\  
 \cline{2-5} 
 & TabSyn & \textcolor{red}{0.31 } & \textcolor{red}{2.02 } & \textcolor{red}{11:44 } \\  
 \cline{2-5} 
 & SMOTE & \textcolor{ForestGreen}{\textbf{0.00} } & \textcolor{ForestGreen}{\textbf{0.00} } & \textcolor{ForestGreen}{\textbf{00:00} } \\  
 \cline{2-5} 
 & UC-SMOTE & 0.00 & 0.01 & 00:02 \\  
 \hline 
\multirow{3}{*}{Moons} & TVAE & 0.01 & 0.08 & 00:20 \\  
 \cline{2-5} 
 & CTGAN & 0.02 & 0.29 & 01:12 \\  
 \cline{2-5} 
 & TabDDPM & 0.01 & 0.08 & 00:12 \\  
 \cline{2-5} 
 & TabSyn & \textcolor{red}{0.05 } & \textcolor{red}{0.87 } & \textcolor{red}{02:35 } \\  
 \cline{2-5} 
 & SMOTE & \textcolor{ForestGreen}{\textbf{0.00} } & \textcolor{ForestGreen}{\textbf{0.00} } & \textcolor{ForestGreen}{\textbf{00:00} } \\  
 \cline{2-5} 
 & UC-SMOTE & 0.00 & 0.00 & 00:00 \\  
 \hline 

\caption{Estimated hyperparameter search cost based on the estimated training cost of the best models presented in Table \ref{table:emissions_estimation_normalized}. All costs associated with TabSyn include those incurred by the VAE and the denoiser. The energy and emissions values are rounded to two decimals and take into account the number of trials we could run in parallel per model. The best values per metric are formatted in \textcolor{ForestGreen}{\textbf{bold green}} and the worse are in \textcolor{red}{red}.}
 \label{table:tuning_cost}
 \end{longtable}
\end{tiny}

\section{Base Models and their Tuned Versions}
\label{appendix:base_vs_tuned}

We also trained the models using their native codes and hyperparameters to provide an additional reference for comparison with their tuned versions.
Table \ref{table:results_baselines} presents the results as evaluated under the same procedure as the tuned models for C2ST, DCR-Rate, ML-Efficacy, \textit{column-wise similarity} (named "Shape"), and \textit{pair-wise correlation} (named "Pair"). 
For TabDDPM, since there is no default hyperparameters provided \cite{kotelnikov2023tabddpm}, we fixed one based on the base configuration provided in the authors' GitHub repository.


            \begin{tiny}\setlength\tabcolsep{4pt}\setstretch{1.5} %
            \begin{longtable}{|c|l|*{5}{l|}}
            
            \hline
            \multirow{2}{*}{Dataset} & \multirow{2}{*}{Model} & \multicolumn{5}{c|}{Metrics} \\
                                                        \cline{3-7}
                                            && C2ST $\downarrow$ & DCR-Rate $\downarrow$ & ML-Efficacy $\uparrow$ & Shape $\uparrow$ & Pair $\uparrow$ \\
            
            \hline
            \endfirsthead
            \caption*{Results per datasets and models under diverse metrics for the \textit{base models} (continued).}\\
            \multirow{2}{*}{Dataset} & \multirow{2}{*}{Model} & \multicolumn{5}{c|}{Metrics} \\
                                            \cline{3-7}
                                            && C2ST $\downarrow$ & DCR-Rate $\downarrow$ & ML-Efficacy $\uparrow$ & Shape $\uparrow$ & Pair $\uparrow$ \\
            
             \hline
            \endhead
            
            \multirow{3}{*}{Abalone} & Train Copy & $0.51 \pm 0.00$ & $1.00 \pm 0.00$ & $0.23 \pm 0.01$ & $0.96 \pm 0.01$ & $0.88 \pm 0.01$ \\  
 \cline{2-7} 
 & CTGAN & $0.99 \pm 0.01$ & \textcolor{ForestGreen}{\bm{$0.60 \pm 0.01$} } & $0.12 \pm 0.03$ & $0.87 \pm 0.02$ & $0.76 \pm 0.01$ \\  
 \cline{2-7} 
 & TVAE & $0.96 \pm 0.00$ & $0.61 \pm 0.02$ & $0.22 \pm 0.02$ & $0.91 \pm 0.02$ & $0.83 \pm 0.03$ \\  
 \cline{2-7} 
 & TabDDPM & \textcolor{red}{$1.00 \pm 0.00$ } & $0.64 \pm 0.01$ & \textcolor{red}{$0.00 \pm 0.00$ } & \textcolor{red}{$0.85 \pm 0.01$ } & \textcolor{red}{$0.70 \pm 0.01$ } \\  
 \cline{2-7} 
 & TabSyn & \textcolor{ForestGreen}{\bm{$0.78 \pm 0.02$} } & \textcolor{red}{$0.64 \pm 0.01$ } & \textcolor{ForestGreen}{\bm{$0.22 \pm 0.01$} } & \textcolor{ForestGreen}{\bm{$0.95 \pm 0.01$} } & \textcolor{ForestGreen}{\bm{$0.88 \pm 0.01$} } \\  
 \hline 
\multirow{3}{*}{Adult} & Train Copy & $0.50 \pm 0.00$ & $1.00 \pm 0.00$ & $0.71 \pm 0.01$ & $0.99 \pm 0.00$ & $0.98 \pm 0.00$ \\  
 \cline{2-7} 
 & CTGAN & \textcolor{red}{$0.96 \pm 0.01$ } & $0.61 \pm 0.01$ & \textcolor{red}{$0.61 \pm 0.05$ } & \textcolor{red}{$0.88 \pm 0.02$ } & \textcolor{red}{$0.82 \pm 0.02$ } \\  
 \cline{2-7} 
 & TVAE & $0.94 \pm 0.01$ & \textcolor{ForestGreen}{\bm{$0.61 \pm 0.00$} } & $0.63 \pm 0.03$ & $0.92 \pm 0.00$ & $0.85 \pm 0.01$ \\  
 \cline{2-7} 
 & TabDDPM & \textcolor{ForestGreen}{\bm{$0.66 \pm 0.01$} } & \textcolor{red}{$0.62 \pm 0.00$ } & \textcolor{ForestGreen}{\bm{$0.67 \pm 0.00$} } & \textcolor{ForestGreen}{\bm{$0.98 \pm 0.00$} } & \textcolor{ForestGreen}{\bm{$0.95 \pm 0.00$} } \\  
 \cline{2-7} 
 & TabSyn & $0.71 \pm 0.06$ & $0.62 \pm 0.00$ & $0.66 \pm 0.01$ & $0.98 \pm 0.01$ & $0.95 \pm 0.02$ \\  
 \hline 
\multirow{3}{*}{\shortstack{Bank\\ marketing}} & Train Copy & $0.50 \pm 0.00$ & $1.00 \pm 0.00$ & $0.54 \pm 0.01$ & $0.99 \pm 0.00$ & $0.98 \pm 0.01$ \\  
 \cline{2-7} 
 & CTGAN & $0.89 \pm 0.01$ & \textcolor{ForestGreen}{\bm{$0.62 \pm 0.01$} } & \textcolor{red}{$0.33 \pm 0.05$ } & $0.92 \pm 0.00$ & $0.86 \pm 0.01$ \\  
 \cline{2-7} 
 & TVAE & \textcolor{red}{$0.95 \pm 0.00$ } & $0.62 \pm 0.00$ & \textcolor{ForestGreen}{\bm{$0.52 \pm 0.02$} } & \textcolor{red}{$0.91 \pm 0.01$ } & \textcolor{red}{$0.84 \pm 0.02$ } \\  
 \cline{2-7} 
 & TabDDPM & $0.68 \pm 0.01$ & \textcolor{red}{$0.63 \pm 0.01$ } & $0.48 \pm 0.02$ & \textcolor{ForestGreen}{\bm{$0.99 \pm 0.01$} } & \textcolor{ForestGreen}{\bm{$0.96 \pm 0.00$} } \\  
 \cline{2-7} 
 & TabSyn & \textcolor{ForestGreen}{\bm{$0.65 \pm 0.04$} } & $0.63 \pm 0.01$ & $0.47 \pm 0.01$ & $0.98 \pm 0.01$ & $0.96 \pm 0.00$ \\  
 \hline 
\multirow{3}{*}{\shortstack{Bike\\ sharing}} & Train Copy & $0.50 \pm 0.01$ & $1.00 \pm 0.00$ & $0.94 \pm 0.00$ & $0.99 \pm 0.00$ & $0.53 \pm 0.00$ \\  
 \cline{2-7} 
 & CTGAN & \textcolor{red}{$1.00 \pm 0.00$ } & \textcolor{ForestGreen}{\bm{$0.61 \pm 0.00$} } & $0.37 \pm 0.03$ & $0.93 \pm 0.01$ & \textcolor{red}{$0.51 \pm 0.01$ } \\  
 \cline{2-7} 
 & TVAE & $1.00 \pm 0.00$ & $0.61 \pm 0.01$ & $0.42 \pm 0.08$ & \textcolor{red}{$0.91 \pm 0.01$ } & $0.51 \pm 0.01$ \\  
 \cline{2-7} 
 & TabDDPM & \textcolor{ForestGreen}{\bm{$0.85 \pm 0.01$} } & $0.61 \pm 0.00$ & \textcolor{ForestGreen}{\bm{$0.67 \pm 0.03$} } & \textcolor{ForestGreen}{\bm{$0.98 \pm 0.00$} } & \textcolor{ForestGreen}{\bm{$0.53 \pm 0.00$} } \\  
 \cline{2-7} 
 & TabSyn & $0.94 \pm 0.03$ & $0.61 \pm 0.00$ & \textcolor{red}{$0.31 \pm 0.04$ } & $0.95 \pm 0.01$ & $0.52 \pm 0.01$ \\  
 \hline 
\multirow{3}{*}{\shortstack{Black\\ friday}} & Train Copy & $0.50 \pm 0.00$ & $1.00 \pm 0.00$ & $0.53 \pm 0.01$ & $1.00 \pm 0.00$ & $0.99 \pm 0.00$ \\  
 \cline{2-7} 
 & CTGAN & $0.92 \pm 0.01$ & \textcolor{ForestGreen}{\bm{$0.66 \pm 0.01$} } & \textcolor{ForestGreen}{\bm{$0.40 \pm 0.01$} } & $0.94 \pm 0.01$ & $0.84 \pm 0.01$ \\  
 \cline{2-7} 
 & TVAE & \textcolor{red}{$0.98 \pm 0.00$ } & $0.67 \pm 0.01$ & $0.29 \pm 0.08$ & \textcolor{red}{$0.83 \pm 0.01$ } & \textcolor{red}{$0.73 \pm 0.01$ } \\  
 \cline{2-7} 
 & TabDDPM & $0.92 \pm 0.00$ & $0.67 \pm 0.01$ & $0.31 \pm 0.03$ & \textcolor{ForestGreen}{\bm{$0.99 \pm 0.00$} } & \textcolor{ForestGreen}{\bm{$0.98 \pm 0.00$} } \\  
 \cline{2-7} 
 & TabSyn & \textcolor{ForestGreen}{\bm{$0.91 \pm 0.01$} } & \textcolor{red}{$0.68 \pm 0.00$ } & \textcolor{red}{$0.13 \pm 0.03$ } & $0.98 \pm 0.00$ & $0.96 \pm 0.01$ \\  
 \hline 
\multirow{3}{*}{Cardio} & Train Copy & $0.50 \pm 0.00$ & $1.00 \pm 0.00$ & $0.72 \pm 0.01$ & $1.00 \pm 0.00$ & $0.98 \pm 0.01$ \\  
 \cline{2-7} 
 & CTGAN & $1.00 \pm 0.01$ & \textcolor{ForestGreen}{\bm{$0.62 \pm 0.01$} } & $0.67 \pm 0.02$ & $0.93 \pm 0.01$ & $0.93 \pm 0.01$ \\  
 \cline{2-7} 
 & TVAE & \textcolor{red}{$1.00 \pm 0.00$ } & $0.63 \pm 0.00$ & \textcolor{red}{$0.67 \pm 0.04$ } & \textcolor{red}{$0.88 \pm 0.01$ } & \textcolor{red}{$0.90 \pm 0.02$ } \\  
 \cline{2-7} 
 & TabDDPM & $0.59 \pm 0.00$ & \textcolor{red}{$0.64 \pm 0.00$ } & \textcolor{ForestGreen}{\bm{$0.72 \pm 0.01$} } & \textcolor{ForestGreen}{\bm{$0.99 \pm 0.00$} } & $0.96 \pm 0.02$ \\  
 \cline{2-7} 
 & TabSyn & \textcolor{ForestGreen}{\bm{$0.58 \pm 0.01$} } & $0.64 \pm 0.01$ & $0.72 \pm 0.01$ & $0.99 \pm 0.01$ & \textcolor{ForestGreen}{\bm{$0.97 \pm 0.01$} } \\  
 \hline 
\multirow{3}{*}{Churn} & Train Copy & $0.50 \pm 0.01$ & $1.00 \pm 0.00$ & $0.59 \pm 0.02$ & $0.95 \pm 0.00$ & $0.87 \pm 0.01$ \\  
 \cline{2-7} 
 & CTGAN & $0.87 \pm 0.02$ & \textcolor{ForestGreen}{\bm{$0.60 \pm 0.00$} } & $0.19 \pm 0.10$ & $0.85 \pm 0.02$ & $0.79 \pm 0.00$ \\  
 \cline{2-7} 
 & TVAE & $0.98 \pm 0.01$ & $0.60 \pm 0.01$ & \textcolor{ForestGreen}{\bm{$0.45 \pm 0.10$} } & $0.75 \pm 0.03$ & $0.63 \pm 0.03$ \\  
 \cline{2-7} 
 & TabDDPM & \textcolor{red}{$1.00 \pm 0.00$ } & \textcolor{red}{$0.65 \pm 0.40$ } & \textcolor{red}{$0.00 \pm 0.00$ } & \textcolor{red}{$0.59 \pm 0.02$ } & \textcolor{red}{$0.56 \pm 0.02$ } \\  
 \cline{2-7} 
 & TabSyn & \textcolor{ForestGreen}{\bm{$0.64 \pm 0.03$} } & $0.61 \pm 0.01$ & $0.43 \pm 0.13$ & \textcolor{ForestGreen}{\bm{$0.92 \pm 0.01$} } & \textcolor{ForestGreen}{\bm{$0.85 \pm 0.00$} } \\  
 \hline 
\multirow{3}{*}{Covertype} & Train Copy & $0.50 \pm 0.00$ & $1.00 \pm 0.00$ & $0.90 \pm 0.00$ & $1.00 \pm 0.00$ & $1.00 \pm 0.01$ \\  
 \cline{2-7} 
 & CTGAN & \textcolor{red}{$1.00 \pm 0.00$ } & \textcolor{ForestGreen}{\bm{$0.60 \pm 0.00$} } & \textcolor{red}{$0.66 \pm 0.01$ } & \textcolor{red}{$0.97 \pm 0.00$ } & \textcolor{red}{$0.94 \pm 0.01$ } \\  
 \cline{2-7} 
 & TVAE & $1.00 \pm 0.00$ & $0.60 \pm 0.00$ & $0.72 \pm 0.01$ & $0.98 \pm 0.00$ & $0.95 \pm 0.01$ \\  
 \cline{2-7} 
 & TabDDPM & $0.82 \pm 0.01$ & $0.60 \pm 0.00$ & $0.73 \pm 0.01$ & \textcolor{ForestGreen}{\bm{$1.00 \pm 0.00$} } & \textcolor{ForestGreen}{\bm{$0.99 \pm 0.00$} } \\  
 \cline{2-7} 
 & TabSyn & \textcolor{ForestGreen}{\bm{$0.72 \pm 0.04$} } & $0.60 \pm 0.00$ & \textcolor{ForestGreen}{\bm{$0.79 \pm 0.03$} } & $0.99 \pm 0.00$ & $0.99 \pm 0.00$ \\  
 \hline 
\multirow{3}{*}{Diamonds} & Train Copy & $0.50 \pm 0.00$ & $1.00 \pm 0.00$ & $0.98 \pm 0.00$ & $0.99 \pm 0.00$ & $0.77 \pm 0.01$ \\  
 \cline{2-7} 
 & CTGAN & \textcolor{red}{$0.98 \pm 0.00$ } & \textcolor{ForestGreen}{\bm{$0.60 \pm 0.00$} } & $0.87 \pm 0.01$ & \textcolor{red}{$0.91 \pm 0.01$ } & $0.75 \pm 0.04$ \\  
 \cline{2-7} 
 & TVAE & $0.97 \pm 0.01$ & $0.60 \pm 0.01$ & $0.91 \pm 0.01$ & $0.92 \pm 0.02$ & \textcolor{ForestGreen}{\bm{$0.76 \pm 0.03$} } \\  
 \cline{2-7} 
 & TabDDPM & \textcolor{ForestGreen}{\bm{$0.76 \pm 0.01$} } & $0.61 \pm 0.00$ & \textcolor{ForestGreen}{\bm{$0.97 \pm 0.00$} } & \textcolor{ForestGreen}{\bm{$0.99 \pm 0.01$} } & \textcolor{red}{$0.71 \pm 0.01$ } \\  
 \cline{2-7} 
 & TabSyn & $0.95 \pm 0.02$ & \textcolor{red}{$0.64 \pm 0.00$ } & \textcolor{red}{$0.19 \pm 0.06$ } & $0.96 \pm 0.00$ & $0.72 \pm 0.01$ \\  
 \hline 
\multirow{3}{*}{Heloc} & Train Copy & $0.50 \pm 0.01$ & $1.00 \pm 0.00$ & $0.70 \pm 0.01$ & $0.99 \pm 0.00$ & $0.97 \pm 0.01$ \\  
 \cline{2-7} 
 & CTGAN & \textcolor{red}{$1.00 \pm 0.00$ } & \textcolor{ForestGreen}{\bm{$0.63 \pm 0.00$} } & \textcolor{red}{$0.40 \pm 0.14$ } & \textcolor{red}{$0.87 \pm 0.05$ } & $0.87 \pm 0.07$ \\  
 \cline{2-7} 
 & TVAE & $0.98 \pm 0.01$ & $0.63 \pm 0.02$ & $0.69 \pm 0.01$ & $0.90 \pm 0.00$ & \textcolor{red}{$0.75 \pm 0.01$ } \\  
 \cline{2-7} 
 & TabDDPM & \textcolor{ForestGreen}{\bm{$0.71 \pm 0.01$} } & $0.64 \pm 0.02$ & \textcolor{ForestGreen}{\bm{$0.70 \pm 0.01$} } & \textcolor{ForestGreen}{\bm{$0.97 \pm 0.00$} } & $0.94 \pm 0.01$ \\  
 \cline{2-7} 
 & TabSyn & $0.75 \pm 0.01$ & \textcolor{red}{$0.65 \pm 0.01$ } & $0.69 \pm 0.01$ & $0.97 \pm 0.00$ & \textcolor{ForestGreen}{\bm{$0.96 \pm 0.01$} } \\  
 \hline 
\multirow{3}{*}{Higgs} & Train Copy & $0.50 \pm 0.00$ & $1.00 \pm 0.00$ & $0.74 \pm 0.00$ & $0.99 \pm 0.00$ & $0.99 \pm 0.00$ \\  
 \cline{2-7} 
 & CTGAN & \textcolor{red}{$1.00 \pm 0.00$ } & \textcolor{ForestGreen}{\bm{$0.60 \pm 0.00$} } & \textcolor{red}{$0.61 \pm 0.08$ } & $0.87 \pm 0.02$ & $0.95 \pm 0.00$ \\  
 \cline{2-7} 
 & TVAE & $1.00 \pm 0.00$ & \textcolor{red}{$0.61 \pm 0.02$ } & $0.68 \pm 0.02$ & \textcolor{red}{$0.78 \pm 0.00$ } & $0.94 \pm 0.01$ \\  
 \cline{2-7} 
 & TabDDPM & $0.79 \pm 0.02$ & $0.60 \pm 0.01$ & $0.72 \pm 0.01$ & \textcolor{ForestGreen}{\bm{$0.98 \pm 0.01$} } & \textcolor{red}{$0.93 \pm 0.02$ } \\  
 \cline{2-7} 
 & TabSyn & \textcolor{ForestGreen}{\bm{$0.59 \pm 0.01$} } & $0.61 \pm 0.00$ & \textcolor{ForestGreen}{\bm{$0.73 \pm 0.00$} } & $0.94 \pm 0.00$ & \textcolor{ForestGreen}{\bm{$0.99 \pm 0.00$} } \\  
 \hline 
\multirow{3}{*}{\shortstack{House\\ 16h}} & Train Copy & $0.50 \pm 0.01$ & $1.00 \pm 0.00$ & $0.64 \pm 0.01$ & $0.99 \pm 0.01$ & $0.99 \pm 0.00$ \\  
 \cline{2-7} 
 & CTGAN & \textcolor{red}{$1.00 \pm 0.00$ } & \textcolor{ForestGreen}{\bm{$0.61 \pm 0.02$} } & \textcolor{red}{$0.20 \pm 0.05$ } & \textcolor{red}{$0.86 \pm 0.01$ } & \textcolor{red}{$0.95 \pm 0.01$ } \\  
 \cline{2-7} 
 & TVAE & $0.98 \pm 0.00$ & $0.61 \pm 0.01$ & $0.36 \pm 0.02$ & $0.89 \pm 0.00$ & $0.96 \pm 0.01$ \\  
 \cline{2-7} 
 & TabDDPM & \textcolor{ForestGreen}{\bm{$0.63 \pm 0.01$} } & $0.61 \pm 0.01$ & \textcolor{ForestGreen}{\bm{$0.57 \pm 0.01$} } & \textcolor{ForestGreen}{\bm{$0.98 \pm 0.00$} } & $0.97 \pm 0.01$ \\  
 \cline{2-7} 
 & TabSyn & $0.84 \pm 0.03$ & \textcolor{red}{$0.62 \pm 0.00$ } & $0.33 \pm 0.06$ & $0.96 \pm 0.01$ & \textcolor{ForestGreen}{\bm{$0.98 \pm 0.01$} } \\  
 \hline 
\multirow{3}{*}{Insurance} & Train Copy & $0.48 \pm 0.01$ & $1.00 \pm 0.00$ & $0.85 \pm 0.03$ & $0.96 \pm 0.01$ & $0.91 \pm 0.01$ \\  
 \cline{2-7} 
 & CTGAN & \textcolor{red}{$0.92 \pm 0.01$ } & \textcolor{red}{$0.67 \pm 0.06$ } & \textcolor{red}{$-0.18 \pm 0.05$ } & \textcolor{red}{$0.84 \pm 0.00$ } & $0.81 \pm 0.01$ \\  
 \cline{2-7} 
 & TVAE & $0.87 \pm 0.02$ & $0.62 \pm 0.02$ & $0.61 \pm 0.03$ & $0.85 \pm 0.03$ & \textcolor{red}{$0.77 \pm 0.03$ } \\  
 \cline{2-7} 
 & TabDDPM & \textcolor{ForestGreen}{\bm{$0.58 \pm 0.01$} } & \textcolor{ForestGreen}{\bm{$0.60 \pm 0.02$} } & \textcolor{ForestGreen}{\bm{$0.83 \pm 0.02$} } & \textcolor{ForestGreen}{\bm{$0.95 \pm 0.01$} } & \textcolor{ForestGreen}{\bm{$0.90 \pm 0.01$} } \\  
 \cline{2-7} 
 & TabSyn & $0.60 \pm 0.03$ & $0.61 \pm 0.02$ & $0.83 \pm 0.01$ & $0.94 \pm 0.01$ & $0.89 \pm 0.01$ \\  
 \hline 
\multirow{3}{*}{King} & Train Copy & $0.50 \pm 0.01$ & $1.00 \pm 0.00$ & $0.87 \pm 0.00$ & $0.98 \pm 0.01$ & $0.97 \pm 0.01$ \\  
 \cline{2-7} 
 & CTGAN & \textcolor{red}{$1.00 \pm 0.00$ } & \textcolor{ForestGreen}{\bm{$0.60 \pm 0.01$} } & $0.54 \pm 0.04$ & $0.89 \pm 0.01$ & $0.93 \pm 0.00$ \\  
 \cline{2-7} 
 & TVAE & $0.99 \pm 0.01$ & $0.60 \pm 0.00$ & \textcolor{ForestGreen}{\bm{$0.70 \pm 0.02$} } & $0.92 \pm 0.01$ & $0.91 \pm 0.01$ \\  
 \cline{2-7} 
 & TabDDPM & $1.00 \pm 0.00$ & \textcolor{red}{$0.91 \pm 0.14$ } & \textcolor{red}{$-192.96 \pm 153.37$ } & \textcolor{red}{$0.33 \pm 0.03$ } & \textcolor{red}{$0.74 \pm 0.02$ } \\  
 \cline{2-7} 
 & TabSyn & \textcolor{ForestGreen}{\bm{$0.97 \pm 0.01$} } & $0.62 \pm 0.01$ & $0.05 \pm 0.04$ & \textcolor{ForestGreen}{\bm{$0.96 \pm 0.01$} } & \textcolor{ForestGreen}{\bm{$0.94 \pm 0.00$} } \\  
 \hline 
\multirow{3}{*}{\shortstack{Miniboo\\ ne}} & Train Copy & $0.50 \pm 0.00$ & $1.00 \pm 0.00$ & $0.89 \pm 0.01$ & $0.99 \pm 0.00$ & $0.99 \pm 0.01$ \\  
 \cline{2-7} 
 & CTGAN & \textcolor{red}{$1.00 \pm 0.00$ } & \textcolor{ForestGreen}{\bm{$0.60 \pm 0.00$} } & \textcolor{red}{$0.54 \pm 0.05$ } & \textcolor{red}{$0.80 \pm 0.03$ } & $0.56 \pm 0.01$ \\  
 \cline{2-7} 
 & TVAE & $1.00 \pm 0.00$ & $0.60 \pm 0.00$ & $0.82 \pm 0.01$ & $0.91 \pm 0.02$ & \textcolor{red}{$0.56 \pm 0.00$ } \\  
 \cline{2-7} 
 & TabDDPM & $0.82 \pm 0.01$ & \textcolor{red}{$0.63 \pm 0.01$ } & $0.88 \pm 0.00$ & $0.95 \pm 0.02$ & $0.83 \pm 0.03$ \\  
 \cline{2-7} 
 & TabSyn & \textcolor{ForestGreen}{\bm{$0.71 \pm 0.09$} } & $0.60 \pm 0.00$ & \textcolor{ForestGreen}{\bm{$0.89 \pm 0.01$} } & \textcolor{ForestGreen}{\bm{$0.98 \pm 0.01$} } & \textcolor{ForestGreen}{\bm{$0.89 \pm 0.05$} } \\  
 \hline 
\multirow{3}{*}{Moons} & Train Copy & $0.50 \pm 0.00$ & $1.00 \pm 0.00$ & $1.00 \pm 0.00$ & $0.99 \pm 0.00$ & $0.99 \pm 0.00$ \\  
 \cline{2-7} 
 & CTGAN & \textcolor{red}{$0.93 \pm 0.01$ } & $0.60 \pm 0.05$ & \textcolor{ForestGreen}{\bm{$1.00 \pm 0.00$} } & $0.94 \pm 0.01$ & \textcolor{red}{$0.70 \pm 0.03$ } \\  
 \cline{2-7} 
 & TVAE & $0.82 \pm 0.02$ & $0.61 \pm 0.03$ & $1.00 \pm 0.00$ & $0.97 \pm 0.01$ & $0.76 \pm 0.02$ \\  
 \cline{2-7} 
 & TabDDPM & $0.61 \pm 0.16$ & \textcolor{ForestGreen}{\bm{$0.53 \pm 0.15$} } & \textcolor{red}{$0.97 \pm 0.06$ } & \textcolor{red}{$0.93 \pm 0.10$ } & $0.90 \pm 0.16$ \\  
 \cline{2-7} 
 & TabSyn & \textcolor{ForestGreen}{\bm{$0.58 \pm 0.03$} } & \textcolor{red}{$0.61 \pm 0.01$ } & $1.00 \pm 0.00$ & \textcolor{ForestGreen}{\bm{$0.98 \pm 0.01$} } & \textcolor{ForestGreen}{\bm{$0.98 \pm 0.01$} } \\  
 \hline

 \caption{Results for \textit{\textbf{base models}}. Models are trained using their default  hyperparameters as provided by the authors in their papers. The best values per metric are formatted in \textcolor{ForestGreen}{\textbf{bold green}} and the worse values are in \textcolor{red}{red}.}
 \label{table:results_baselines}
 \end{longtable}
\end{tiny}

\newpage
\section{Quick Search on Reduced Hyperparameter Space}
\label{appendix:light_experiment}

\begin{figure}[!htb]
     \centering

    \begin{adjustbox}{minipage=\linewidth,scale=0.9}

    \subfigure[C2ST]{\includegraphics[width=0.9\textwidth]{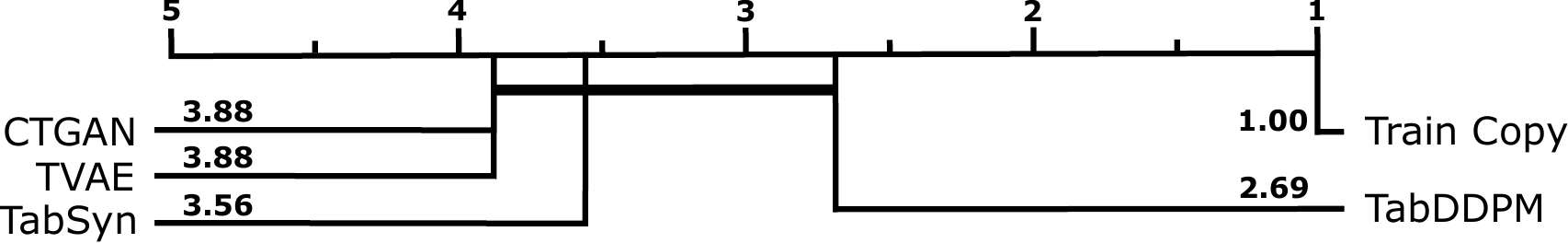}
    \label{fig:cdd_c2st_light}
    }

    \subfigure[Column-wise similarity]{\includegraphics[width=0.9 
 \textwidth]{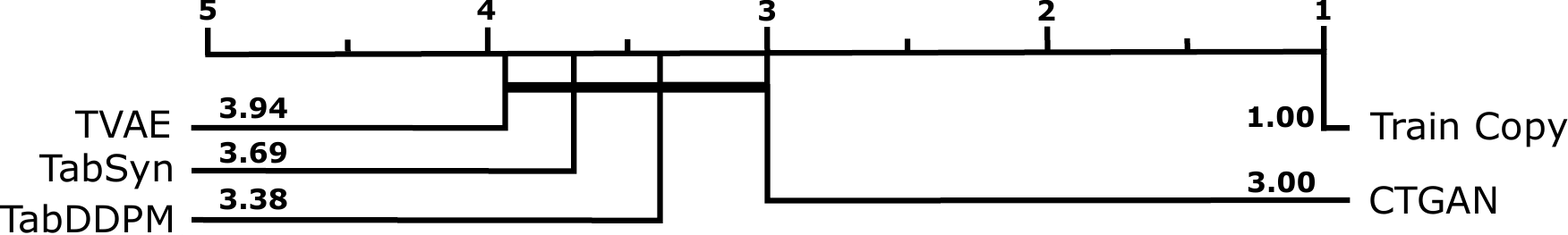}
    \label{fig:cdd_shape_light}
    }

        \subfigure[Pair-wise correlation]{\includegraphics[width=0.9 
 \textwidth]{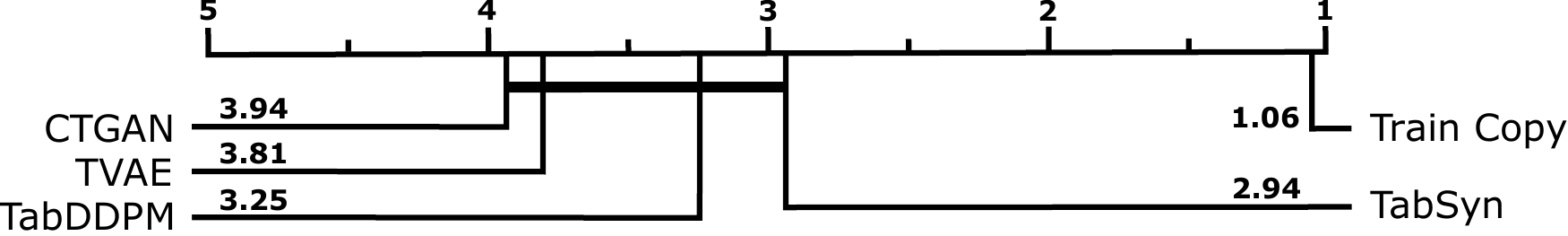}
    \label{fig:cdd_pair_light}
    }

        \subfigure[DCR-Rate]{\includegraphics[width=0.9 
 \textwidth]{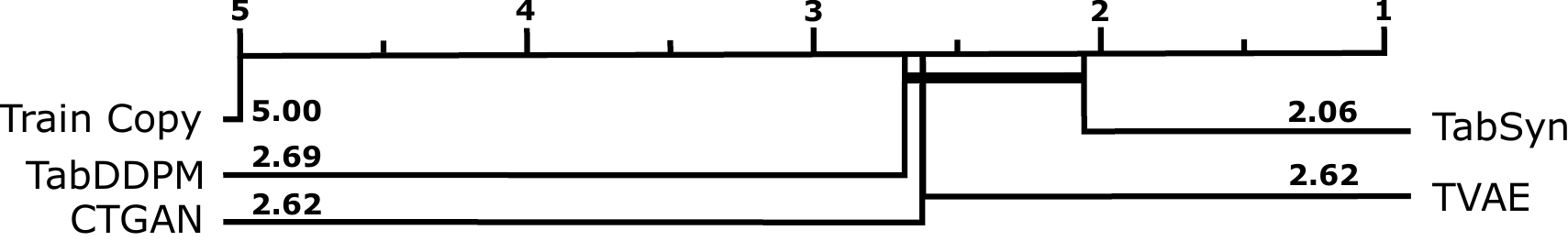}
    \label{fig:cdd_dcr_rate_light}
    }
    
    \subfigure[ML-Efficacy]{\includegraphics[width=0.9 
 \textwidth]{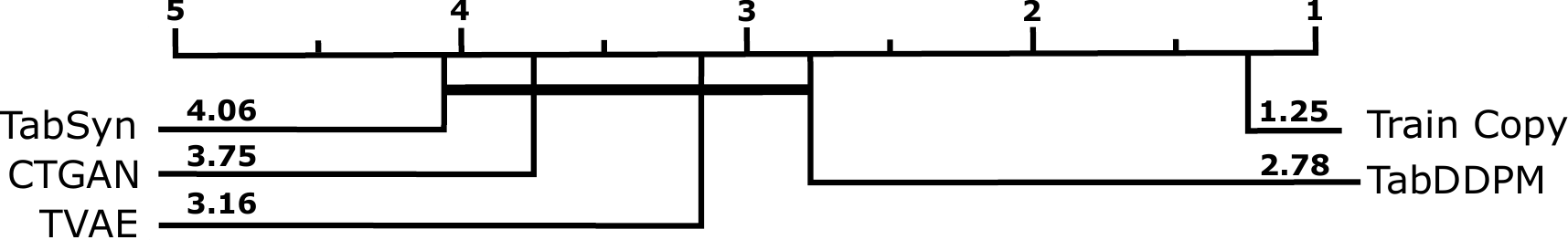}
    \label{fig:cdd_mle_light}
    }
    
    \end{adjustbox}

    \caption{Models' ranking under the \textit{light} (limited-budget) experiment setup with critical difference diagrams on C2ST, \textit{column-wise similarity}, \textit{pair-wise correlation}, DCR-Rate, and ML-Efficacy metrics over all datasets.}
    \label{fig:cdd_light_experiment}
    
\end{figure}

With the reduced hyperparameters search spaces presented in \ref{appendix:hp_search_space}, we ran a limited-budget hyperparameter tuning described in Section \ref{sec:hpsens}. The experiment was done on all dataset (Table \ref{tab:datasets}). We performed 50 trials per fold with 3 folds, meaning we ran a total of 150 trials for this limited-budget experiment. The results par datasets are shown in Table~\ref{table:results_light}.

            

            \begin{tiny}\setlength\tabcolsep{4pt}\setstretch{1.5} %
            \begin{longtable}{|c|l|*{6}{l|}}
            
            \hline
            \multirow{2}{*}{Dataset} & \multirow{2}{*}{Model} & \multicolumn{6}{c|}{Metrics} \\
                                            \cline{3-8}
                                            && C2ST $\downarrow$ & DCR-Rate $\downarrow$ & ML-Efficacy $\uparrow$ & Shape $\uparrow$ & Pair $\uparrow$ & Sampling time $\downarrow$ \\
            
            \hline
            \endfirsthead
            \caption*{Results per datasets and models under diverse metrics for the limited-budget search (continued).}\\
            \multirow{2}{*}{Dataset} & \multirow{2}{*}{Model} & \multicolumn{6}{c|}{Metrics} \\
                                            \cline{3-8}
                                            && C2ST $\downarrow$ & DCR-Rate $\downarrow$ & ML-Efficacy $\uparrow$ & Shape $\uparrow$ & Pair $\uparrow$ & Sampling time $\downarrow$ \\
            
             \hline
            \endhead
            
            \multirow{3}{*}{Abalone} & Train Copy & $0.51 \pm 0.00$ & $1.00 \pm 0.00$ & $0.23 \pm 0.01$ & $0.96 \pm 0.01$ & $0.88 \pm 0.01$ & - \\  
 \cline{2-8} 
 & CTGAN & $0.71 \pm 0.03$ & $0.63 \pm 0.00$ & \textcolor{red}{$0.17 \pm 0.01$ } & $0.93 \pm 0.02$ & $0.86 \pm 0.04$ & $00 \pm 0.01$ \\  
 \cline{2-8} 
 & TVAE & \textcolor{ForestGreen}{\bm{$0.64 \pm 0.02$} } & $0.67 \pm 0.00$ & $0.23 \pm 0.01$ & \textcolor{red}{$0.93 \pm 0.02$ } & \textcolor{ForestGreen}{\bm{$0.89 \pm 0.03$} } & \textcolor{ForestGreen}{\bm{$00 \pm 0.00$} } \\  
 \cline{2-8} 
 & TabDDPM & $0.78 \pm 0.01$ & \textcolor{red}{$0.69 \pm 0.03$ } & \textcolor{ForestGreen}{\bm{$0.23 \pm 0.01$} } & \textcolor{ForestGreen}{\bm{$0.95 \pm 0.00$} } & $0.88 \pm 0.02$ & \textcolor{red}{$03 \pm 0.88$ } \\  
 \cline{2-8} 
 & TabSyn & \textcolor{red}{$0.80 \pm 0.00$ } & \textcolor{ForestGreen}{\bm{$0.62 \pm 0.01$} } & $0.19 \pm 0.03$ & $0.93 \pm 0.01$ & \textcolor{red}{$0.85 \pm 0.01$ } & $00 \pm 0.01$ \\  
 \hline 
\multirow{3}{*}{Adult} & Train Copy & $0.50 \pm 0.00$ & $1.00 \pm 0.00$ & $0.71 \pm 0.01$ & $0.99 \pm 0.00$ & $0.98 \pm 0.00$ & - \\  
 \cline{2-8} 
 & CTGAN & $0.77 \pm 0.01$ & $0.72 \pm 0.00$ & \textcolor{red}{$0.65 \pm 0.00$ } & \textcolor{red}{$0.96 \pm 0.01$ } & \textcolor{red}{$0.89 \pm 0.02$ } & $00 \pm 0.02$ \\  
 \cline{2-8} 
 & TVAE & \textcolor{red}{$0.77 \pm 0.01$ } & \textcolor{red}{$0.72 \pm 0.01$ } & $0.65 \pm 0.02$ & $0.96 \pm 0.01$ & $0.93 \pm 0.00$ & \textcolor{ForestGreen}{\bm{$00 \pm 0.00$} } \\  
 \cline{2-8} 
 & TabDDPM & \textcolor{ForestGreen}{\bm{$0.67 \pm 0.01$} } & \textcolor{ForestGreen}{\bm{$0.62 \pm 0.00$} } & \textcolor{ForestGreen}{\bm{$0.67 \pm 0.01$} } & \textcolor{ForestGreen}{\bm{$0.97 \pm 0.01$} } & \textcolor{ForestGreen}{\bm{$0.94 \pm 0.01$} } & \textcolor{red}{$11 \pm 0.63$ } \\  
 \cline{2-8} 
 & TabSyn & $0.73 \pm 0.01$ & $0.62 \pm 0.00$ & $0.66 \pm 0.01$ & $0.97 \pm 0.01$ & $0.94 \pm 0.01$ & $02 \pm 0.01$ \\  
 \hline 
\multirow{3}{*}{\shortstack{Bank\\ marketing}} & Train Copy & $0.50 \pm 0.00$ & $1.00 \pm 0.00$ & $0.54 \pm 0.01$ & $0.99 \pm 0.00$ & $0.98 \pm 0.01$ & - \\  
 \cline{2-8} 
 & CTGAN & $0.72 \pm 0.01$ & \textcolor{red}{$0.63 \pm 0.00$ } & $0.45 \pm 0.04$ & \textcolor{ForestGreen}{\bm{$0.98 \pm 0.00$} } & $0.94 \pm 0.01$ & $00 \pm 0.01$ \\  
 \cline{2-8} 
 & TVAE & \textcolor{red}{$0.78 \pm 0.02$ } & $0.63 \pm 0.00$ & $0.47 \pm 0.03$ & \textcolor{red}{$0.97 \pm 0.01$ } & \textcolor{red}{$0.94 \pm 0.00$ } & \textcolor{ForestGreen}{\bm{$00 \pm 0.00$} } \\  
 \cline{2-8} 
 & TabDDPM & \textcolor{ForestGreen}{\bm{$0.68 \pm 0.00$} } & $0.63 \pm 0.01$ & \textcolor{ForestGreen}{\bm{$0.52 \pm 0.03$} } & $0.97 \pm 0.00$ & \textcolor{ForestGreen}{\bm{$0.95 \pm 0.01$} } & \textcolor{red}{$16 \pm 2.18$ } \\  
 \cline{2-8} 
 & TabSyn & $0.75 \pm 0.00$ & \textcolor{ForestGreen}{\bm{$0.62 \pm 0.00$} } & \textcolor{red}{$0.42 \pm 0.04$ } & $0.97 \pm 0.01$ & $0.95 \pm 0.00$ & $02 \pm 0.01$ \\  
 \hline 
\multirow{3}{*}{\shortstack{Bike\\ sharing}} & Train Copy & $0.50 \pm 0.01$ & $1.00 \pm 0.00$ & $0.94 \pm 0.00$ & $0.99 \pm 0.00$ & $0.53 \pm 0.00$ & - \\  
 \cline{2-8} 
 & CTGAN & $0.89 \pm 0.01$ & $0.61 \pm 0.01$ & $0.71 \pm 0.01$ & \textcolor{ForestGreen}{\bm{$0.98 \pm 0.01$} } & \textcolor{ForestGreen}{\bm{$0.53 \pm 0.00$} } & $00 \pm 0.00$ \\  
 \cline{2-8} 
 & TVAE & $0.81 \pm 0.01$ & $0.61 \pm 0.01$ & $0.69 \pm 0.03$ & $0.96 \pm 0.00$ & $0.53 \pm 0.00$ & \textcolor{ForestGreen}{\bm{$00 \pm 0.00$} } \\  
 \cline{2-8} 
 & TabDDPM & \textcolor{ForestGreen}{\bm{$0.80 \pm 0.02$} } & \textcolor{red}{$0.63 \pm 0.01$ } & \textcolor{ForestGreen}{\bm{$0.83 \pm 0.01$} } & $0.96 \pm 0.02$ & $0.53 \pm 0.01$ & \textcolor{red}{$06 \pm 1.34$ } \\  
 \cline{2-8} 
 & TabSyn & \textcolor{red}{$0.91 \pm 0.01$ } & \textcolor{ForestGreen}{\bm{$0.61 \pm 0.01$} } & \textcolor{red}{$0.32 \pm 0.06$ } & \textcolor{red}{$0.95 \pm 0.01$ } & \textcolor{red}{$0.52 \pm 0.00$ } & $00 \pm 0.01$ \\  
 \hline 
\multirow{3}{*}{\shortstack{Black\\ friday}} & Train Copy & $0.50 \pm 0.00$ & $1.00 \pm 0.00$ & $0.53 \pm 0.01$ & $1.00 \pm 0.00$ & $0.99 \pm 0.00$ & - \\  
 \cline{2-8} 
 & CTGAN & \textcolor{ForestGreen}{\bm{$0.87 \pm 0.01$} } & \textcolor{red}{$0.68 \pm 0.00$ } & $0.36 \pm 0.01$ & \textcolor{ForestGreen}{\bm{$0.97 \pm 0.01$} } & \textcolor{red}{$0.91 \pm 0.02$ } & $01 \pm 0.17$ \\  
 \cline{2-8} 
 & TVAE & \textcolor{red}{$0.95 \pm 0.01$ } & $0.68 \pm 0.00$ & $0.27 \pm 0.02$ & $0.97 \pm 0.00$ & $0.95 \pm 0.00$ & \textcolor{ForestGreen}{\bm{$00 \pm 0.02$} } \\  
 \cline{2-8} 
 & TabDDPM & $0.89 \pm 0.01$ & \textcolor{ForestGreen}{\bm{$0.67 \pm 0.00$} } & \textcolor{ForestGreen}{\bm{$0.44 \pm 0.02$} } & \textcolor{red}{$0.97 \pm 0.01$ } & \textcolor{ForestGreen}{\bm{$0.96 \pm 0.01$} } & \textcolor{red}{$50 \pm 17.19$ } \\  
 \cline{2-8} 
 & TabSyn & $0.89 \pm 0.01$ & $0.68 \pm 0.00$ & \textcolor{red}{$0.16 \pm 0.01$ } & $0.97 \pm 0.01$ & $0.95 \pm 0.01$ & $07 \pm 0.01$ \\  
 \hline 
\multirow{3}{*}{Cardio} & Train Copy & $0.50 \pm 0.00$ & $1.00 \pm 0.00$ & $0.72 \pm 0.01$ & $1.00 \pm 0.00$ & $0.98 \pm 0.01$ & - \\  
 \cline{2-8} 
 & CTGAN & $0.64 \pm 0.01$ & \textcolor{red}{$0.64 \pm 0.01$ } & $0.72 \pm 0.01$ & \textcolor{ForestGreen}{\bm{$0.99 \pm 0.00$} } & \textcolor{red}{$0.96 \pm 0.00$ } & $00 \pm 0.06$ \\  
 \cline{2-8} 
 & TVAE & \textcolor{red}{$0.71 \pm 0.01$ } & $0.64 \pm 0.01$ & \textcolor{ForestGreen}{\bm{$0.73 \pm 0.01$} } & \textcolor{red}{$0.96 \pm 0.01$ } & $0.96 \pm 0.01$ & \textcolor{ForestGreen}{\bm{$00 \pm 0.01$} } \\  
 \cline{2-8} 
 & TabDDPM & \textcolor{ForestGreen}{\bm{$0.57 \pm 0.01$} } & \textcolor{ForestGreen}{\bm{$0.64 \pm 0.00$} } & \textcolor{red}{$0.72 \pm 0.01$ } & $0.99 \pm 0.00$ & \textcolor{ForestGreen}{\bm{$0.98 \pm 0.01$} } & \textcolor{red}{$12 \pm 4.58$ } \\  
 \cline{2-8} 
 & TabSyn & $0.59 \pm 0.01$ & $0.64 \pm 0.00$ & $0.72 \pm 0.01$ & $0.98 \pm 0.01$ & $0.97 \pm 0.01$ & $03 \pm 0.01$ \\  
 \hline 
\multirow{3}{*}{Churn} & Train Copy & $0.50 \pm 0.01$ & $1.00 \pm 0.00$ & $0.59 \pm 0.02$ & $0.95 \pm 0.00$ & $0.87 \pm 0.01$ & - \\  
 \cline{2-8} 
 & CTGAN & \textcolor{ForestGreen}{\bm{$0.63 \pm 0.01$} } & $0.63 \pm 0.01$ & $0.39 \pm 0.01$ & \textcolor{ForestGreen}{\bm{$0.93 \pm 0.00$} } & \textcolor{ForestGreen}{\bm{$0.84 \pm 0.01$} } & $00 \pm 0.01$ \\  
 \cline{2-8} 
 & TVAE & $0.64 \pm 0.00$ & $0.63 \pm 0.01$ & \textcolor{ForestGreen}{\bm{$0.53 \pm 0.01$} } & $0.92 \pm 0.00$ & $0.84 \pm 0.00$ & \textcolor{ForestGreen}{\bm{$00 \pm 0.00$} } \\  
 \cline{2-8} 
 & TabDDPM & \textcolor{red}{$0.98 \pm 0.03$ } & \textcolor{red}{$0.67 \pm 0.18$ } & \textcolor{red}{$0.02 \pm 0.03$ } & \textcolor{red}{$0.59 \pm 0.05$ } & \textcolor{red}{$0.55 \pm 0.02$ } & \textcolor{red}{$18 \pm 1.06$ } \\  
 \cline{2-8} 
 & TabSyn & $0.68 \pm 0.06$ & \textcolor{ForestGreen}{\bm{$0.61 \pm 0.00$} } & $0.46 \pm 0.03$ & $0.91 \pm 0.01$ & $0.84 \pm 0.01$ & $00 \pm 0.01$ \\  
 \hline 
 \pagebreak
\multirow{3}{*}{Covertype} & Train Copy & $0.50 \pm 0.00$ & $1.00 \pm 0.00$ & $0.90 \pm 0.00$ & $1.00 \pm 0.00$ & $1.00 \pm 0.01$ & - \\  
 \cline{2-8} 
 & CTGAN & \textcolor{red}{$0.98 \pm 0.00$ } & \textcolor{ForestGreen}{\bm{$0.60 \pm 0.00$} } & $0.69 \pm 0.00$ & \textcolor{ForestGreen}{\bm{$0.98 \pm 0.00$} } & $0.95 \pm 0.01$ & $06 \pm 1.58$ \\  
 \cline{2-8} 
 & TVAE & $0.90 \pm 0.01$ & $0.60 \pm 0.00$ & \textcolor{ForestGreen}{\bm{$0.77 \pm 0.01$} } & $0.98 \pm 0.00$ & \textcolor{ForestGreen}{\bm{$0.96 \pm 0.00$} } & $02 \pm 0.07$ \\  
 \cline{2-8} 
 & TabDDPM & $0.97 \pm 0.01$ & \textcolor{red}{$0.64 \pm 0.02$ } & $0.69 \pm 0.00$ & \textcolor{red}{$0.94 \pm 0.02$ } & $0.89 \pm 0.03$ & \textcolor{red}{$355 \pm 10.90$ } \\  
 \cline{2-8} 
 & TabSyn & \textcolor{ForestGreen}{\bm{$0.87 \pm 0.00$} } & $0.61 \pm 0.01$ & \textcolor{red}{$0.65 \pm 0.03$ } & $0.98 \pm 0.00$ & \textcolor{red}{$0.68 \pm 0.00$ } & \textcolor{ForestGreen}{\bm{$02 \pm 0.02$} } \\  
 \hline 
\multirow{3}{*}{Diamonds} & Train Copy & $0.50 \pm 0.00$ & $1.00 \pm 0.00$ & $0.98 \pm 0.00$ & $0.99 \pm 0.00$ & $0.77 \pm 0.01$ & - \\  
 \cline{2-8} 
 & CTGAN & $0.89 \pm 0.01$ & \textcolor{red}{$0.64 \pm 0.01$ } & $0.94 \pm 0.00$ & \textcolor{red}{$0.95 \pm 0.00$ } & \textcolor{ForestGreen}{\bm{$0.74 \pm 0.02$} } & $00 \pm 0.08$ \\  
 \cline{2-8} 
 & TVAE & $0.76 \pm 0.02$ & $0.64 \pm 0.00$ & $0.96 \pm 0.01$ & $0.95 \pm 0.01$ & $0.73 \pm 0.01$ & \textcolor{ForestGreen}{\bm{$00 \pm 0.01$} } \\  
 \cline{2-8} 
 & TabDDPM & \textcolor{ForestGreen}{\bm{$0.75 \pm 0.01$} } & \textcolor{ForestGreen}{\bm{$0.61 \pm 0.01$} } & \textcolor{ForestGreen}{\bm{$0.97 \pm 0.00$} } & \textcolor{ForestGreen}{\bm{$0.97 \pm 0.01$} } & \textcolor{red}{$0.70 \pm 0.02$ } & \textcolor{red}{$10 \pm 1.26$ } \\  
 \cline{2-8} 
 & TabSyn & \textcolor{red}{$0.92 \pm 0.01$ } & $0.64 \pm 0.01$ & \textcolor{red}{$0.65 \pm 0.09$ } & $0.95 \pm 0.01$ & $0.72 \pm 0.02$ & $02 \pm 0.01$ \\  
 \hline 
\multirow{3}{*}{Heloc} & Train Copy & $0.50 \pm 0.01$ & $1.00 \pm 0.00$ & $0.70 \pm 0.01$ & $0.99 \pm 0.00$ & $0.97 \pm 0.01$ & - \\  
 \cline{2-8} 
 & CTGAN & \textcolor{red}{$0.93 \pm 0.02$ } & $0.64 \pm 0.01$ & \textcolor{ForestGreen}{\bm{$0.70 \pm 0.01$} } & \textcolor{ForestGreen}{\bm{$0.97 \pm 0.00$} } & $0.93 \pm 0.01$ & $00 \pm 0.02$ \\  
 \cline{2-8} 
 & TVAE & $0.92 \pm 0.01$ & \textcolor{ForestGreen}{\bm{$0.63 \pm 0.01$} } & $0.69 \pm 0.02$ & \textcolor{red}{$0.94 \pm 0.00$ } & \textcolor{red}{$0.88 \pm 0.01$ } & \textcolor{ForestGreen}{\bm{$00 \pm 0.00$} } \\  
 \cline{2-8} 
 & TabDDPM & \textcolor{ForestGreen}{\bm{$0.71 \pm 0.02$} } & \textcolor{red}{$0.67 \pm 0.01$ } & \textcolor{red}{$0.69 \pm 0.02$ } & $0.95 \pm 0.01$ & $0.92 \pm 0.03$ & \textcolor{red}{$03 \pm 1.99$ } \\  
 \cline{2-8} 
 & TabSyn & $0.82 \pm 0.01$ & $0.65 \pm 0.01$ & $0.69 \pm 0.01$ & $0.95 \pm 0.01$ & \textcolor{ForestGreen}{\bm{$0.95 \pm 0.01$} } & $00 \pm 0.01$ \\  
 \hline 
\multirow{3}{*}{Higgs} & Train Copy & $0.50 \pm 0.00$ & $1.00 \pm 0.00$ & $0.74 \pm 0.00$ & $0.99 \pm 0.00$ & $0.99 \pm 0.00$ & - \\  
 \cline{2-8} 
 & CTGAN & $0.87 \pm 0.00$ & \textcolor{ForestGreen}{\bm{$0.60 \pm 0.01$} } & \textcolor{red}{$0.70 \pm 0.01$ } & \textcolor{ForestGreen}{\bm{$0.98 \pm 0.00$} } & \textcolor{ForestGreen}{\bm{$0.98 \pm 0.00$} } & $00 \pm 0.09$ \\  
 \cline{2-8} 
 & TVAE & \textcolor{red}{$0.92 \pm 0.01$ } & $0.60 \pm 0.01$ & $0.71 \pm 0.02$ & $0.94 \pm 0.01$ & $0.97 \pm 0.00$ & \textcolor{ForestGreen}{\bm{$00 \pm 0.03$} } \\  
 \cline{2-8} 
 & TabDDPM & \textcolor{ForestGreen}{\bm{$0.61 \pm 0.02$} } & \textcolor{red}{$0.62 \pm 0.01$ } & \textcolor{ForestGreen}{\bm{$0.73 \pm 0.00$} } & $0.98 \pm 0.01$ & \textcolor{red}{$0.95 \pm 0.01$ } & \textcolor{red}{$08 \pm 1.00$ } \\  
 \cline{2-8} 
 & TabSyn & $0.76 \pm 0.05$ & $0.60 \pm 0.01$ & $0.71 \pm 0.02$ & \textcolor{red}{$0.92 \pm 0.01$ } & $0.98 \pm 0.00$ & $04 \pm 0.01$ \\  
 \hline 
\multirow{3}{*}{\shortstack{House\\ 16h}} & Train Copy & $0.50 \pm 0.01$ & $1.00 \pm 0.00$ & $0.64 \pm 0.01$ & $0.99 \pm 0.01$ & $0.99 \pm 0.00$ & - \\  
 \cline{2-8} 
 & CTGAN & \textcolor{red}{$0.84 \pm 0.02$ } & \textcolor{red}{$0.62 \pm 0.00$ } & $0.45 \pm 0.02$ & \textcolor{ForestGreen}{\bm{$0.97 \pm 0.01$} } & $0.97 \pm 0.01$ & \textcolor{ForestGreen}{\bm{$00 \pm 0.01$} } \\  
 \cline{2-8} 
 & TVAE & $0.84 \pm 0.00$ & $0.62 \pm 0.01$ & $0.46 \pm 0.03$ & \textcolor{red}{$0.94 \pm 0.00$ } & $0.98 \pm 0.00$ & $00 \pm 0.00$ \\  
 \cline{2-8} 
 & TabDDPM & \textcolor{ForestGreen}{\bm{$0.64 \pm 0.01$} } & \textcolor{ForestGreen}{\bm{$0.61 \pm 0.01$} } & \textcolor{ForestGreen}{\bm{$0.61 \pm 0.01$} } & $0.96 \pm 0.01$ & \textcolor{red}{$0.94 \pm 0.03$ } & \textcolor{red}{$02 \pm 0.75$ } \\  
 \cline{2-8} 
 & TabSyn & $0.83 \pm 0.03$ & $0.62 \pm 0.01$ & \textcolor{red}{$0.35 \pm 0.01$ } & $0.95 \pm 0.01$ & \textcolor{ForestGreen}{\bm{$0.99 \pm 0.01$} } & $01 \pm 0.01$ \\  
 \hline 
\multirow{3}{*}{Insurance} & Train Copy & $0.48 \pm 0.01$ & $1.00 \pm 0.00$ & $0.85 \pm 0.03$ & $0.96 \pm 0.01$ & $0.91 \pm 0.01$ & - \\  
 \cline{2-8} 
 & CTGAN & \textcolor{red}{$0.67 \pm 0.02$ } & \textcolor{red}{$0.92 \pm 0.01$ } & \textcolor{red}{$0.66 \pm 0.04$ } & \textcolor{ForestGreen}{\bm{$0.94 \pm 0.01$} } & \textcolor{red}{$0.85 \pm 0.02$ } & $00 \pm 0.01$ \\  
 \cline{2-8} 
 & TVAE & $0.65 \pm 0.02$ & $0.91 \pm 0.02$ & $0.80 \pm 0.03$ & \textcolor{red}{$0.93 \pm 0.01$ } & $0.88 \pm 0.01$ & \textcolor{ForestGreen}{\bm{$00 \pm 0.00$} } \\  
 \cline{2-8} 
 & TabDDPM & \textcolor{ForestGreen}{\bm{$0.61 \pm 0.02$} } & $0.62 \pm 0.02$ & \textcolor{ForestGreen}{\bm{$0.83 \pm 0.03$} } & $0.94 \pm 0.01$ & \textcolor{ForestGreen}{\bm{$0.89 \pm 0.02$} } & \textcolor{red}{$06 \pm 2.76$ } \\  
 \cline{2-8} 
 & TabSyn & $0.63 \pm 0.02$ & \textcolor{ForestGreen}{\bm{$0.60 \pm 0.03$} } & $0.80 \pm 0.01$ & $0.93 \pm 0.01$ & $0.88 \pm 0.02$ & $00 \pm 0.01$ \\  
 \hline 
\multirow{3}{*}{King} & Train Copy & $0.50 \pm 0.01$ & $1.00 \pm 0.00$ & $0.87 \pm 0.00$ & $0.98 \pm 0.01$ & $0.97 \pm 0.01$ & - \\  
 \cline{2-8} 
 & CTGAN & $0.96 \pm 0.01$ & $0.62 \pm 0.00$ & $0.66 \pm 0.03$ & \textcolor{ForestGreen}{\bm{$0.97 \pm 0.00$} } & \textcolor{ForestGreen}{\bm{$0.94 \pm 0.01$} } & $00 \pm 0.13$ \\  
 \cline{2-8} 
 & TVAE & \textcolor{ForestGreen}{\bm{$0.95 \pm 0.01$} } & \textcolor{ForestGreen}{\bm{$0.61 \pm 0.01$} } & \textcolor{ForestGreen}{\bm{$0.80 \pm 0.01$} } & $0.95 \pm 0.01$ & $0.94 \pm 0.00$ & \textcolor{ForestGreen}{\bm{$00 \pm 0.01$} } \\  
 \cline{2-8} 
 & TabDDPM & \textcolor{red}{$1.00 \pm 0.00$ } & \textcolor{red}{$0.70 \pm 0.46$ } & $0.03 \pm 0.68$ & \textcolor{red}{$0.28 \pm 0.07$ } & \textcolor{red}{$0.76 \pm 0.00$ } & \textcolor{red}{$05 \pm 0.43$ } \\  
 \cline{2-8} 
 & TabSyn & $0.98 \pm 0.01$ & $0.62 \pm 0.01$ & \textcolor{red}{$-0.01 \pm 0.01$ } & $0.94 \pm 0.01$ & $0.93 \pm 0.00$ & $01 \pm 0.01$ \\  
 \hline 
\multirow{3}{*}{\shortstack{Miniboo\\ ne}} & Train Copy & $0.50 \pm 0.00$ & $1.00 \pm 0.00$ & $0.89 \pm 0.01$ & $0.99 \pm 0.00$ & $0.99 \pm 0.01$ & - \\  
 \cline{2-8} 
 & CTGAN & $0.90 \pm 0.01$ & \textcolor{ForestGreen}{\bm{$0.60 \pm 0.01$} } & $0.87 \pm 0.01$ & \textcolor{red}{$0.93 \pm 0.01$ } & \textcolor{red}{$0.58 \pm 0.01$ } & $01 \pm 0.03$ \\  
 \cline{2-8} 
 & TVAE & \textcolor{red}{$0.94 \pm 0.01$ } & $0.60 \pm 0.00$ & $0.87 \pm 0.00$ & $0.94 \pm 0.00$ & $0.59 \pm 0.01$ & \textcolor{ForestGreen}{\bm{$00 \pm 0.03$} } \\  
 \cline{2-8} 
 & TabDDPM & \textcolor{ForestGreen}{\bm{$0.73 \pm 0.01$} } & \textcolor{red}{$0.62 \pm 0.00$ } & \textcolor{ForestGreen}{\bm{$0.89 \pm 0.01$} } & $0.96 \pm 0.01$ & \textcolor{ForestGreen}{\bm{$0.83 \pm 0.05$} } & \textcolor{red}{$12 \pm 1.35$ } \\  
 \cline{2-8} 
 & TabSyn & $0.87 \pm 0.01$ & $0.60 \pm 0.01$ & \textcolor{red}{$0.87 \pm 0.00$ } & \textcolor{ForestGreen}{\bm{$0.96 \pm 0.00$} } & $0.76 \pm 0.14$ & $06 \pm 0.08$ \\  
 \hline 
 \pagebreak
\multirow{3}{*}{Moons} & Train Copy & $0.50 \pm 0.00$ & $1.00 \pm 0.00$ & $1.00 \pm 0.00$ & $0.99 \pm 0.00$ & $0.99 \pm 0.00$ & - \\  
 \cline{2-8} 
 & CTGAN & \textcolor{red}{$0.70 \pm 0.02$ } & \textcolor{red}{$0.62 \pm 0.02$ } & \textcolor{ForestGreen}{\bm{$1.00 \pm 0.00$} } & \textcolor{red}{$0.96 \pm 0.01$ } & $0.96 \pm 0.01$ & $00 \pm 0.02$ \\  
 \cline{2-8} 
 & TVAE & $0.69 \pm 0.01$ & $0.62 \pm 0.01$ & $1.00 \pm 0.00$ & $0.96 \pm 0.01$ & \textcolor{red}{$0.93 \pm 0.02$ } & \textcolor{ForestGreen}{\bm{$00 \pm 0.00$} } \\  
 \cline{2-8} 
 & TabDDPM & \textcolor{ForestGreen}{\bm{$0.53 \pm 0.01$} } & \textcolor{ForestGreen}{\bm{$0.61 \pm 0.01$} } & $1.00 \pm 0.00$ & \textcolor{ForestGreen}{\bm{$0.99 \pm 0.00$} } & \textcolor{ForestGreen}{\bm{$0.98 \pm 0.01$} } & \textcolor{red}{$08 \pm 2.13$ } \\  
 \cline{2-8} 
 & TabSyn & $0.59 \pm 0.01$ & $0.61 \pm 0.01$ & $1.00 \pm 0.00$ & $0.98 \pm 0.01$ & $0.97 \pm 0.01$ & $01 \pm 0.00$ \\  
 \hline 

 \caption{Limited-budget experiment results under various metrics. Results are averaged over 3 folds with 5 synthetic samples per fold as done in the extensive hyperparameter tuning. The best values per metric are formatted in \textcolor{ForestGreen}{\textbf{bold green}} and the worse values are in \textcolor{red}{red}.}
 \label{table:results_light}
 \end{longtable}
\end{tiny}

\end{document}